\documentclass[10pt,twocolumn,letterpaper]{article}

\usepackage[pagenumbers]{cvpr}      
\usepackage[scaled]{inconsolata}
 
\usepackage{makecell}
\usepackage[export]{adjustbox}
\usepackage[most]{tcolorbox}

\definecolor{indigo}{RGB}{63,81,181} 

\newtcolorbox{bluebox}[1][]{
  enhanced, breakable, sharp corners,
  boxrule=0pt, 
  colback=indigo!5!white, colframe=indigo!5!white, 
  borderline west={2pt}{0pt}{indigo!75!black}, 
  width=\linewidth,
  enlarge left by=0mm,
  enlarge right by=0mm,
  #1
}

\newtcolorbox{DefBox}[1][]{
  enhanced, breakable, sharp corners,
  boxrule=0pt,
  colback=green!5!white, colframe=green!5!white,
  borderline west={2pt}{0pt}{green!75!black},
  width=\linewidth,
  enlarge left by=0mm,
  enlarge right by=0mm,
  #1
}

\newtcolorbox{HypBox}[1][]{
  enhanced, breakable, sharp corners,
  boxrule=0pt,
  colback=orange!15!white, colframe=orange!15!white,
  borderline west={2pt}{0pt}{orange!80!black},
  width=\linewidth,
  enlarge left by=0mm,
  enlarge right by=0mm,
  #1
}

\newtcolorbox{PromptBox}[1][]{
  enhanced, breakable, sharp corners,
  boxrule=0pt,
  colback=gray!10!white, colframe=gray!10!white,
  borderline west={2pt}{0pt}{gray!75!black},
  width=\linewidth,
  enlarge left by=0mm,
  enlarge right by=0mm,
  #1
}

\newif\ifshownotes
\shownotestrue 

\ifdefined\MakeWithNotes
  \shownotestrue
\fi
\ifdefined\MakeWithoutNotes
  \shownotesfalse
\fi

\ifshownotes
  \newcommand{\colornote}[3]{{\color{#1}\bf{#2: #3}\normalfont}}
  \newcommand{\colornoteTwo}[3]{{\color{#1}\bf{#3}\normalfont}}
  \newcommand{\colornoteThree}[2]{{\color{#1}\bf{#2}\normalfont}}      
\else
  \newcommand{\colornote}[3]{}
  \newcommand{\colornoteTwo}[3]{}
  \newcommand{\colornoteThree}[2]{}      
\fi

\definecolor{darkgreen}{rgb}{0.0,0.65,0}

\usepackage{graphicx}
\usepackage{array}
\usepackage{multirow}
\usepackage{booktabs}
\usepackage{makecell}
\usepackage{placeins}

\usepackage{graphicx}

\newtcolorbox{GrayBox}[1][]{
  enhanced, breakable, sharp corners,
  boxrule=0pt,
  colback=gray!10!white, colframe=gray!10!white,
  borderline west={2pt}{0pt}{gray!75!black},
  width=\linewidth,
  enlarge left by=0mm,
  enlarge right by=0mm,
  #1
} 

\definecolor{cvprblue}{rgb}{0.21,0.49,0.74}
\usepackage[pagebackref,breaklinks,colorlinks,allcolors=cvprblue]{hyperref}

\def\paperID{9082}
\def\confName{CVPR}
\def\confYear{2026}

\title{\textit{LouvreSAE}: Sparse Autoencoders for Interpretable \\ and Controllable Style Transfer}

\author{%
Raina Panda$^{1}$
\hspace{3pt}
Daniel Fein$^{2}$
\hspace{3pt}
Arpita Singhal$^{2}$
\hspace{3pt}
Mark Fiore$^{2}$
\hspace{3pt}
Maneesh Agrawala$^{2}$
\hspace{3pt}
Matyas Bohacek$^{2}$
\vspace{12pt}\\
$^1$Washington High School
\hspace{5pt}
$^2$Stanford University
\\
}

\begin{document}
\twocolumn[{%
\renewcommand\twocolumn[1][]{#1}%
\maketitle
\vspace{-10pt}
\centering

\includegraphics[width=\linewidth]{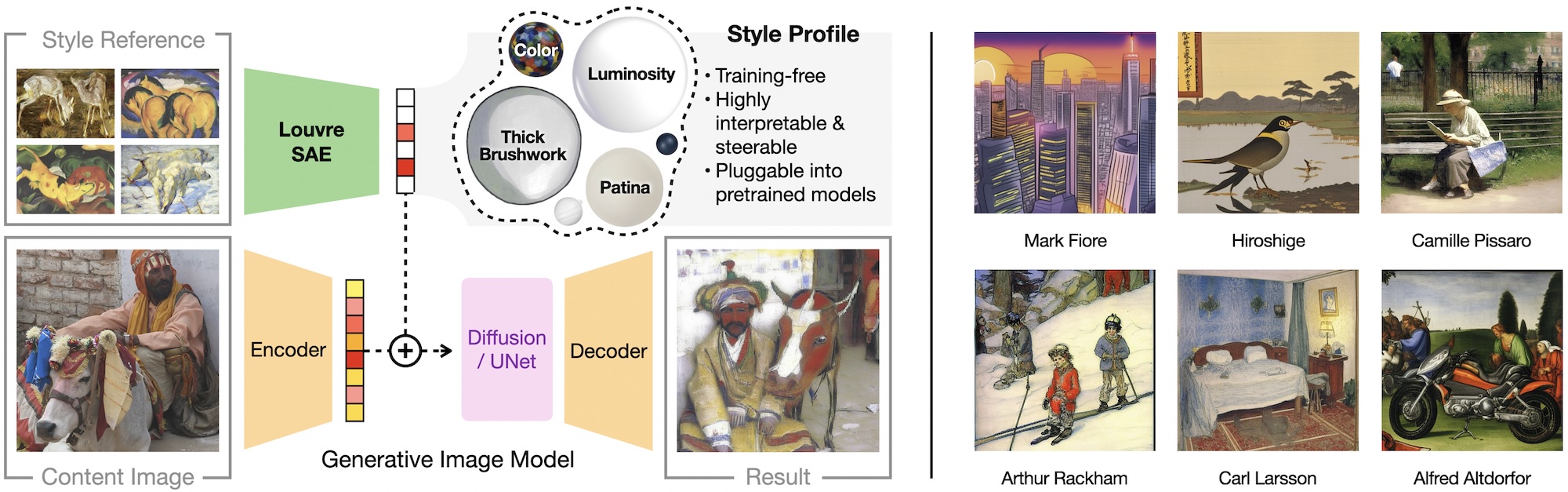}

\captionof{figure}{\textbf{Overview of \textit{LouvreSAE}.} Our method performs style transfer using an art-specific sparse autoencoder (SAE) on top of latent embeddings of a pre-trained image-to-image (I2I) models, as shown on the left. The SAE enables us to construct style profiles, which contain interpretable concepts and intensities that can be inspected and adjusted. Shown on the right are qualitative, representative examples of images generated with our approach, where a photograph (content image) was turned into styles of six prominent artists.\vspace{1em}}

\label{fig:teaser}
}]

\begin{abstract}

\noindent Artistic style transfer in generative models remains a significant challenge, as existing methods often introduce style only via model fine-tuning, additional adapters, or prompt engineering, all of which can be computationally expensive and may still entangle style with subject matter. In this paper, we introduce a training- and inference-light, interpretable method for representing and transferring artistic style. Our approach leverages an art-specific Sparse Autoencoder (SAE) on top of latent embeddings of generative image models. Trained on artistic data, our SAE learns an emergent, largely disentangled set of stylistic and compositional concepts, corresponding to style-related elements pertaining brushwork, texture, and color palette, as well as semantic and structural concepts. We call it \textit{LouvreSAE}\footnote{Similar to louvres (also spelled 'louvers'), which steer light to control shadows, our SAE steers how activations flow through a generative AI model in order to control style. The name is also a nod to \textit{the} Louvre.} and use it to construct style profiles: compact, decomposable steering vectors that enable style transfer without any model updates or optimization. Unlike prior concept-based style transfer methods, our method requires no fine-tuning, no LoRA training, and no additional inference passes, enabling direct steering of artistic styles from only a few reference images. We validate our method on ArtBench10, achieving or surpassing existing methods on style evaluations (VGG Style Loss and CLIP Score Style) while being 1.7-20x faster and, critically, interpretable.
\end{abstract}


\vspace{-1em}
\begin{GrayBox}
\textbf{\small{Correspondence}} \hfill \href{mailto:maty@stanford.edu}{\texttt{\small{maty@stanford.edu}}} \\[2pt]
\textbf{\small{Project Page (Code \& Data)}} \hfill \href{https://louvresae.github.io}{\texttt{\small{louvresae.github.io}}}
\end{GrayBox}
\section{Introduction}
\label{sec:intro}

Since the early successes of AI image generation, there has been a growing interest in adapting these models to specific artistic styles~\cite{gatys2015neural,ruiz2023dreambooth,hu2022lora}. Such capabilities promise to help creative professionals meaningfully enhance and augment their work~\cite{zhou2024generative,ko2023large}, enable consistent visual generation~\cite{pascual2024enhancing,akdemir2024oracle}, and support a range of downstream use cases~\cite{johnson2016perceptual}. Despite considerable effort devoted to this goal, to date, existing methods have received mixed reception and seen limited adoption~\cite{inie2023designing}.

This is due to several challenges. (a) The style transfer literature often lacks a precise, \emph{operational} definition of style, relying instead on loose descriptions that cannot be algorithmically captured or reliably disentangled from content~\cite{ioannou2024evaluation,wang2017multimodal,lang1987concept}. (b) Style representations in current models are frequently opaque; this lack of interpretability makes it difficult to align the model's encoding of style with artists' intent or established art historical understanding~\cite{wu2023uncovering,wang2023stylediffusion}. (c) Many existing style transfer methods require lengthy and unstable fine-tuning procedures~\cite{gupta2017characterizing,shen2018neural}. Moreover, while many approaches explicitly aim to support artistic styles and serve artists, few have engaged practicing creative professionals in the research process itself~\cite{inie2023designing,holmer2015constructing}.

In this paper, we aim to address these limitations, focusing on representing and transferring the artistic styles of individual artists. We build on the intuition that style constitutes the characteristic visual elements that consistently appear across an artist’s body of work~\cite{finn2024learning}. To move beyond this intuition and address challenge (a), we propose a precise, operational definition grounded in the representations learned by generative models. 

\begin{bluebox}[]
\paragraph{\textcolor{indigo}{Operational Definition of Style.}} We start from the established premise that the latent embeddings and internal activations of deep vision models effectively capture these stylistic properties, encoding both local textures and global structures~\cite{gatys2016image, radford2021learning}, to the point they can be reconstructed back into image domain at high quality. Consequently, we argue that an artist's unique style must correspond to specific, recurring patterns within these high-dimensional representations. We define style operationally as the shared component of the latent representation that persists across an artist's body of work, even as the semantic content varies. Notably, for this operational definition to work, the reference set must be semantically diverse.
\end{bluebox}

However, operationalizing this definition is challenging because model representations are typically dense and entangled; models often compress information through superposition, making features difficult to interpret or isolate directly~\cite{elhage2022toy}. We therefore employ Sparse Autoencoders (SAEs)~\cite{cunningham2023sparse,bricken2023towards} as the essential tool to access these patterns. SAEs decompose dense activations into a sparser, more disentangled, and interpretable set of features. By applying SAEs and analyzing activations across a semantically diverse set of images, we can identify the specific features that consistently activate. This process effectively collapses the content variance, allowing the recurring features (the style) to be isolated and utilized.

To implement this definition, we train SAEs specifically on artistic data to learn an emergent, disentangled set of concepts. To make these concepts interpretable and align them with human understanding---addressing challenge (b)---we introduce a novel taxonomy of artistic concepts. This taxonomy was constructed by systematically integrating established art historical literature with autointerpretability techniques applied to our SAEs, providing a structured vocabulary for analyzing and manipulating style.

We use this framework to build compact style profiles: steering vectors for individual artists or art movements from just a handful of reference images. Because these profiles are composed of features categorized by our taxonomy, they offer a degree of transparency and control that we hypothesize will better align with artists' preferences than opaque methods. This approach enables computationally efficient and controllable image generation—directly addressing challenge (c).

The contributions of this paper are three-fold, with code available at \url{https://louvresae.github.io}:

\begin{enumerate} 

    \item \textbf{An Operational Definition and Taxonomy of Style.} We propose an operational definition of style based on consistent feature activations in the latent space across diverse semantic content. To this end, we develop and release a set of art-specific SAEs. Alongside these SAEs, we introduce a taxonomy of artistic concepts derived in line with art scholarships and state-of-the-art autointerpretability methods, providing a structured framework for fine-grained analysis and interpretation. 
    
    \item \textbf{Interpretable Style Representation.} We use our framework to build style profiles, which are interpretable and decomposable. This means users of the style transfer method can intervene and precisely control the model's representation of the style at hand. 
    
    \item \textbf{A Lightweight, Controllable Style Transfer Method.} By identifying, filtering, and steering SAE concepts, we introduce a novel method for style transfer. Grounded in our operational definition of style, this approach is 1.7-20x faster than existing methods while still outperforming them on ArtBench10.

\end{enumerate}

\section{Related Work}
\label{sec:rel_work}

\subsection{Style Transfer}

Recent neural style transfer methods leverage diffusion and attention to overcome limitations of previous convolutional neural network-based approaches. DiffuseST~\cite{hu2024diffusest} injects style and content at separate diffusion steps using BLIP-2 and spatial DDIM features. Style Injection in Diffusion~\cite{chung2024style} modifies self-attention by swapping style key–value pairs, with query preservation and temperature scaling. InST~\cite{zhang2023inversion} uses attention-based textual inversion and cross-attention to learn style from a single painting.

Transformer-based approaches offer stronger long-range style reasoning. StyTr2~\cite{deng2022stytr2} uses dual encoders and content-aware positional encoding; StyleFormer~\cite{park2022styleformer} enables real-time arbitrary transfer through parametric style composition; and Dynamic Style Kernels~\cite{xu2023learning} generates per-pixel adaptive kernels via style alignment and gating.

Region-aware attention is also on the rise. STRAT~\cite{qi2025strat} fuses CNN and transformer features with SNR-guided selective attention, inspiring 3D work like Style-NeRF2NeRF~\cite{fujiwara2024style}, which ensures view-consistent, text-driven stylization in neural radiance fields.

Recent work improves content-style disentanglement. Methods like UnZipLoRA~\cite{liu2025unziplora} and CSD-VAR~\cite{nguyen2025csdvar} explicitly decompose images, while SCFlow~\cite{ma2025scflow} learns disentanglement implicitly with flow models. To prevent artifacts, StyleKeeper~\cite{jeong2025stylekeeper} uses negative visual query guidance (NVQG) to block content leakage, and Balanced Image Stylization~\cite{jiang2025balanced} employs a Style Matching Score (SMS) to harmonize the transfer. Separately, IntroStyle~\cite{kumar2025introstyle} offers style attribution using the statistical fingerprint of some diffusion features.

\begin{figure*}
    \centering
    \includegraphics[width=1.0\linewidth]{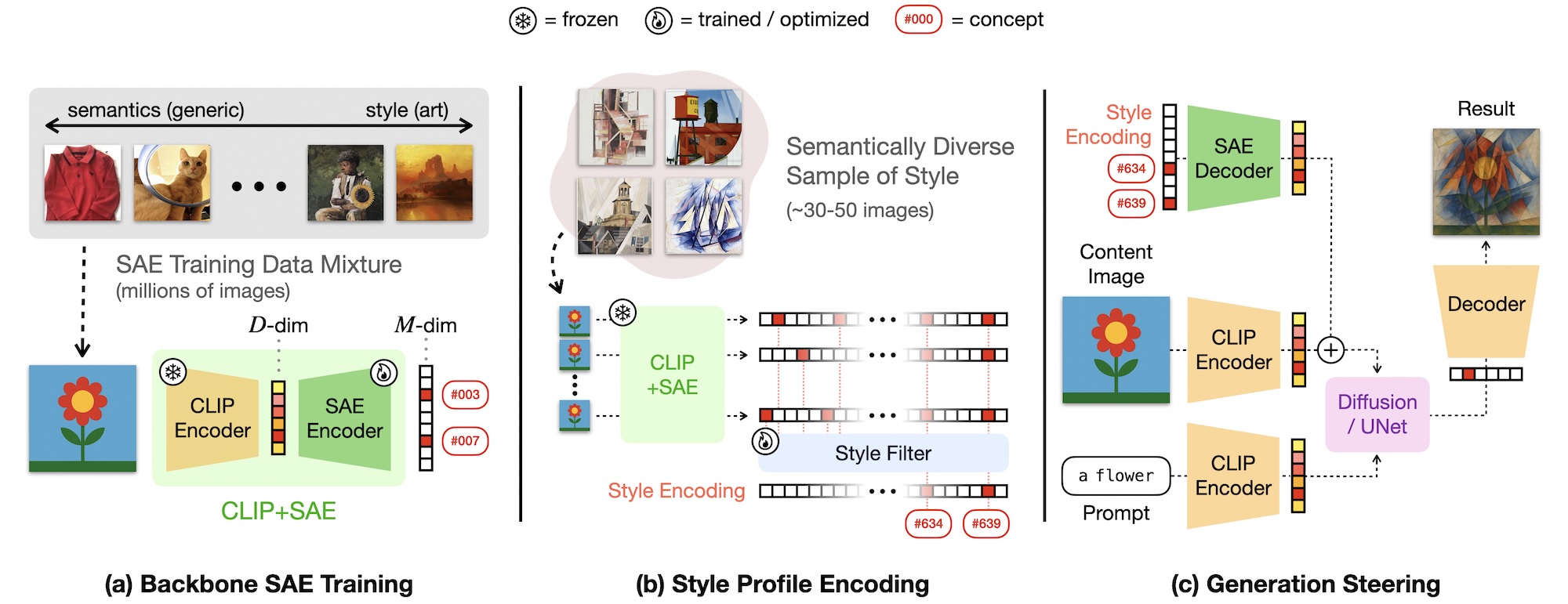}
    \caption{\textbf{\textit{LouvreSAE} Training \& Inference Stages.} The method comprises three stages: \textbf{(a) Backbone SAE Training}, where the SAE is presented with a wide range of data sources (including both generic and art-specific datasets); \textbf{(b) Style Profile Encoding}, where a handful of sparse codes obtained from samples of the target style are overlapped to find overlapping concepts that make up the style; and \textbf{(c) Generation Steering}, where the style profile us used as a steering vector during image-to-image pipeline inference.}
    \label{fig:method_overview}
\end{figure*}

\subsection{Style Transfer Datasets \& Benchmarks}

Benchmarks like WikiArt~\cite{tan2018improved}, BAM~\cite{wilber2017bam}, and ArtBench-10~\cite{liao2022artbench} support style classification and transfer evaluation, offering 60K–80K+ artworks across genres, styles, and media. Recent datasets address style diversity and labeling quality: StyleBench~\cite{guo2025stylebench} provides 490 expert-curated samples across 73 styles; OmniStyle-1M~\cite{wang2025omnistyle} includes 1M triplets spanning 1K style categories; and LAPIS~\cite{maerten2025lapis} offers 11,723 images with hierarchical expert annotations. The Style Transfer Dataset~\cite{kitov2024style} augments these with 10K human-rated stylizations to predict perceptual preference and optimal content–style match.

\subsection{Sparse Autoencoders}

Sparse autoencoders (SAEs) aim to isolate dense neural network features by reconstructing activations using a small, sparse subset of learned units that are more interpretable. This helps address polysemanticity~\cite{cunningham2023sparse} (neurons encoding unrelated concepts) and superposition~\cite{rajamanoharan2024improving} (shared representation of distinct concepts).

Prominent SAE variants include JumpReLU~\cite{rajamanoharan2024jumping} for improved reconstruction via discontinuous activations; BatchTopK~\cite{bussmann2024batchtopk} for global batch-level sparsity; Group-SAEs~\cite{ghilardi2024efficient} for layer clustering; Archetypal SAEs~\cite{fel2025archetypal} using convex hull constraints; and Matryoshka SAEs~\cite{bussmann2025learning} with nested hierarchical dictionaries.

Though originally developed for language models, SAEs are increasingly used for interpretability and control in computer vision. Concept Steerers~\cite{kim2025concept} apply k-sparse SAEs to guide diffusion outputs; TIDE~\cite{huang2025tide} adapts them for temporal-aware diffusion transformers.

Several works train SAEs on CLIP embeddings, including PatchSAE~\cite{lim2024sparse}, which extracts localized, interpretable patch-level features from CLIP ViTs~\cite{dosovitskiy2020image}. Some SAEs have also been trained for UNets~\cite{ronneberger2015u} in generative models, such as SDXL-Unbox~\cite{surkov2025unpacking,podell2023sdxl}. These methods demonstrated successful feature disentanglement in the majority of their respective SAE dictionaries, with 10–15\% of features found to be highly steerable~\cite{joseph2025steering}.

\section{\textit{LouvreSAE} Steering Method}
\label{sec:methods}

Our method, which we call \textit{LouvreSAE} Steering, comprises three stages: (1) training of the backbone SAE; (2) encoding of the style profile; and (3) steering of the image generation. These are illustrated in Figure~\ref{fig:method_overview}. The method operates on top of a pre-trained I2I model of arbitrary architecture, provided it utilizes a CLIP-like prior.

\subsection{Backbone SAE Training}
\label{subsec:backbone_sae}

As shown in Figure~\ref{fig:method_overview}(a), we train a Sparse Autoencoder (SAE) on the representations extracted from a pre-trained, frozen vision encoder. We utilize the image embeddings from a CLIP-like encoder as it provides a rich representation, which is already able to preserve artistic styles and is used in many generative models, allowing for direct, out-of-the-box conditioning of various models and architectures.

\textbf{Data Considerations.} By incorporating a variety of generic and art-specific data sources (emphasizing the latter), we expect the SAE to be better equipped to surface semantic and stylistic concepts at a finer granularity than conventional SAEs trained exclusively on generic datasets such as LAION-5B~\cite{schuhmann2022laion} or ImageNet~\cite{deng2009imagenet}. We note, however, that style elements can be expected to entangle with layout and semantic concepts, as a result of the training data\footnote{For example, impressionist brushwork appears mostly in images depicting landscapes, gardens, and parks, and thus the two co-occur significantly. One can therefore reasonably expect them to be highly correlated in the model’s activation geometry, and, by extension, in the SAE’s learned concepts.}.  

\textbf{Architecture.} We adopt the BatchTopK SAE architecture~\cite{bussmann2024batchtopk}. Given an input dimension $D$ (the CLIP embedding dimension), the SAE learns an overcomplete dictionary of $M$ concepts, where $M \gg D$. The encoder maps an input embedding $x \in \mathbb{R}^{D}$ to a sparse activation vector $z \in \mathbb{R}^{M}$, and the decoder reconstructs the input $\hat{x}$ from $z$.

\textbf{Training Objective.} The training objective minimizes the reconstruction error under a sparsity constraint. We employ the Mean Squared Error (MSE) loss combined with an auxiliary reanimation loss term designed to mitigate "dead features" (features that never activate across a batch):

\begin{equation}
\mathcal{L}(x, \hat{x}, z_{pre}) = \|x - \hat{x}\|^2_2 - \lambda \cdot \text{mean}(z_{pre} \cdot I_{dead})
\end{equation}

\noindent where $z_{pre}$ are the pre-activation codes, $I_{dead}$ is a binary indicator for inactive features, and $\lambda$ is the reanimation coefficient. 

Sparsity is enforced globally using the BatchTopK mechanism, which retains only the top $K_{batch}$ activations across the entire batch during the forward pass. $K_{batch}$ is defined as $k \times B$, where $k$ is the target average per-sample sparsity and $B$ is the batch size. This global constraint encourages better utilization of the dictionary compared to per-sample L1 regularization.

\textbf{Optimization.} The model is optimized using Adam~\cite{adam2014method} for $E$ epochs, a learning rate of $\eta$, and $W$ linear warmup steps. The key hyperparameters governing the learned representation are the dictionary size $M$ and sparsity level $k$.

\textbf{Autointerpretability.} Once the SAE is trained, we employ a set of analyses to understand what individual concepts represent, and classify them into a taxonomy, see Appendix~\ref{app:autointerp}.

\subsection{Style Profile Encoding}

Once the SAE is trained, we encode a target artistic style into a sparse vector of concept activations, termed the \textit{style profile}, illustrated in Figure~\ref{fig:method_overview}(b). This process identifies the subset of learned concepts that consistently appear across a corpus of representative artworks, consistent with the used definition of style~\cite{finn2024learning}, effectively filtering out incidental semantic content.

\textbf{Concept Collection.} We begin with a curated dataset of reference images $I_{S} = \{i_1, ..., i_N\}$ representative of the target style. We extract the CLIP embeddings for each image and process them through the trained SAE encoder to obtain a concept activation matrix $C \in \mathbb{R}^{N \times M}$.

\textbf{Characteristic Concept Identification.} To robustly disentangle style, we identify concepts that are characteristic by analyzing their consistency across the style sample, aligning with the used definition of style as recurring visual elements~\cite{finn2024learning}. We introduce a required overlap threshold $P$. A concept $j$ is selected if it is actived ($C_{i,j} > 0$) in at least $P$ share of the reference images.

\textbf{Style Profile Vector Construction.} The final style profile $V_{S} \in \mathbb{R}^{M}$ is a sparse vector. For the selected concepts, we compute the mean activation strength, considering only the images where the concept is active. The magnitude is then scaled by a strength multiplier $S$:

\begin{equation}
V_{S}[j] = S \cdot \frac{\sum_{i=1}^{N} C_{i,j} \cdot \mathbb{I}(C_{i,j} > 0)}{\sum_{i=1}^{N} \mathbb{I}(C_{i,j} > 0)}
\end{equation}

\noindent provided the activation frequency exceeds the threshold $P$; otherwise $V_{S}[j] = 0$. The parameters $P$ and $S$ control the breadth and intensity of the encoded style, respectively.

\begin{figure*}[t]
\centering
\renewcommand{\arraystretch}{1.05}
\setlength{\tabcolsep}{4pt}

\begin{tabular}{
    >{\centering\arraybackslash}m{3.5cm}
    >{\centering\arraybackslash}m{2.0cm}|
    >{\centering\arraybackslash}m{2.0cm}
    >{\centering\arraybackslash}m{2.0cm}
    >{\centering\arraybackslash}m{2.0cm}
    >{\centering\arraybackslash}m{2.0cm}
     >{\centering\arraybackslash}m{2.0cm}
}
\toprule
\multicolumn{2}{c|}{\textbf{Concept}} &
\multicolumn{4}{c}{\textbf{Representative Exemplars}} \\
\midrule

\raisebox{0pt}[0pt][0pt]{%
\parbox{3.6cm}{\centering
    Surreal Volumetric Curvature\\[3pt]
    \colorbox{gray!20}{\parbox{2.8cm}{\centering Artistic\\Aesthetic}}
}}
&
\includegraphics[width=2cm,height=2cm]{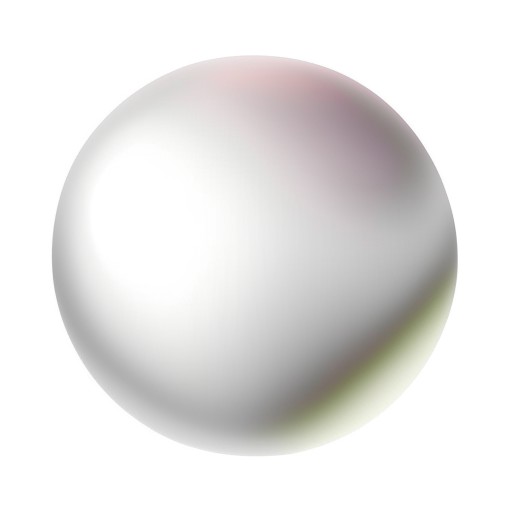}
&
\includegraphics[width=2cm,height=2cm]{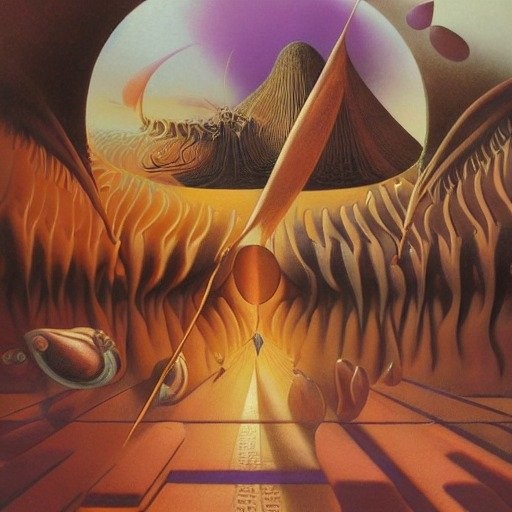}
&
\includegraphics[width=2cm,height=2cm]{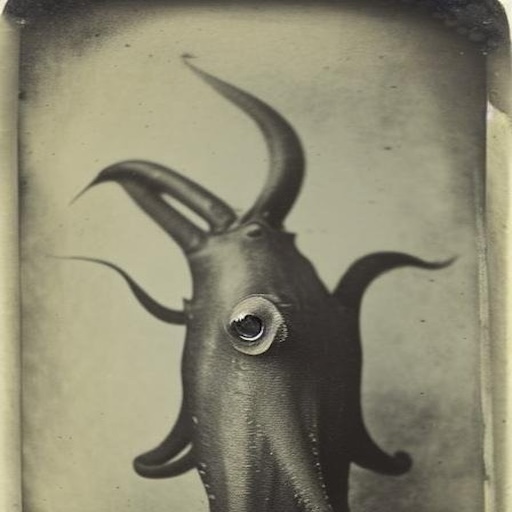}
&
\includegraphics[width=2cm,height=2cm]{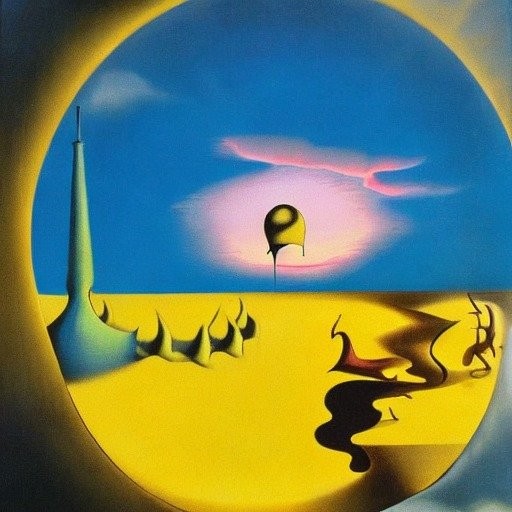}
&
\includegraphics[width=2cm,height=2cm]{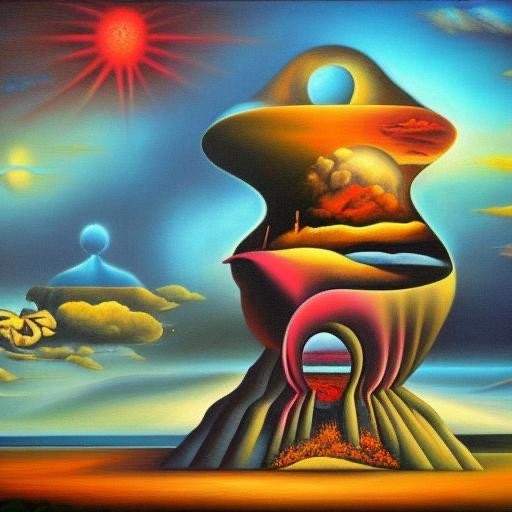}
&
\includegraphics[width=2cm,height=2cm]{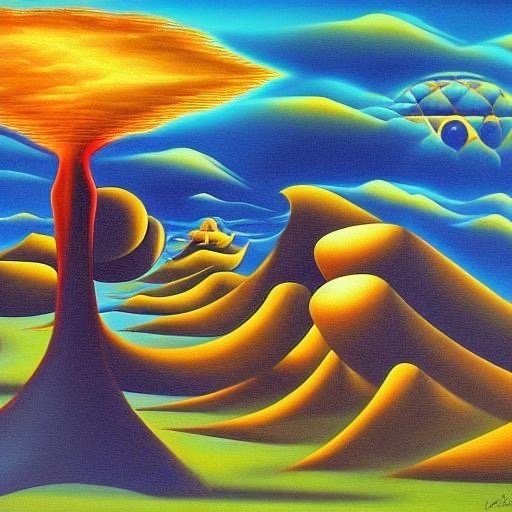}
\\

\raisebox{0pt}[0pt][0pt]{%
\parbox{3.6cm}{\centering
   Watercolor\\[3pt]
    \colorbox{gray!20}{\parbox{2.8cm}{\centering Artistic \\ Aesthetics}}
}}
&
\includegraphics[width=2cm,height=2cm]{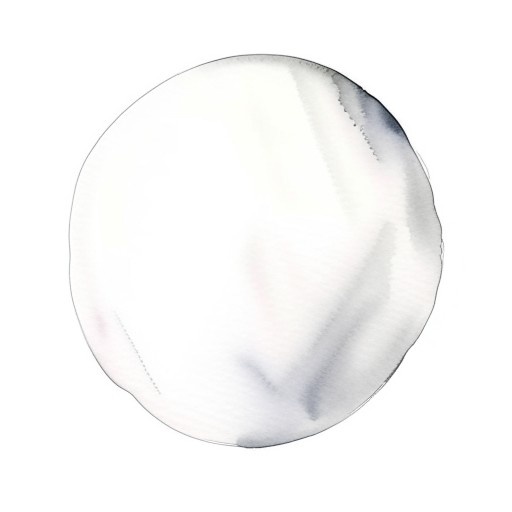}
&
\includegraphics[width=2cm,height=2cm]{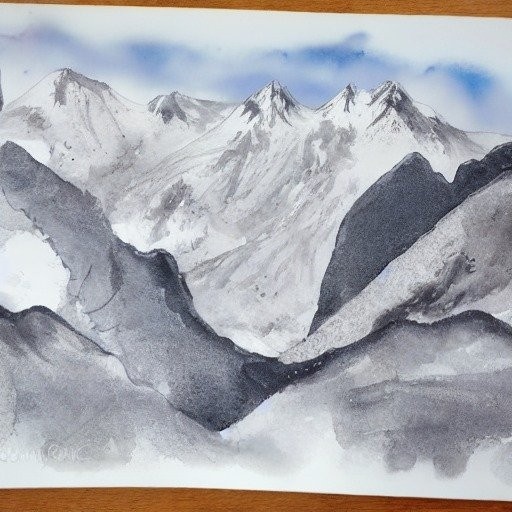}
&
\includegraphics[width=2cm,height=2cm]{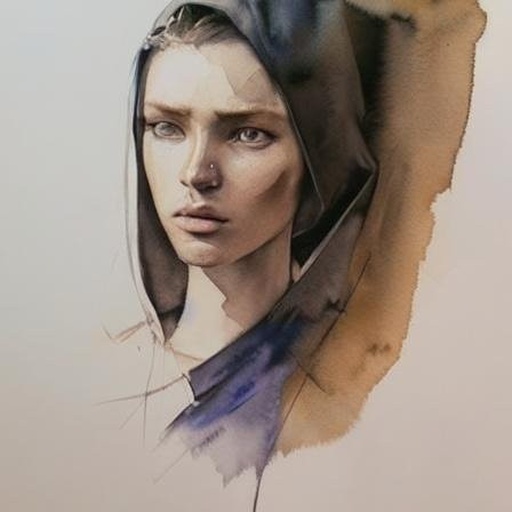}
&
\includegraphics[width=2cm,height=2cm]{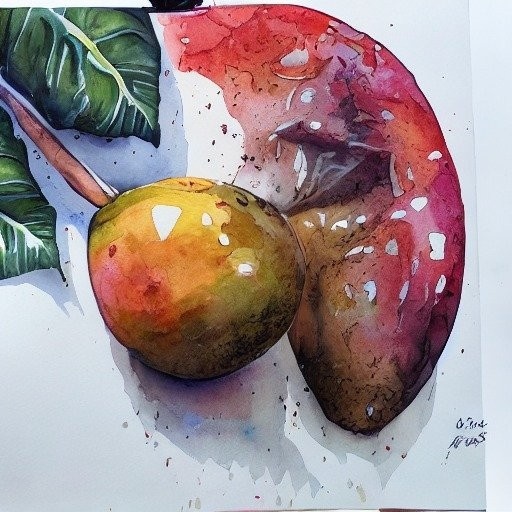}
&
\includegraphics[width=2cm,height=2cm]{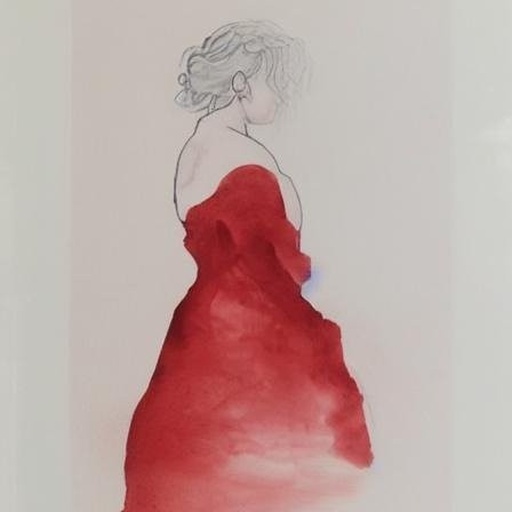}
&
\includegraphics[width=2cm,height=2cm]{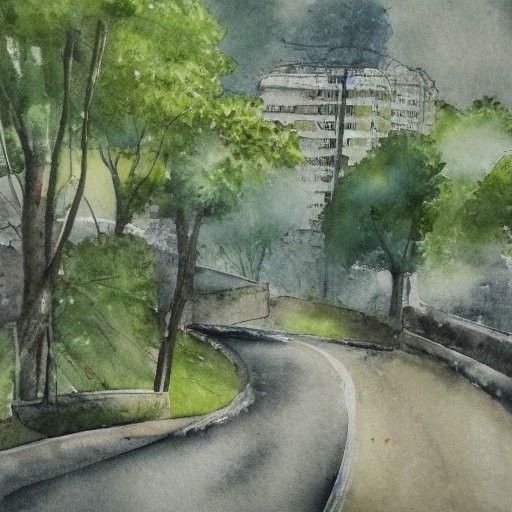}
\\

\bottomrule
\end{tabular}

\caption{
\textbf{Examples of SAE Concepts.} Shown above are the autointerpretability name and label, the prototype visualization, and four representative exemplars for two representative concepts learned by \textit{LouvreSAE}-20.}
\label{fig:concept_overview}
\end{figure*}

\begin{figure}[t]
\centering
\begin{tabular}{
    >{\centering\arraybackslash}m{2.7cm}|
    >{\centering\arraybackslash}m{2.15cm}
    >{\centering\arraybackslash}m{2.15cm}
}
\toprule
\shortstack{\textbf{Content Image}} &
\shortstack{\textbf{FeatureLab}} &
\shortstack{\textbf{\textit{LouvreSAE}}} \\
\midrule

\includegraphics[width=2.6cm,height=2.1cm]{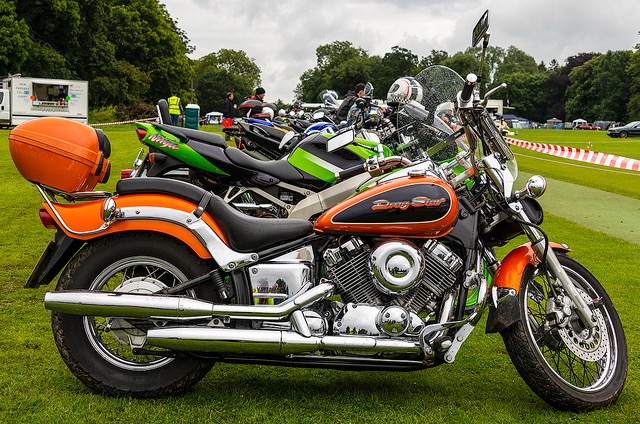} &
\includegraphics[width=2.1cm,height=2.1cm]{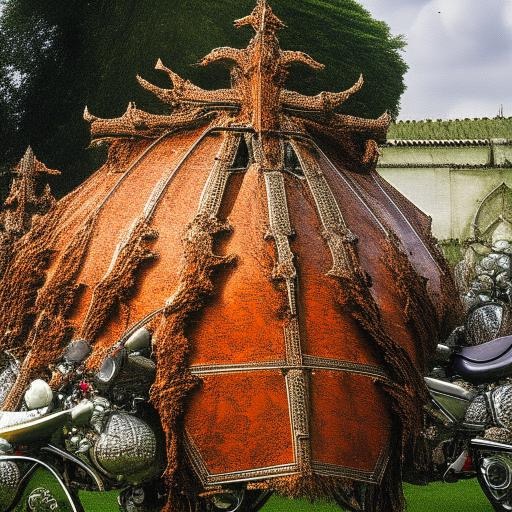} &
\includegraphics[width=2.1cm,height=2.1cm]{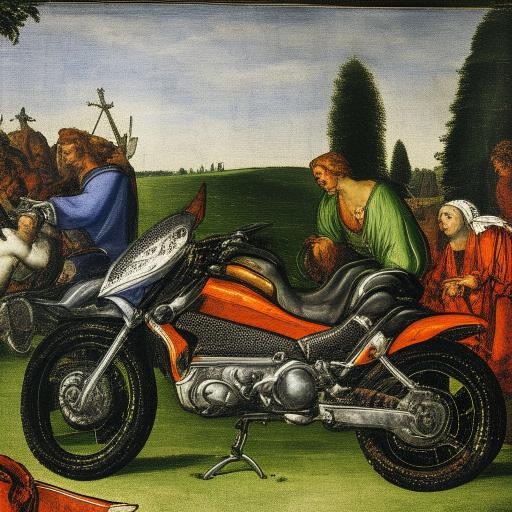}
\\[0.30cm]

\includegraphics[width=2.6cm,height=2.1cm]{fig/sae-comp/example-1/content.jpg} &
\includegraphics[width=2.1cm,height=2.1cm]{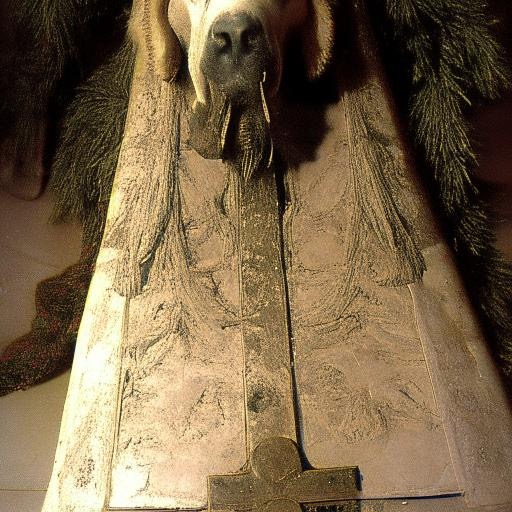} &
\includegraphics[width=2.1cm,height=2.1cm]{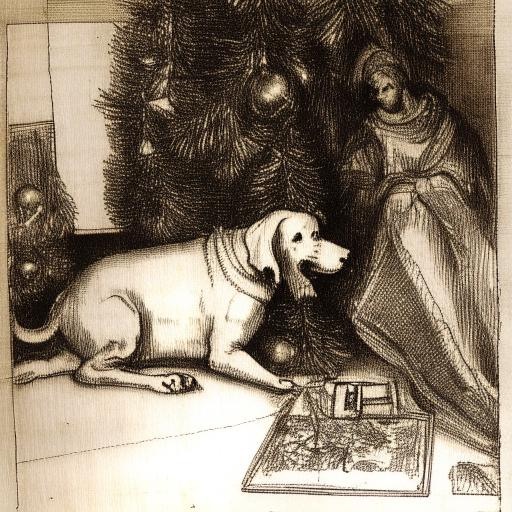}
\\[0.30cm]

\includegraphics[width=2.6cm,height=2.1cm]{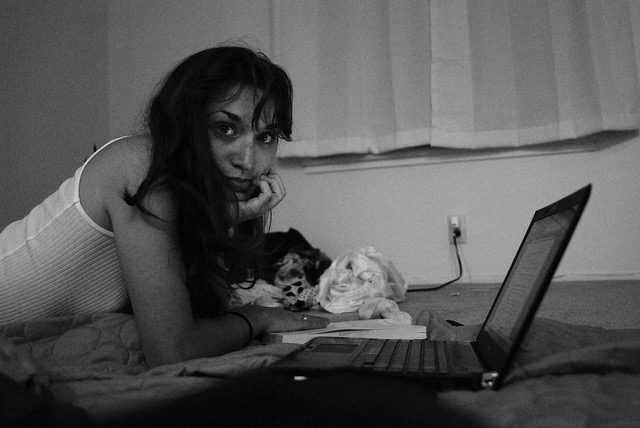} &
\includegraphics[width=2.1cm,height=2.1cm]{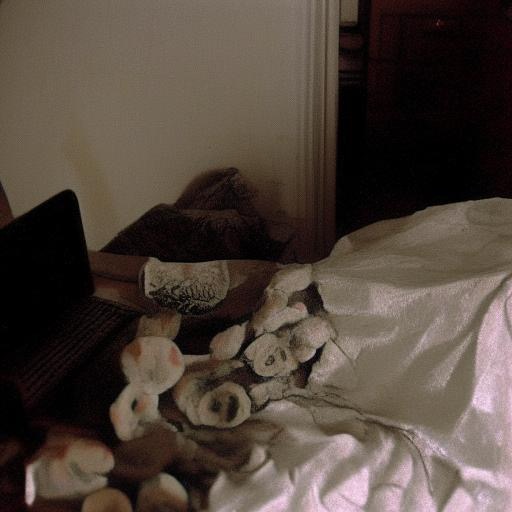} &
\includegraphics[width=2.1cm,height=2.1cm]{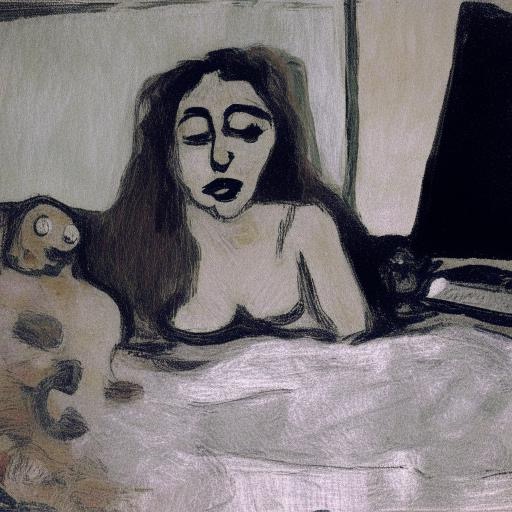}
\\



\bottomrule
\end{tabular}

\caption{\textbf{Qualitative Comparison of SAE Baselines.} We compare \textit{LouvreSAE}-20K with the FeatureLab SAE by swapping the backbone SAE module of the Kandinsky 2.2 pipeline with \textit{LouvreSAE} Steering. In each row, the content image is transferred to the style of a different artist: (from top to bottom) Alfred Altdorfor, Durer, Matisse, and Hiroshige.}
\label{fig:sae_comp}
\end{figure}

\subsection{Generation Steering}


The encoded style profile enables lightweight and targeted style transfer via direct intervention in the CLIP embedding space, requiring no fine-tuning of the base diffusion model, as shown in Figure~\ref{fig:method_overview}(c).

\textbf{Residual Decoding.} First, we calculate the style residual $\Delta_{S} \in \mathbb{R}^{D}$ by decoding the sparse style profile $V_{S}$ using the SAE's decoder $D_{SAE}(\cdot)$: $\Delta_{S} = D_{SAE}(V_{S})$.

This residual represents the vector shift in the embedding space necessary to introduce the stylistic elements captured by $V_{S}$.

\textbf{Latent Space Intervention.} We utilize a latent diffusion architecture comprising a prior (mapping inputs to the embedding space) and a Decoder. Given a content image $I_{C}$, we obtain its embedding $E_{C}$ using the Prior. Steering is performed via simple vector arithmetic by adding the style residual to the content embedding: $E_{steered} = E_{C} + \Delta_{S}$.

\textbf{Image Synthesis.} Finally, $E_{steered}$ is passed to the Decoder model to generate the stylized output image using a standard DDIM scheduler~\cite{song2020denoising} for $N_{steps}$ inference steps. This additive approach inherently supports the composition of multiple styles by linearly combining the residuals of different style profiles.
\section{Data}
\label{sec:data}

\subsection{Backbone SAE Training}

The training mixture for the backbone SAE was composed by randomly sampling from the following datasets\footnote{Some images had to be skipped as the URLs were no longer available. We also attempted deduplication based on URLs across these datasets.}: WikiArt~\cite{karayev2013recognizing} (81,444), LAION-5B~\cite{schuhmann2022laion} (1,500,000), LAION Art~\cite{schuhmann2022laion} (1,566,090), StyleBreeder~\cite{zheng2024stylebreeder} (1,126,400), and ImageNet~\cite{deng2009imagenet} (880,640). In total, this data mixture contained 5,154,574 images, ranging from generic imagery capturing semantic diversity to art- and style-specific imagery, in line with data considerations for this stage required for our operational style definition to work (per Section~\ref{subsec:backbone_sae}).

\subsection{Style Profile Encoding \& Generation Steering}

For style transfer experiments, we used a subset of artist samples in ArtBench10~\cite{akdemir2024oracle} that contained at least 50 images, devising an 80-20 split (40 train and 10 test images).

\begin{figure*}
    \centering
    \includegraphics[width=1.0\linewidth]{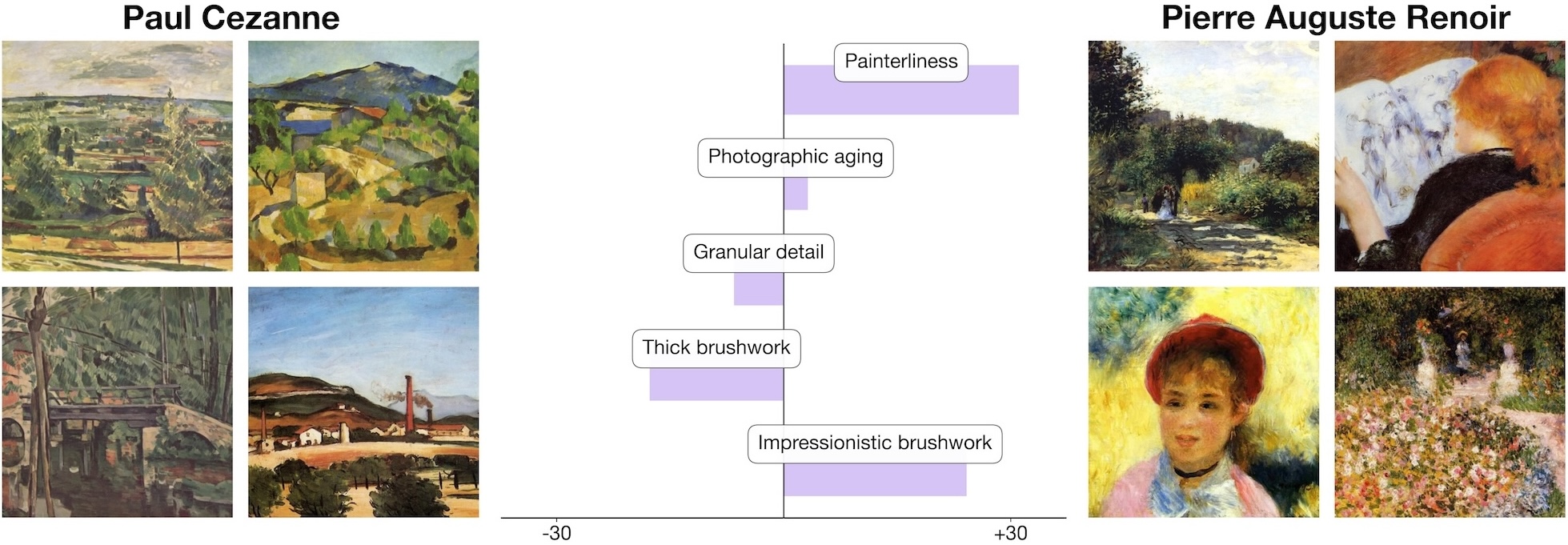}
    \caption{\textbf{Style Profile Comparison.} Paul Cézanne and Pierre-Auguste Renoir, two Impressionist artists with visually similar styles, nevertheless exhibit distinct and interpretable style profiles. Shown here is a subset of concepts they share, with the difference in concept intensity visualized. Additional unique concepts and further differences also exist beyond those displayed.}
    \label{fig:style_profile_diff}
\end{figure*}
\section{Experiments}
\label{sec:experiments}


While the \textit{LouvreSAE} steering method can operate on top of arbitrary I2I models (provided they utilize a CLIP-like prior), we opted for the pre-trained Kandinsky 2.2 model~\cite{razzhigaev2023kandinsky}, which allowed us to fairly compare our custom \textit{LouvreSAE} with the existing SAE baseline.

We trained four variants of \textit{LouvreSAE}: \textit{LouvreSAE}-10K, -20K, -40K, and -80K with 10,240, 20,480, 40,960, and 81,920 concepts, respectively. We found suitable hyperparameters for these models through a standard grid search. These models are open-sourced alongside autointerpretability and image steering code at \url{https://louvresae.github.io}.

Based on SAE-specific metrics (namely, $R^2$, $z$-sparsity / $L0$, and dictionary rank) and manual inspection of the learned concepts, we opted for \textit{LouvreSAE}-20K (see Appendix~\ref{app:louvresae_training}). The remainder of our style transfer experiments is Kandinsky 2.2 + \textit{LouvreSAE}-20K. See Appendix~\ref{app:louvresae_training} for complete configuration and implementation details.

\subsection{Baselines}

\paragraph{Backbone SAE.} We compare \textit{LouvreSAE}-20K to the FeatureLab SAE~\cite{daujotas2024interpreting} with 163,000 concepts (only 58,000 activate; 65\% are reported to be dead), trained on top of the same CLIP prior as \textit{LouvreSAE}-20K, but only on LAION-5B, not additional art- and style-specific data. See Appendix~\ref{app:sae_baseline_considerations} for more detailed SAE baseline considerations.

\paragraph{Style Transfer.} We compare Kandinsky 2.2 + \textit{LouvreSAE} against three state-of-the-art style transfer methods that also represent style at the input as a collection of artist images. These are namely B-LoRA~\cite{frenkel2024implicit}, InstantStyle~\cite{wang2024instantstyle}, and StyleShot~\cite{gao2025styleshot}. See Appendix~\ref{app:style_transfer_baseline_considerations} for more detailed style transfer baseline considerations.

\subsection{Metrics \& Evaluation}

\paragraph{Backbone SAE.} Direct quantitative comparison of SAEs remains limited, and so we compare the FeatureLab SAE baseline to our \textit{LouvreSAE}-20K directly in the style transfer pipeline. We present both a quantitative and a qualitative comparison.

\paragraph{Style Transfer.} We quantitatively compare different methods using five metrics for style and content quality: VGG Loss~\cite{gatys2015neural}: Style and Content, LPIPS~\cite{zhang2018unreasonable}, and CLIP Score~\cite{radford2021learning}: Style and Content (see Appendix~\ref{app:impl_metrics}).

\section{Results}
\label{sec:results}

\begin{table*}[t]
\centering
\renewcommand{\arraystretch}{1.2}
\setlength{\tabcolsep}{3pt}
\begin{tabular}{
    l l l | *{6}{>{\centering\arraybackslash}m{1.911cm}}
}
\toprule
\textbf{Method} & \textbf{I2I Backbone} & \textbf{SAE Backbone} & \textbf{VGG Loss: Style} $\downarrow$ & \textbf{VGG Loss: Content} $\downarrow$ & \textbf{LPIPS} $\downarrow$ & \textbf{CLIP Score: Style} $\uparrow$ & \textbf{CLIP Score: Content} $\uparrow$ \\
\midrule
B-LoRA        & SD XL~\cite{podell2023sdxl}         & ---                 & 1.63\ensuremath{\times}10\ensuremath{^{-4}} & 42.53 & 0.75 & 0.21 & 0.18 \\
InstantStyle  & SD XL~\cite{podell2023sdxl}         & ---                 & 2.48\ensuremath{\times}10\ensuremath{^{-5}}  & \textbf{22.94} & \textbf{0.64} & 0.25 & 0.24 \\
StyleShot     & SD 1.5~\cite{rombach2022high}        & ---                 & 3.74\ensuremath{\times}10\ensuremath{^{-5}}  & 29.17 & 0.71 & 0.25 & 0.20 \\
\midrule
\textit{L-SAE} Steer.     & Kan. 2.2~\cite{razzhigaev2023kandinsky}     & FeatLab SAE~\cite{daujotas2024interpreting}    & 2.18\ensuremath{\times}10\ensuremath{^{-5}} & 37.32 & 0.77 & 0.21 & \textbf{0.30} \\
\textit{L-SAE} Steer.     & Kan. 2.2~\cite{razzhigaev2023kandinsky}     & \textit{L-SAE} (Ours)                & \textbf{1.73\ensuremath{\times}10\ensuremath{^{-5}}} & 35.04 & 0.73 & \textbf{0.27} & 0.26 \\
\bottomrule
\end{tabular}
\caption{\textbf{Quantitative Comparison with Baselines.} Three state-of-the-art style transfer methods are compared to MosaicSAE on a combination of style- and content-based metrics.}
\label{tab:baselines}
\end{table*}

\begin{figure*}[t]
    \centering
    \begin{tabular}{
        >{\centering\arraybackslash}m{2.4cm}|
        >{\centering\arraybackslash}m{2.0cm}
        >{\centering\arraybackslash}m{2.0cm}|
        >{\centering\arraybackslash}m{2.0cm}
        >{\centering\arraybackslash}m{2.0cm}
        >{\centering\arraybackslash}m{2.0cm}|
        >{\centering\arraybackslash}m{2.0cm}
    }
    \toprule
    \shortstack{\textbf{Style}} &
    \shortstack{\textbf{Reference}\\\textbf{Sample}} &
    \shortstack{\textbf{Content}\\\textbf{Image}} &
    \shortstack{\textbf{B-LoRA}} &
    \shortstack{\textbf{InstantStyle}} &
    \shortstack{\textbf{StyleShot}} &
    \shortstack{\textbf{\textit{LouvreSAE}}} \\
    \midrule
    
    \shortstack{ Arthur \\ Rackham} &
    \includegraphics[width=2cm,height=2cm]{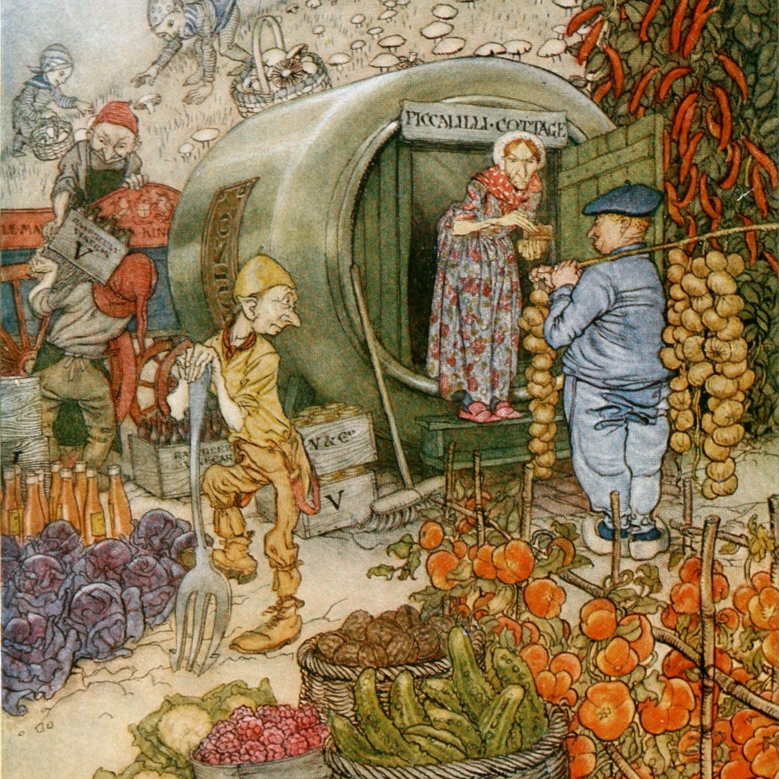} &
    \includegraphics[width=2cm,height=2cm]{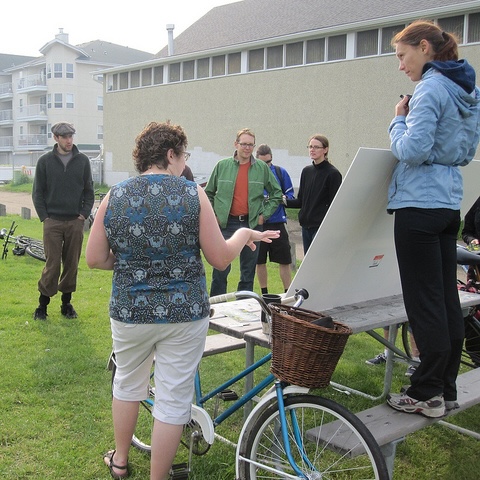} &
    \includegraphics[width=2cm,height=2cm]{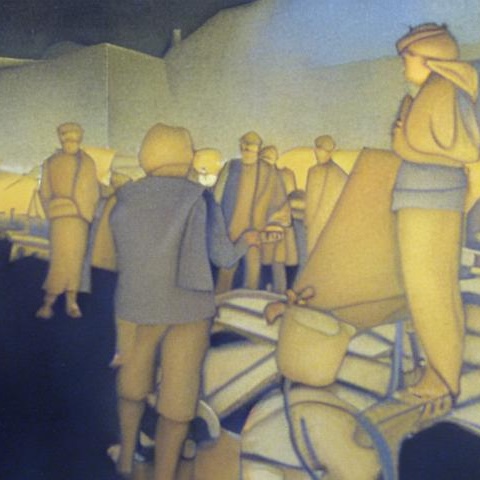} &
    \includegraphics[width=2cm,height=2cm]{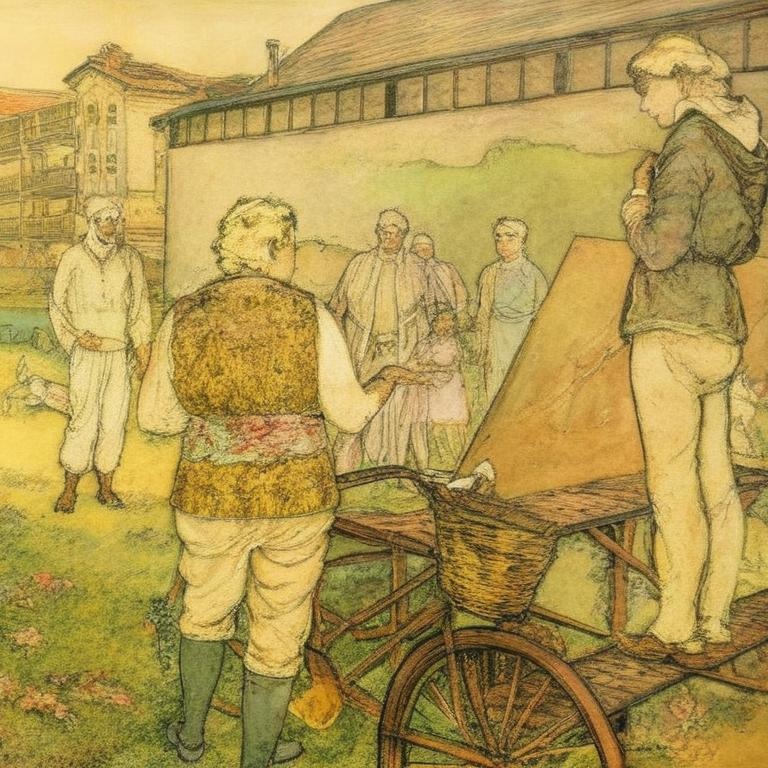} &
    \includegraphics[width=2cm,height=2cm]{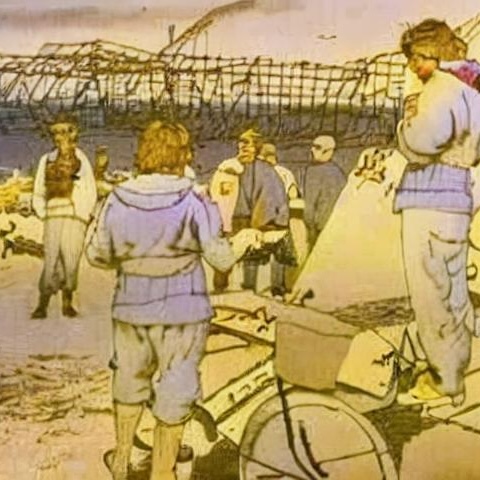} &
    \includegraphics[width=2cm,height=2cm]{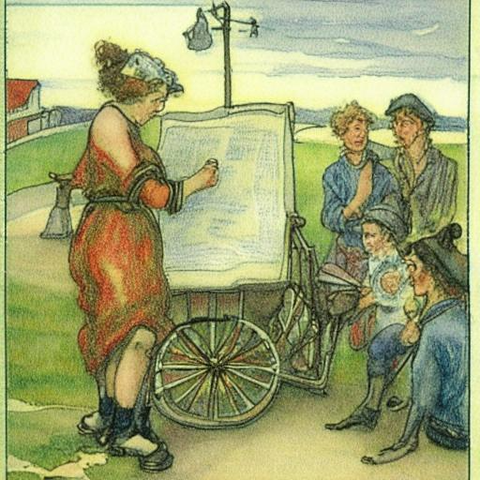} \\[0.4cm]
    
    \multirow{1}{*}{\centering Franz Marc} &
    \includegraphics[width=2cm,height=2cm]{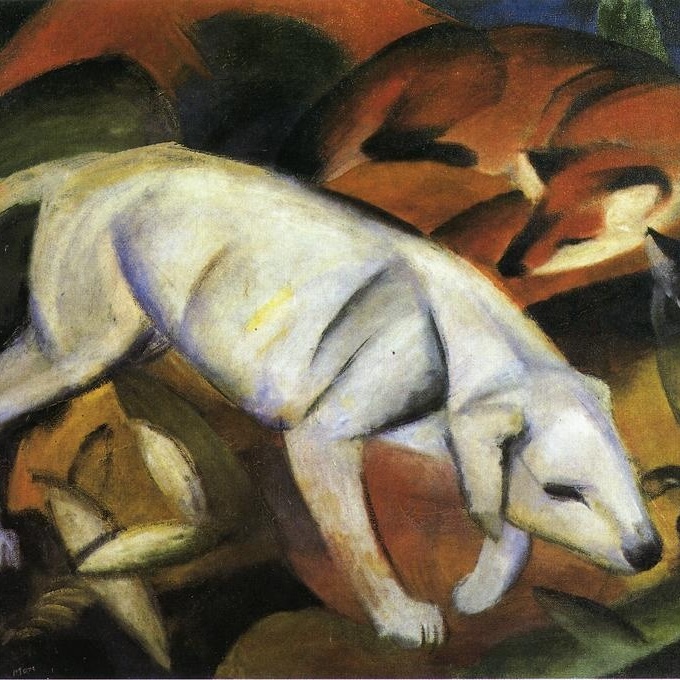} &
    \includegraphics[width=2cm,height=2cm]{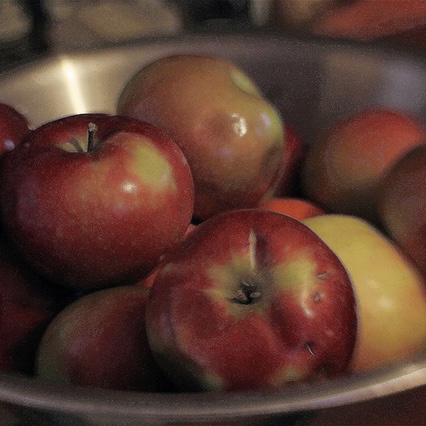} &
    \includegraphics[width=2cm,height=2cm]{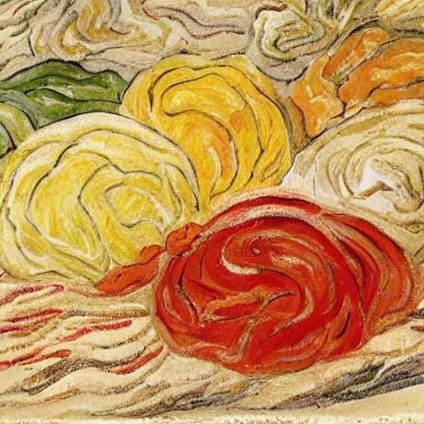} &
    \includegraphics[width=2cm,height=2cm]{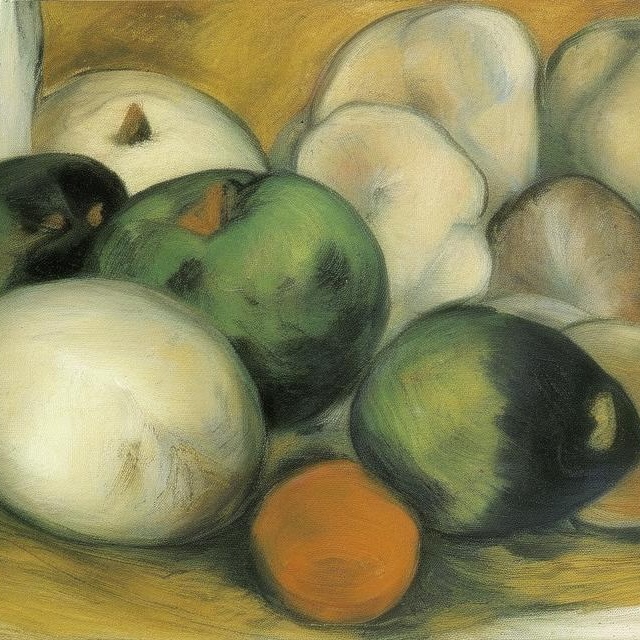} &
    \includegraphics[width=2cm,height=2cm]{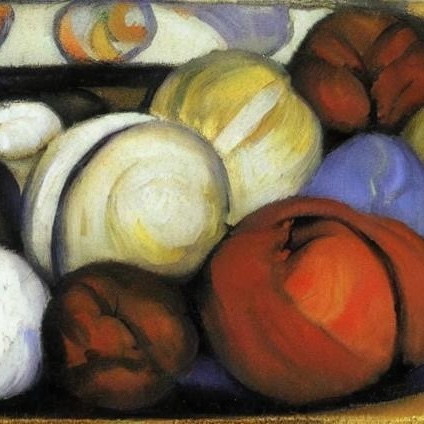} &
    \includegraphics[width=2cm,height=2cm]{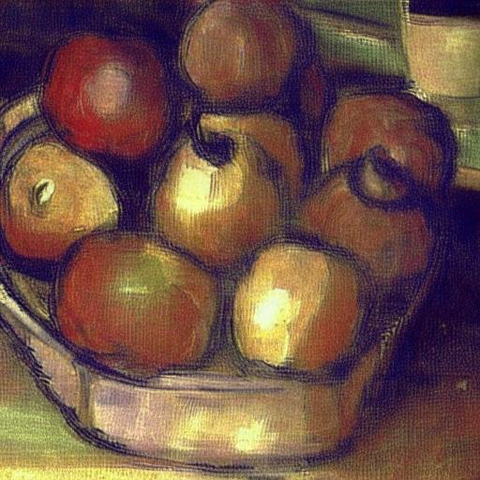} \\[0.4cm]

    \multirow{1}{*}{\centering Hiroshige} &
    \includegraphics[width=2cm,height=2cm]{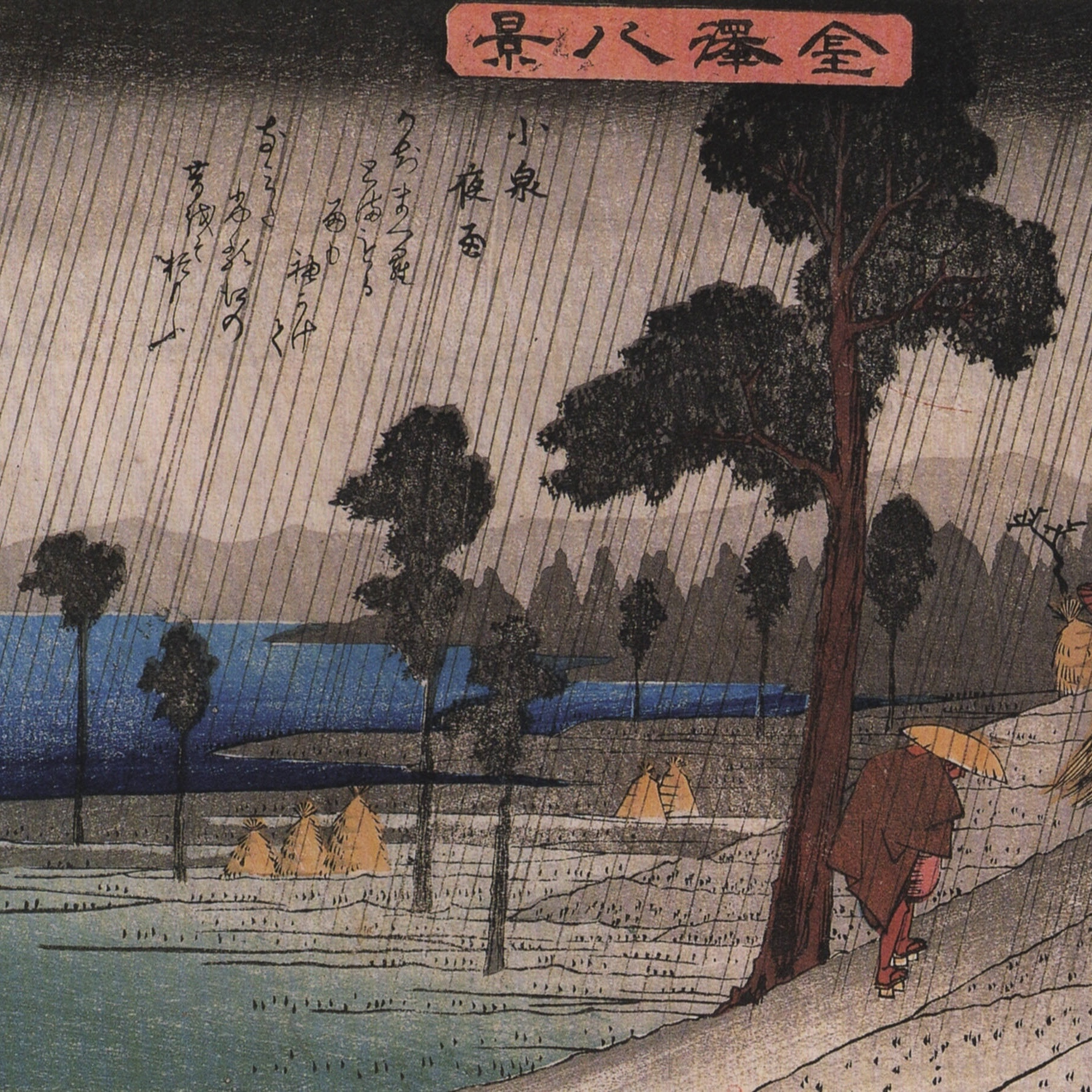} &
    \includegraphics[width=2cm,height=2cm]{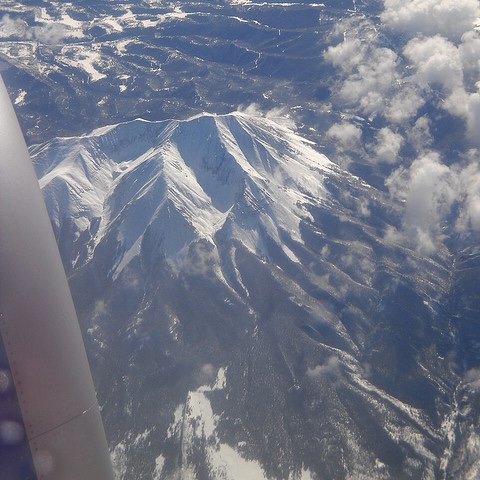} &
    \includegraphics[width=2cm,height=2cm]{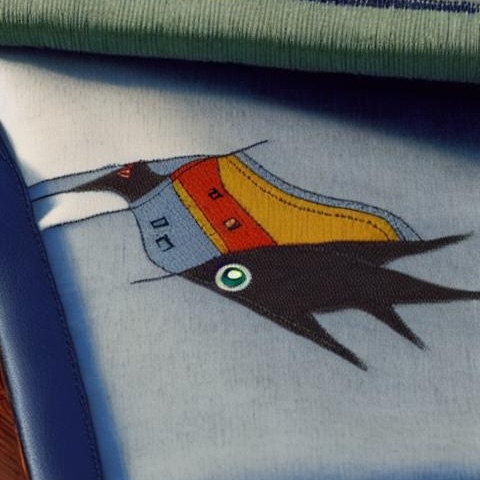} &
    \includegraphics[width=2cm,height=2cm]{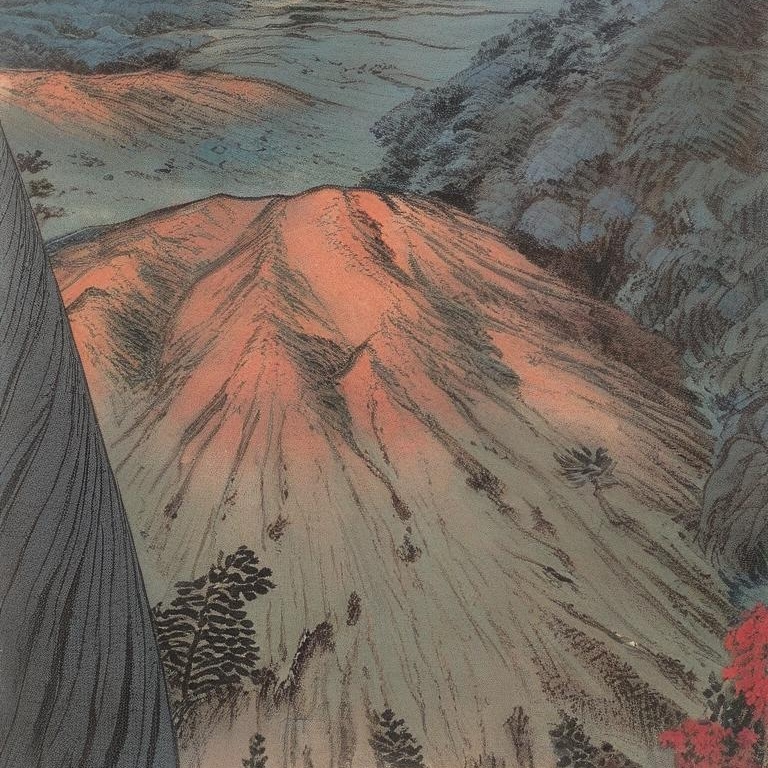} &
    \includegraphics[width=2cm,height=2cm]{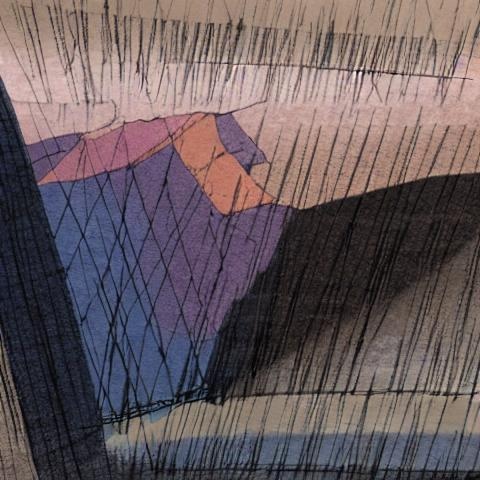} &
    \includegraphics[width=2cm,height=2cm]{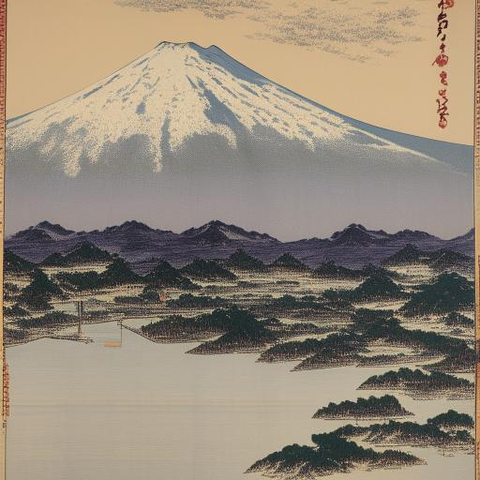} \\[0.4cm]

    \shortstack{ Theodor \\ Severin \\ Kittelsen } &
    \includegraphics[width=2cm,height=2cm]{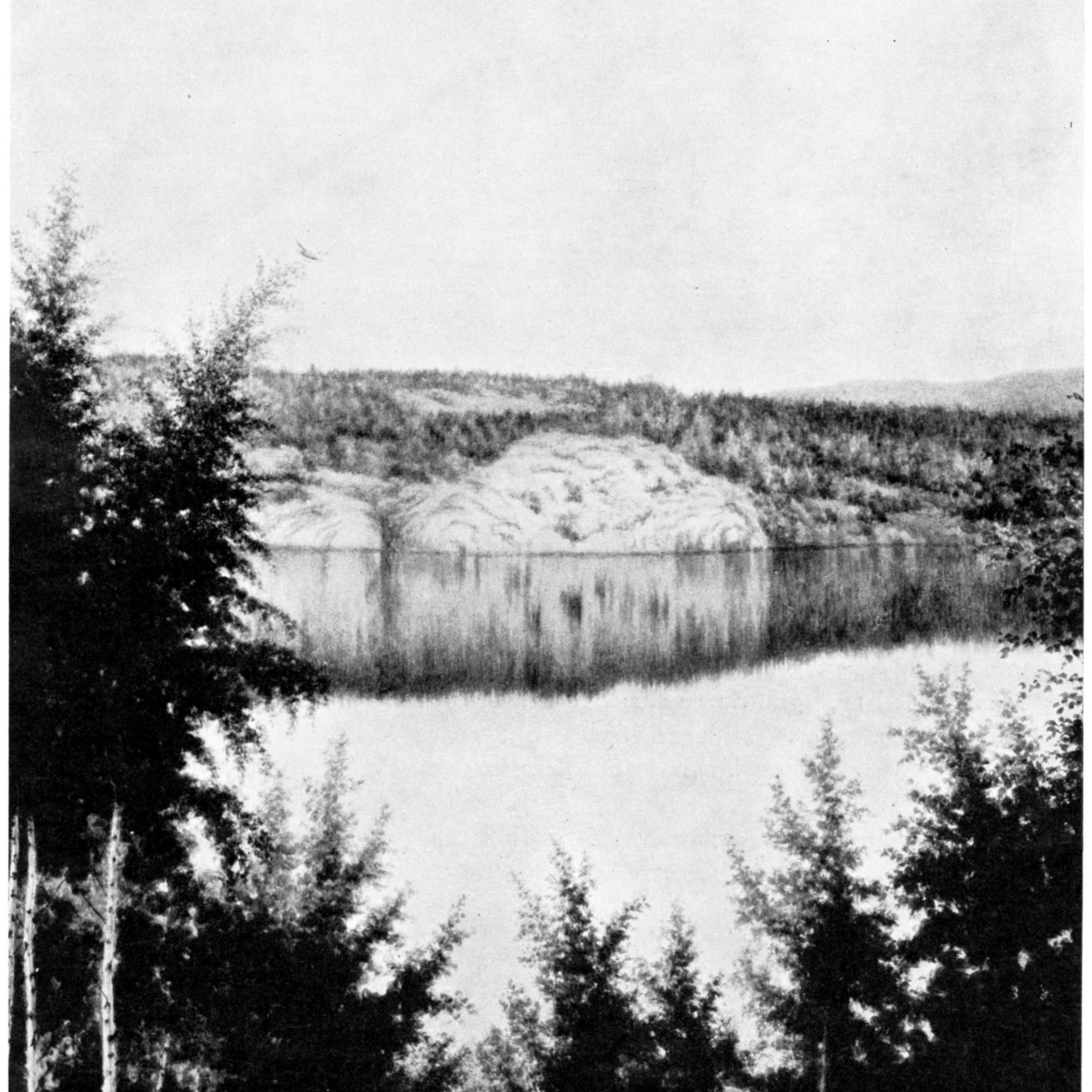} &
    \includegraphics[width=2cm,height=2cm]{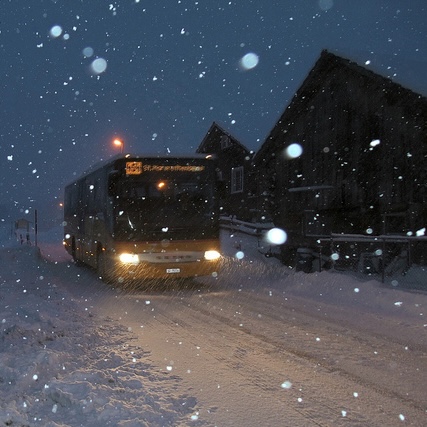} &
    \includegraphics[width=2cm,height=2cm]{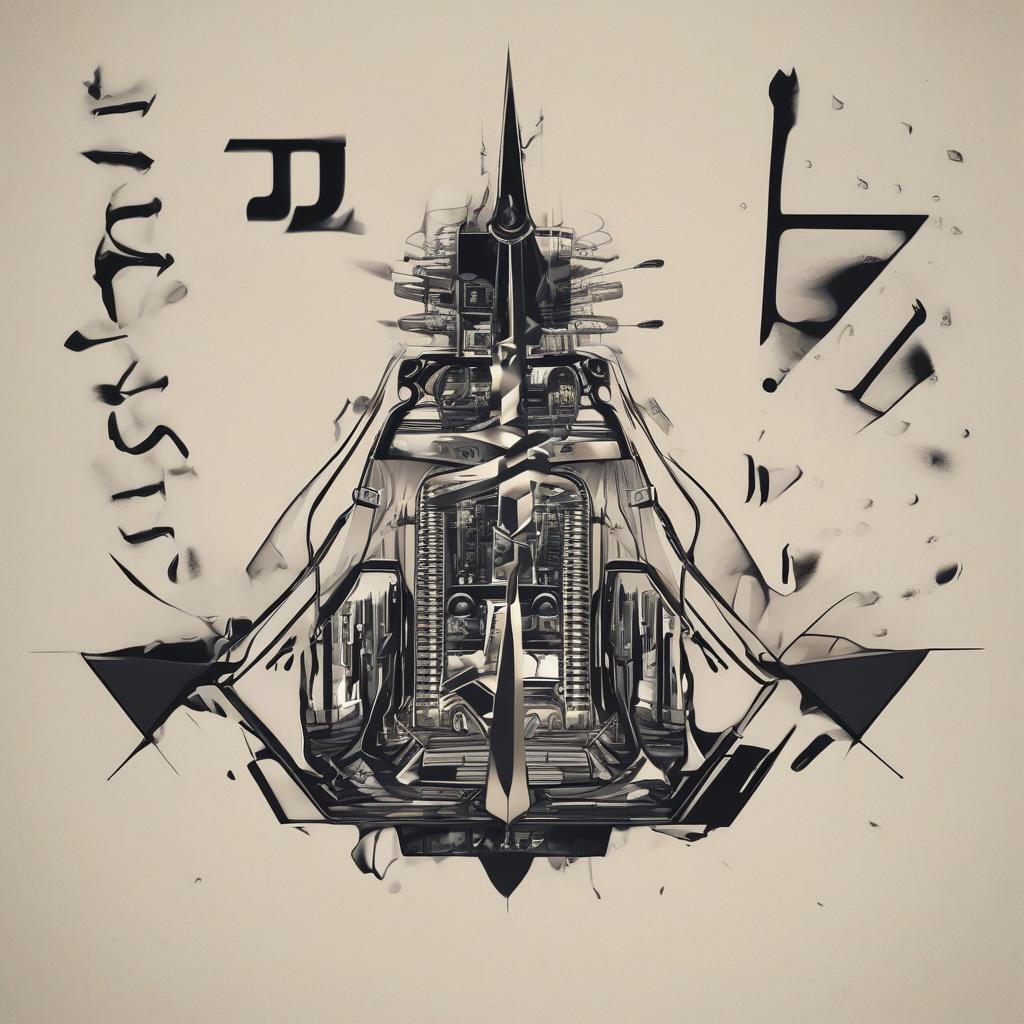} &
    \includegraphics[width=2cm,height=2cm]{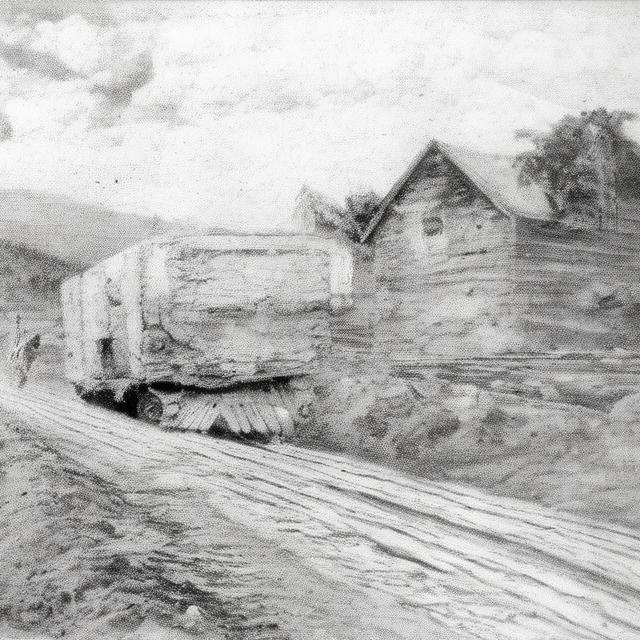} &
    \includegraphics[width=2cm,height=2cm]{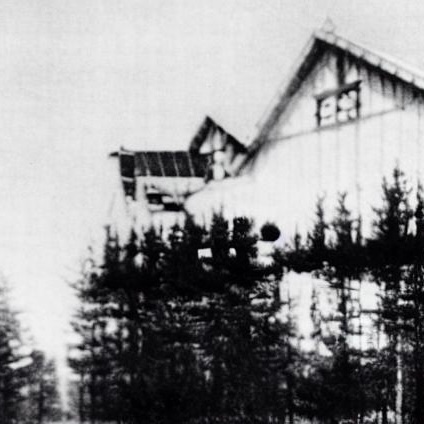} &
    \includegraphics[width=2cm,height=2cm]{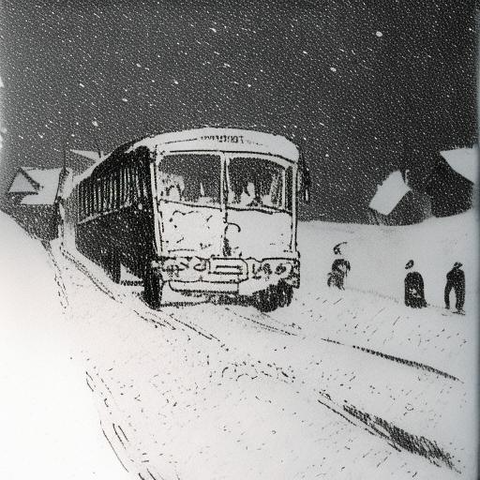} \\
    \bottomrule
    \end{tabular}

    \caption{\textbf{Qualitative Baseline Comparison.} We compare \textit{LouvreSAE} Steering (Kandinsky 2.2 + \textit{LouvreSAE}-20K) to B-LoRA, InstantStyle, and StyleShot, presenting representative examples of style transfer results for identical inputs. In each method, the content image is transferred to the respective target style, whose reference sample is shown.}
    \label{fig:qual_baseline_comp}
\end{figure*}

\subsection{Backbone SAE Training}

\textit{LouvreSAE}-20 achieved an $R^2$ of $0.734$, with a mean activation sparsity of $z$-sparsity of $32.48$. Dead features made up $19.9\%$ of the dictionary, suggesting strong and diverse feature utilization. Representative examples of concepts learned by the SAE are shown in Figure~\ref{fig:concept_overview}. Through a qualitative inspection, we found the model to capture a wide range of concepts from semantic, object-oriented ones to more abstract and art-specific concepts. Additional concept examples are provided in Appendix~\ref{app:add_res_concepts}.

Table~\ref{tab:baselines} provides a direct quantitative comparison of \textit{LouvreSAE}-20 and FeatLab SAE when used as backbone SAEs in the full style transfer pipeline. Our SAE outperforms the baseline on four out of five metrics; the exception is CLIP Score Content, which primarily measures semantic preservation rather than stylistic fidelity. Qualitative inspection of results in Figure~\ref{fig:sae_comp} seems to support this: FeatLab SAE alters semantic content rather than altering the style.

\begin{table}[t]
\centering
\renewcommand{\arraystretch}{1.2}
\setlength{\tabcolsep}{3pt}
\begin{tabular}{
    l l | *{2}{>{\centering\arraybackslash}m{1.9cm}}
}
\toprule
\textbf{Method} & \textbf{Backbone} & \textbf{Style Extraction} & \textbf{Image Generation} \\
\midrule
B-LoRA & SD XL~\cite{podell2023sdxl} & 660 & 30 \\
InstantStyle & SD XL~\cite{podell2023sdxl} & --- & 78 \\
StyleShot & SD 1.5~\cite{rombach2022high} & --- & 60 \\
\midrule
LouvreSAE & Kan. 2.2~\cite{razzhigaev2023kandinsky} & 6 & 29 \\
\bottomrule
\end{tabular}
\caption{
\textbf{Runtime Comparison with Baselines.} Average runtime of style extraction (where applicable) and image generation comparing MosaicSAE to three baselines, in seconds, averaging 50 repetitions on an NVIDIA A100 SXM 40GB with a single job running. Style extraction assumes 40 training images. Image sizes were fixed across methods.}
\label{tab:runtime}
\end{table}

\begin{figure*}[t]
\centering
\renewcommand{\arraystretch}{1.05}
\setlength{\tabcolsep}{4pt}

\begin{tabular}{
    >{\centering\arraybackslash}m{0.7cm}
    >{\centering\arraybackslash}m{3.6cm}
    >{\centering\arraybackslash}m{2.0cm}|
    >{\centering\arraybackslash}m{2.0cm}
    >{\centering\arraybackslash}m{2.0cm}
    >{\centering\arraybackslash}m{2.0cm}
    >{\centering\arraybackslash}m{2.0cm}
    >{\centering\arraybackslash}m{0.7cm}
}
\toprule
& \multicolumn{2}{c|}{\textbf{Concept}} &
$\boldsymbol{\alpha_1}$ &
$\boldsymbol{\alpha_2}$ &
$\boldsymbol{\alpha_3}$ &
$\boldsymbol{\alpha_4}$ & \\
\midrule

&
\raisebox{0pt}[0pt][0pt]{%
\parbox{3.6cm}{\centering
    Painterliness\\[3pt]
    \colorbox{gray!20}{\parbox{2.8cm}{\centering Artistic Aesthetics}}
}}
&
\includegraphics[width=2cm,height=2cm]{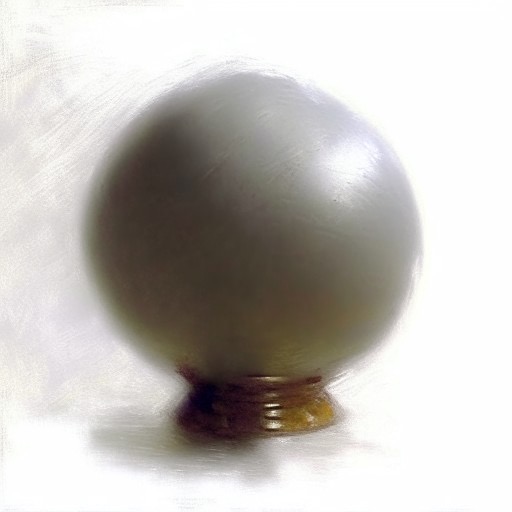}
&
\includegraphics[width=2cm,height=2cm]{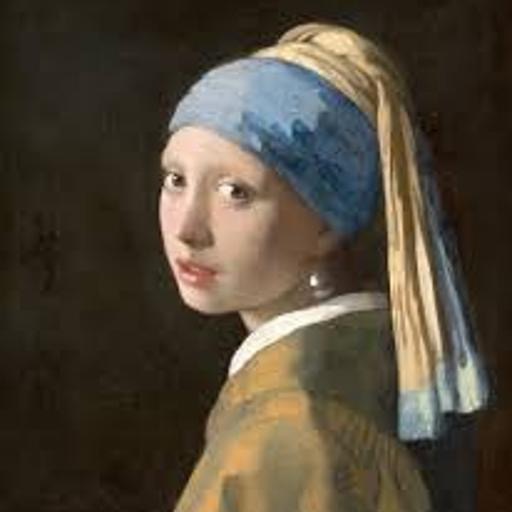}
&
\includegraphics[width=2cm,height=2cm]{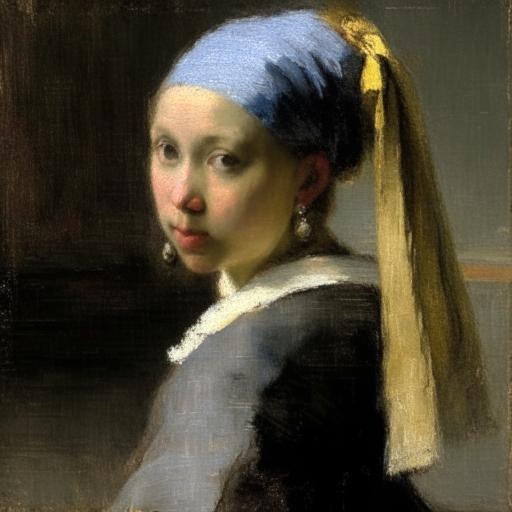}
&
\includegraphics[width=2cm,height=2cm]{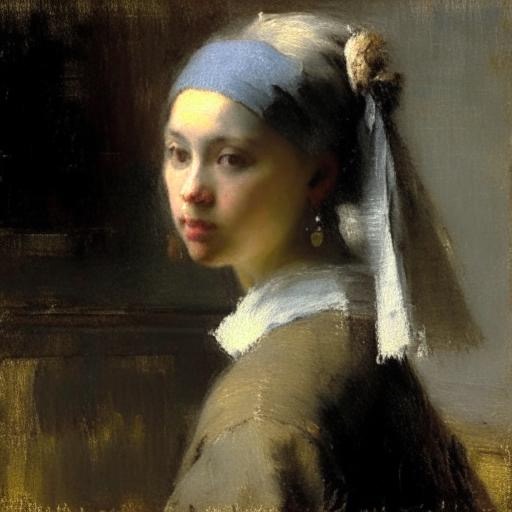}
&
\includegraphics[width=2cm,height=2cm]{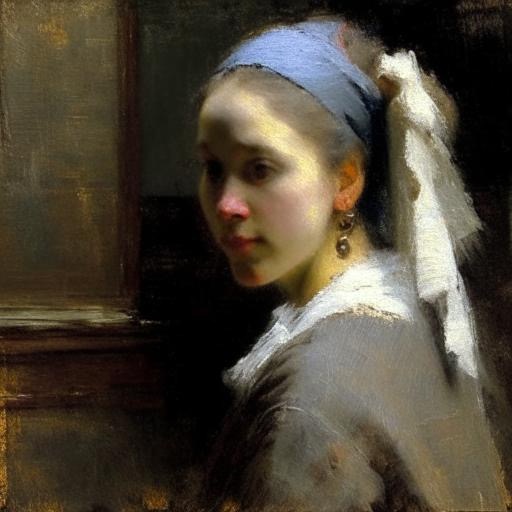} & 
\\[0.35cm]

&
\raisebox{0pt}[0pt][0pt]{%
\parbox{3.6cm}{\centering
    Aerial Perspective\\[3pt]
    \colorbox{gray!20}{\parbox{2.8cm}{\centering Form, Volume, Geometry}}
}}
&
\includegraphics[width=2cm,height=2cm]{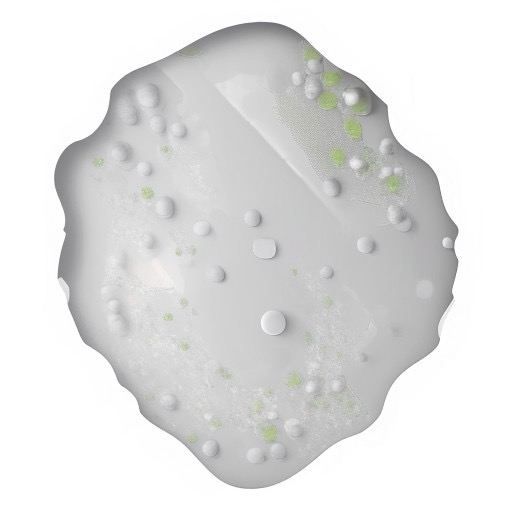}
&
\includegraphics[width=2cm,height=2cm]{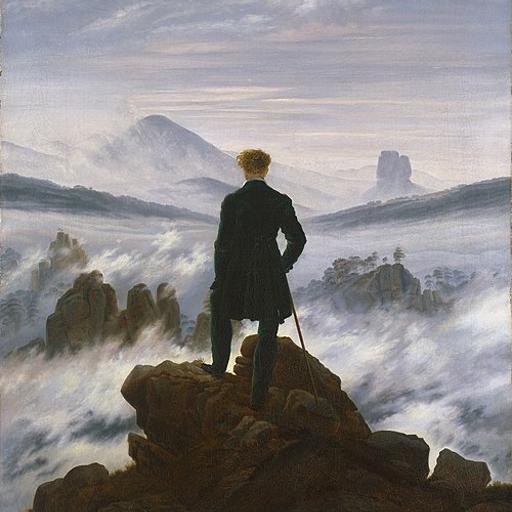}
&
\includegraphics[width=2cm,height=2cm]{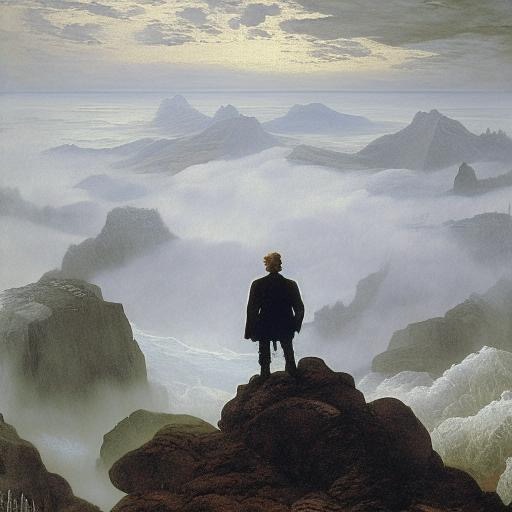}
&
\includegraphics[width=2cm,height=2cm]{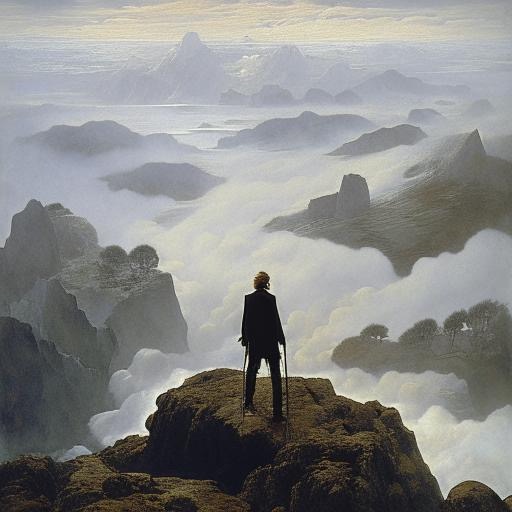}
&
\includegraphics[width=2cm,height=2cm]{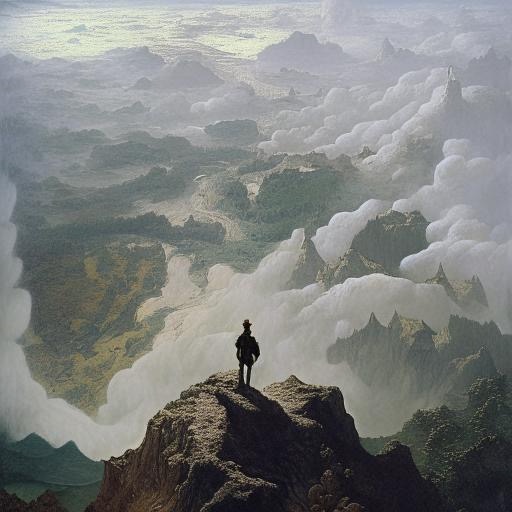} &
\\

\bottomrule
\end{tabular}

\caption{\textbf{Concept Steering Examples.} Increasing steering intensity governed by the $\alpha$ coefficient is shown above for two concepts.}
\label{fig:qual_concept_steer}
\end{figure*}

\subsection{Style Profiles}

With the backbone trained, we next construct style profiles. While difficult to evaluate quantitatively due to a lack of ground-truth supervision, differences become apparent when comparing closely related styles. Figure~\ref{fig:style_profile_diff} highlights two perceptually and historically similar Impressionist styles: Paul Cézanne and Pierre-Auguste Renoir. Despite shared cultural context, their profiles reveal distinct preferences for certain concepts, while others differ mainly in relative activation strength: for instance, \texttt{Thick brushwork} is more strongly expressed by Cézanne.

\subsection{Style Transfer}

We further validate the benefits of our backbone SAE through extensive baseline comparisons on ArtBench10. As summarized in Table~\ref{tab:baselines}, our method achieves the strongest performance on style-focused metrics, yielding a VGG Style Loss of $1.73!\times!10^{-5}$ compared to $2.48!\times!10^{-5}$ for InstantStyle, $3.74!\times!10^{-5}$ for StyleShot, and $1.63!\times!10^{-4}$ for B-LoRA. Likewise, we achieve the highest CLIP Style Score ($0.27$ vs. $0.25$ and $0.21$).

On content-focused metrics, our approach remains competitive: VGG Content Loss of $35.04$ sits within the baseline range of $22.94$–$42.53$, LPIPS of $0.73$ within $0.64$–$0.77$, and CLIP Content Score of $0.26$ within $0.18$–$0.30$.

Qualitative results in Figure~\ref{fig:qual_baseline_comp} further reinforce these findings. While baselines often adjust palette and texture without significant structural change, visible in the Arthur Rackham and Hiroshige cases, our method more faithfully inherits stylistic properties such as layout alteration and detail modulation. Although this can lead to greater semantic deviation, it may better reflect the full nature of artistic style, according to our operational definition.

Runtime results in Table~\ref{tab:runtime} demonstrate a practical advantage: despite improved style fidelity, our method is $1.7$–$20\times$ faster overall, due to efficient style extraction and image generation.

\subsection{Fine-grained Control}

Finally, the decomposable and interpretable structure of our backbone SAE allows downstream manipulation of individual style concepts, applied either at the level of entire profiles or on a per-image basis. As shown in Figure~\ref{fig:qual_concept_steer}, this enables flexible and user-controlled adjustments that modulate stylistic attributes without retraining.

\section{Discussion}
\label{sec:discussion}

This paper introduced \textit{LouvreSAE}, a method leveraging art-specific SAEs to operationalize artistic style as a sparse, decomposable set of  concepts. By constructing style profiles from these features, our approach enables lightweight, interpretable, and steerable style transfer without requiring model fine-tuning. We demonstrate the viability of using SAEs as an easily integrable module to move beyond opaque style representations.




%




%


\paragraph{Limitations.} \textit{LouvreSAE} Steering ultimately depends on the quality and disentanglement of its backbone SAE. While some style and semantic features remain correlated in training data, leading to occasional content leakage into style profile, our approach provides strong control across diverse styles. In some cases, this even leads to more desirable outputs given that structure, perspective, and other features get transferred. Our representation of a style profile captures common stylistic cues, though it may overlook multimodal variations within an oeuvre. Moreover, steering currently uses a linear intervention, which can be imperfect for highly nonlinear style effects. Finally, style transfer evaluation remains challenging, and perceptual nuances still benefit from qualitative examination alongside metrics.

\paragraph{Future Work.} Looking ahead, interpretable generative modeling is poised to move toward more robust, large-scale concept-level control. A key community challenge remains in addressing entanglement, requiring new architectures or training paradigms that more perfectly disentangle semantic content from stylistic or abstract features. Concurrently, there is a clear need for expanding the autointerpretability and taxonomy of discovered features, building larger and more standardized dictionaries to serve as a common vocabulary for model control.




\section*{Acknowledgments}

This work was supported by a Brown Institute Magic Grant. The authors thank Yael Vinker for helpful discussions. MB also thanks the friendly staff of his local IHOP, where much of this paper was written.

\newpage

{
    \small
    \bibliographystyle{ieeenat_fullname}
    \bibliography{main}
}


\newpage
\appendix

\section{Baseline Considerations}
\label{app:baseline_consideration}

\subsection{Backbone SAE}
\label{app:sae_baseline_considerations}

We train the SAE on CLIP embeddings to enable the direct projection of style reference images into the steerable latent space. Unlike architectures that target internal U-Net activations, which necessitate complex optimization or inversion to map reference signals to learned features, our configuration permits immediate, inversion-free encoding of the target style. 

Accordingly, we benchmark against the FeatureLab SAE~\cite{daujotas2024interpreting}, the only comparable model trained on the same CLIP embedding space. This ensures a direct evaluation of dictionary quality and steerability within an identical latent topology.

\subsection{Style Transfer}
\label{app:style_transfer_baseline_considerations}

We restrict our baseline comparison to methods that share our operational paradigm: reference-based style encoding applied to I2I synthesis. This selection isolates methods capable of transforming a specific content image using visual exemplars, ensuring a direct comparison of content preservation and stylistic fidelity.

\section{\textit{LouvreSAE} Deployment Insights}
\label{app:louvresae_training}

Training and deploying SAEs is often unstable and sensitive to hyperparameter configurations. During the development of \textit{LouvreSAE}, we gathered several practical insights regarding effective training practices and inference strategies. We provide these here to aid researchers intending to build upon our results. The optimization and precise resulting configuration of \textit{LouvreSAE}-20, which we ended up using, are described in Appendix~\ref{app:implementation}.

\begin{enumerate}
    \item \textbf{Granularity and Dictionary Size.} 
    Dictionary size ($M$) correlates with feature granularity. While SAEs successfully disentangle dense CLIP embeddings, we found a trade-off in that smaller dictionaries (${<}$10K) yield broad concepts, while larger ones (${>}$80K) capture fine textures but increase redundancy. We found $M{=}$20K optimal for our use case of style transfer.

    \item \textbf{The Reconstruction-Rank Trade-off.} 
    Higher learning rates improved reconstruction ($R^2$) but collapsed the decoder's stable rank, indicating reliance on fewer effective dimensions. We prioritized configurations with higher stable rank over maximum $R^2$ to ensure a diverse vocabulary of steerable concepts.

    \item \textbf{Efficacy of Reanimation Mechanisms.} 
    We found the global \texttt{BatchTopK} architecture more effective for maximizing feature utilization than auxiliary "dead neuron" losses (e.g., ghost grads). Explicit reanimation terms often introduced noise without yielding interpretable semantic or stylistic features.

    \item \textbf{Steering Mechanics.} 
    Additive steering ($E_{steered} = E_{content} + \alpha \Delta_{style}$) proved more stable than coordinate clamping. Clamping disrupts the semantic structure of CLIP embeddings, whereas additive intervention shifts stylistic representation while preserving the underlying content geometry.

    \item \textbf{Autointerpretability Heuristics.} 
    VLMs performed poorly when provided with negative examples. The most reliable labeling strategy paired positive exemplars with a prototype image. Additionally, the most stylistically relevant concepts appeared consistently within the top-1000 features by mean activation; long-tail features often represented noise or non-generalizable artifacts. See Appendix~\ref{app:autointerp} for more details.
\end{enumerate}

\section{Implementation Details}
\label{app:implementation}
\subsection{Kandinsky 2.2 + \textit{LouvreSAE}}
\label{subsec:impl_louvresae}

\paragraph{Backbone SAE Training.}

We used Weights \& Biases to conduct a grid search over the training hyperparameters of our \textit{LouvreSAE}, which include the number of concepts \(M \in \{10{,}240, 20{,}480, 81{,}920\}\) (expansion factors \(\{8\times, 16\times, 64\times\}\) given \(D=1280\)), the top-\(k \in \{16, 32\}\), and the learning rate \(\eta \in \{10^{-4}, 5\times 10^{-5}, 10^{-5}\}\). This comprised \(N = 3 \times 2 \times 3 = 18\) runs over all combinations. For all runs, we fixed the batch size \(B=8192\), epochs \(E=30\), warmup \(W=100\), Adam \((\beta_1,\beta_2)=(0.9, 0.999)\) with zero weight decay, and enforced sparsity via the architectural BatchTop-\(K\) mechanism. We report reconstruction \(R^2\), dictionary stable rank, and $L_2$ loss for these runs in Table~\ref{tab:grid_search}.

We compute the stable rank of the decoder \(W\) as
\[
\mathrm{sr}(W)\;=\;\frac{\|W\|_F^2}{\|W\|_2^2}
\;=\;\frac{\sum_{i=1}^{r}\sigma_i^2}{\sigma_1^2},
\]
where \(\|W\|_F^2=\sum_{i,j} W_{ij}^2\) is the squared Frobenius norm, \(\|W\|_2=\sigma_1\) is the spectral norm, and \(\sigma_1\ge\sigma_2\ge\cdots\ge 0\) are the singular values of \(W\). We observe a consistent effect of learning rate on stable rank: higher learning rates compress the spectrum, yielding lower stable rank and more concentrated dictionaries.

\begin{table}[t]
\centering
\begin{tabular}{r r r r r r}
\toprule
$M$ & $k$ & $\eta$ & stable rank & $R^2$ & $L_2$ loss
\\
\midrule
10240 & 16 & 1e-5 & 639.999 & 0.5481 & 0.6248 \\
10240 & 16 & 5e-5 & 521.528 & 0.6395 & 0.4976 \\
10240 & 16 & 1e-4 & 254.441 & 0.6415 & 0.4951 \\
10240 & 32 & 1e-5 & 625.484 & 0.6486 & 0.4862 \\
10240 & 32 & 5e-5 & 508.804 & 0.7128 & 0.3970 \\
10240 & 32 & 1e-4 & 211.631 & 0.7159 & 0.3931 \\
\addlinespace\midrule
20480 & 16 & 1e-5 & 752.641 & 0.5643 & 0.6024 \\
20480 & 16 & 5e-5 & 637.256 & 0.6496 & 0.4836 \\
20480 & 16 & 1e-4 & 266.289 & 0.6655 & 0.4617 \\
20480 & 32 & 1e-5 & 707.774 & 0.6591 & 0.4716 \\
\textbf{20480} & \textbf{32} & \textbf{5e-5} & \textbf{674.755} & \textbf{0.7337} & \textbf{0.3681} \\
20480 & 32 & 1e-4 & 158.385 & 0.7348 & 0.3669 \\
\addlinespace\midrule
81920 & 16 & 1e-5 & 977.434 & 0.5761 & 0.5862 \\
81920 & 16 & 5e-5 & 804.337 & 0.6919 & 0.4251 \\
81920 & 16 & 1e-4 & 663.620 & 0.7167 & 0.3907 \\
81920 & 32 & 1e-5 & 902.814 & 0.6942 & 0.4233 \\
81920 & 32 & 5e-5 & 711.387 & 0.7760 & 0.3096 \\
81920 & 32 & 1e-4 & 471.910 & 0.7963 & 0.2816 \\
\bottomrule
\end{tabular}
\caption{\textbf{Hyperparameter sweep for \textit{LouvreSAE}.} We report stable rank, reconstruction $R^2$, and $L_2$ loss across all 18 training configurations. The bolded setting $(M=20{,}480,\, k=32,\, \eta=5\times10^{-5})$ was selected because it balances high reconstruction performance and high stable rank on the Pareto frontier, and it qualitatively produced the most coherent concepts relevant to style transfer.}
\label{tab:grid_search}
\end{table}

\paragraph{Style Profile Encoding.}

Given an image $x$, we obtain its $D$-dimensional prior embedding $e \in \mathbb{R}^D$ and encode it using the trained sparse autoencoder (SAE) to produce a nonnegative concept vector $z \in \mathbb{R}^M$:
\[
e = f_{\text{enc}}(x), \qquad 
z = f_{\text{SAE}}(e).
\]
For a reference set of $N$ images, concept activations are stacked into $Z \in \mathbb{R}^{N \times M}$.  
A concept $j$ is considered characteristic if it is active in at least $P$ of the images.  
In all of our experiments, we use a presence threshold of $P = 0.6$.

The style profile $V_S \in \mathbb{R}^M$ assigns each selected concept its mean positive activation, scaled by a profile-strength parameter $S$:
\[
V_S[j] \;=\;
S \cdot 
\frac{
    \sum_{i=1}^{N} Z_{i,j}\,\mathbf{1}[Z_{i,j}>0]
}{
    \sum_{i=1}^{N}\mathbf{1}[Z_{i,j}>0]
}
\]
\[
\text{valid when } 
\frac{1}{N}\sum_{i=1}^{N} \mathbf{1}[Z_{i,j}>0] \ge P,
\text{and else } V_S[j]=0 \text{.}
\]
We use $S = 5$ for constructing all profiles, adjusting it only when explicitly noted.
This produces a sparse, interpretable vector whose magnitude reflects the characteristic strength of the style.


\paragraph{Presence Threshold.}
\label{app:presence_threshold}

To decide the value of the presence threshold $P$, we compared the performance of our method at $P=\{0.4,\ 0.5,\ 0.6,\ 0.7,\ 0.8\}$. We report the results in Table~\ref{tab:p_sweep}. Based on these results, we opted for $P=0.6$.

\begin{table}[t]
\centering
\renewcommand{\arraystretch}{1.2}
\setlength{\tabcolsep}{4pt}
\begin{tabular}{c | c c c c}
\toprule
$P$ 
& \makecell{\textbf{VGG Loss} \\ \textbf{Style} $\downarrow$}
& \makecell{\textbf{VGG Loss} \\ \textbf{Content} $\downarrow$}
& \textbf{LPIPS} $\downarrow$
& \makecell{\textbf{CLIP Score} \\ \textbf{Style} $\uparrow$} \\
\midrule
0.4 & 1.55\ensuremath{\times}10\ensuremath{^{-5}} & 35.77 & 0.7220 & 0.280 \\
0.5 & 1.56\ensuremath{\times}10\ensuremath{^{-5}} & 35.85 & 0.7747 & 0.273 \\
0.6 & 1.73\ensuremath{\times}10\ensuremath{^{-5}} & 35.04 & 0.7300 & 0.270 \\
0.7 & 2.02\ensuremath{\times}10\ensuremath{^{-5}} & 36.43 & 0.7803 & 0.253 \\
0.8 & 2.40\ensuremath{\times}10\ensuremath{^{-5}} & 37.12 & 0.7824 & 0.239 \\
\bottomrule
\end{tabular}
\caption{\textbf{Effect of Parameter $P$ on Performance.} Lower $P$ is better for VGG losses and LPIPS, while higher is better for CLIP style score.}
\label{tab:p_sweep}
\end{table}


\paragraph{Generation Steering.}

Let $W \in \mathbb{R}^{M \times D}$ denote the SAE decoder (dictionary).  
The style residual in the embedding space is obtained by
\[
\Delta_S = V_S W.
\]
For a content image with embedding $E_C$, we apply steering by simple additive intervention:
\[
E_{\text{steered}} = E_C + \alpha\,\Delta_S.
\]
The steered embedding is then passed to the diffusion decoder to generate the final stylized image.  
Across all experiments, we use a steering gain of $\alpha = 2$.

\paragraph{Single-Concept Steering.}
For experiments that steer using a \emph{single} SAE concept, we construct a one-hot vector whose nonzero entry corresponds to the selected concept. Its magnitude is set to that concept's maximum activation observed in the dataset. The resulting residual is obtained by decoding this one-hot vector through the SAE dictionary, identical to the multi-concept procedure. During generation, this residual is added to the content embedding in exactly the same manner as multi-concept style steering, with
\[
\alpha_{\text{single}} \in [0.5,\, 3.5],
\]
which we find yields stable and interpretable images.

\subsection{Metrics \& Evaluation}
\label{app:impl_metrics}

\paragraph{VGG Loss~\cite{gatys2015neural}.} We use a pre-trained VGG19 model (trained on ImageNet), extracting features from 5 layers (ReLU 1-5). Style loss computed via Gram matrices of features with equal weighting (1/5 per layer) and MSE. Content loss uses MSE between ReLU 4 features. Images are resized to 224x224 with standard ImageNet normalization.

\paragraph{LPIPS~\cite{zhang2018unreasonable}.} Implemented via \texttt{torchmetrics} with a VGG backbone. Images are resized to 256x256 and normalized to [-1,1].

\paragraph{CLIP Score~\cite{radford2021learning}.} We use the \texttt{openai/clip-vit-base-patch32} model and compute the cosine similarity between the normalized image and text embedding. For content loss, the ImageNet caption is used; for style loss, the following prompt is used: "A painting in [Artist Name] signature style."

\subsection{Baselines}
\label{app:impl_baselines}

We use the official open-source releases of baselines with default parameters.

\paragraph{B-LoRA~\cite{frenkel2024implicit}.} \ For each comparison, we trained two B-LoRAs: a style B-LoRA which was trained on the 10 test images per style and a content B-LoRA per individual content image. We used an empty prompt to perform an unbiased style transfer and other default training parameters provided on the project GitHub (\url{https://github.com/yardenfren1996/B-LoRA}) to train the B-LoRAs. Then, we performed image stylization based on the reference style images using these trained B-LoRAs and used the input prompt provided by the authors on the official GitHub repository. At inference time, we generated 4 images per style and content pair and modified \texttt{style\_alpha}=$0.5$. 

\paragraph{InstantStyle~\cite{wang2024instantstyle}.} \ InstantStyle takes one content image and one style image per generation, so we performed a generation for each style image from each style per content image, resulting in 10x more generations per style since we had 10 test images for each style. We utilized the script provided in the official HuggingFace app (\url{https://huggingface.co/spaces/InstantX/InstantStyle}) and inputted an empty prompt to perform an unbiased style transfer. We utilized an empty prompt to perform an unbiased style transfer and other default parameters provided with the official GitHub release (\url{https://github.com/instantX-research/InstantStyle}).

\paragraph{StyleShot~\cite{gao2025styleshot}.} \ Similar to InstantStyle, StyleShot takes one content image and one style image per generation. We, thus, performed a generation for each style image from each style per content image. We utilized the \texttt{styleshot\_image\_driven\_demo.py} script provided with the official GitHub release (\url{https://github.com/open-mmlab/StyleShot}) with an empty prompt to perform an unbiased style transfer during inference. 

\section{Autointerpretability and Taxonomy \\ Construction}
\label{app:autointerp}

To make the overcomplete dictionary of our SAEs understandable, we employ a multi-stage autointerpretability pipeline described herein. This process entails using a Vision-Language Model (VLM) for initial labeling, a Large Language Model (LLM) for taxonomy induction, and human verification to ensure art-historical accuracy.

For autointerpretability labels of \textit{LouvreSAE}-20, we used Gemini 2.5 Flash (queried through the Google API) with $k=12$. We open-source these labels at \url{https://louvresae.github.io}.


\subsection{Automated Concept Labeling}

For each concept in the dictionary learned by our SAEs, we first construct a visual demonstration intended to isolate its semantic or stylistic function. We scan the training dataset to retrieve the top-$k$ images that maximally activate the specific concept. These exemplars are combined into a grid image alongside a prototype visualization: a neutral white sphere image steered along the direction of the concept. This prototype visualization assists in isolating texture and brushwork from semantic object content. 

This composite visual grid is passed to a VLM with the following prompt:

\begin{PromptBox}
The grid of images represents a visual concept or feature. Your task is to concisely label this feature. The top row contains a protoype of this concept: a blank white sphere followed by increasingly powerful expressions of the feature. The images underneath are those which contain the described feature. Use this information to provide a label for the feature. Do not give a complete sentence, just your label for the feature.
\end{PromptBox}

\subsection{Taxonomy Induction and Refinement}

To organize the labels from the previous step into a coherent framework, we employ a hybrid hierarchical clustering approach facilitated by an LLM. We feed the aggregate list of generated labels to the LLM and prompt it to categorize them into high-level dimensions (e.g., light effects, artistic styles, textures). We prioritized the top 1,000 highest mean-activation features for this categorization and performed a manual human-in-the-loop refinement stage.

First, we visually inspected the concepts' visual demonstration (from the previous step) to verify that the VLM-generated label accurately reflected the visual activation. Ambiguous or generic labels (e.g., ``lighting") were manually refined to specific terms (e.g., ``oil on canvas" or ``chiaroscuro") where appropriate. Features were then re-sorted into the final taxonomy to ensure that the style profiles reflected semantically meaningful groupings.

\onecolumn

\newpage
\section{Additional Results: SAE Concept Examples}
\label{app:add_res_concepts}

\begin{figure*}[h]
\centering
\renewcommand{\arraystretch}{1.05}
\setlength{\tabcolsep}{4pt}

\begin{tabular}{
    >{\centering\arraybackslash}m{0.8cm}
    >{\centering\arraybackslash}m{3.6cm}
    >{\centering\arraybackslash}m{2.0cm}|
    >{\centering\arraybackslash}m{2.0cm}
    >{\centering\arraybackslash}m{2.0cm}
    >{\centering\arraybackslash}m{2.0cm}
    >{\centering\arraybackslash}m{2.0cm}
    >{\centering\arraybackslash}m{0.8cm}
}
\toprule
& \multicolumn{2}{c|}{\textbf{Concept}} &
\multicolumn{4}{c}{\textbf{Representative Exemplars}} & \\
\midrule

&
\raisebox{0pt}[0pt][0pt]{%
\parbox{3.6cm}{\centering
   Graphic Style\\[3pt]
    \colorbox{gray!20}{\parbox{2.8cm}{\centering Style \& Cultural / Artistic Aesthetics}}
}}
&
\includegraphics[width=2cm,height=2cm]{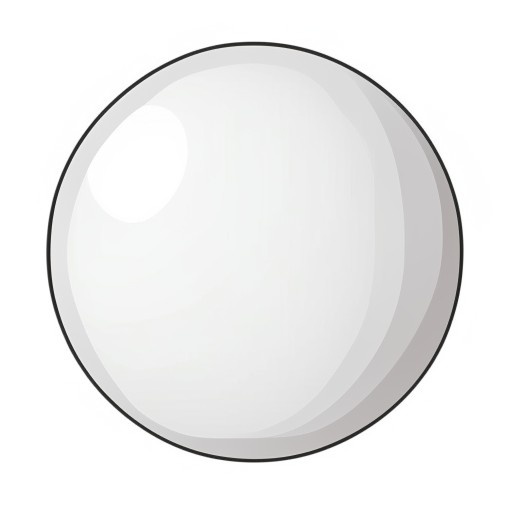}
&
\includegraphics[width=2cm,height=2cm]{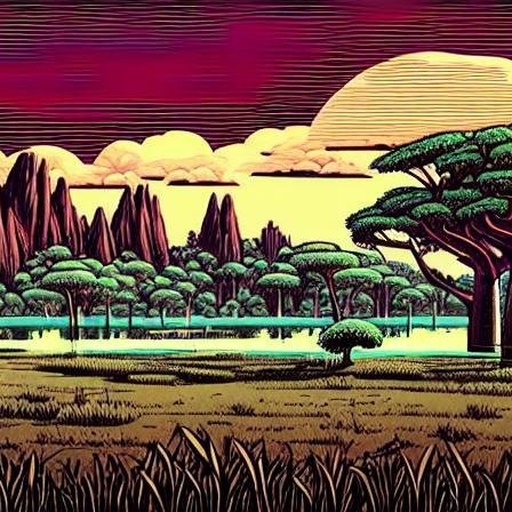}
&
\includegraphics[width=2cm,height=2cm]{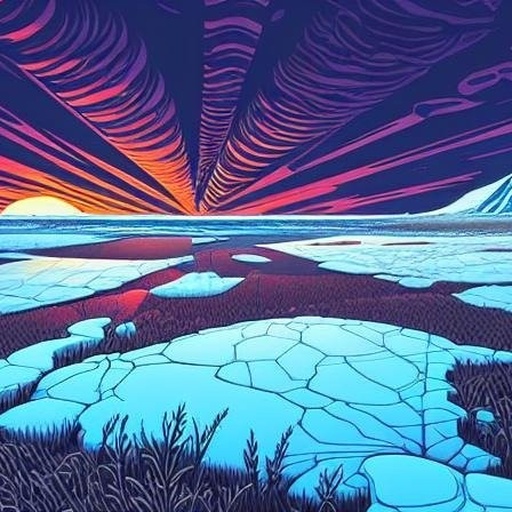}
&
\includegraphics[width=2cm,height=2cm]{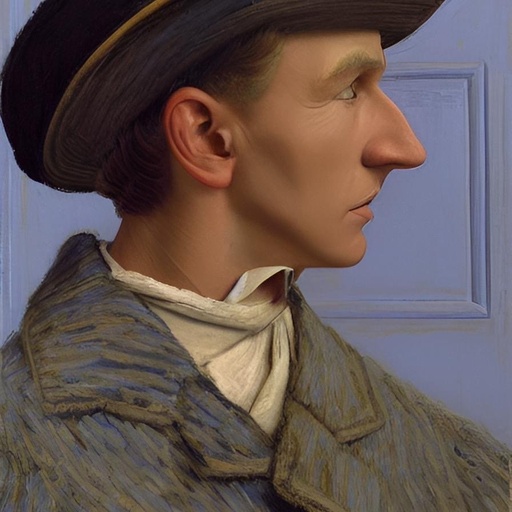}
&
\includegraphics[width=2cm,height=2cm]{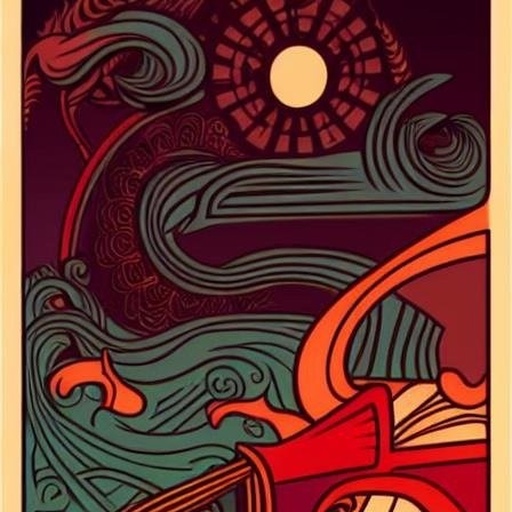} &
\\

&
\raisebox{0pt}[0pt][0pt]{%
\parbox{3.6cm}{\centering
   Water reflection\\[3pt]
    \colorbox{gray!20}{\parbox{2.8cm}{\centering Reflectivity \& Specular Light Behavior}}
}}
&
\includegraphics[width=2cm,height=2cm]{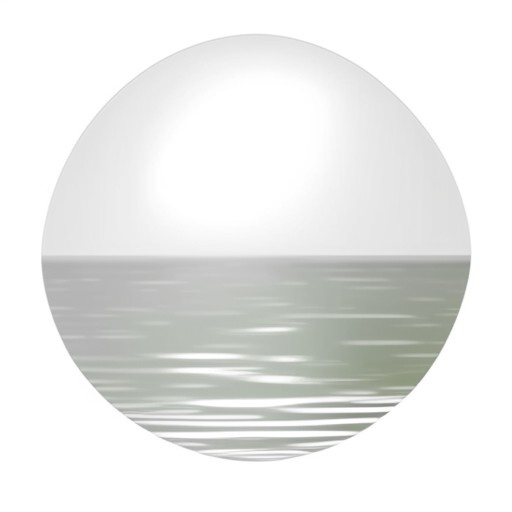}
&
\includegraphics[width=2cm,height=2cm]{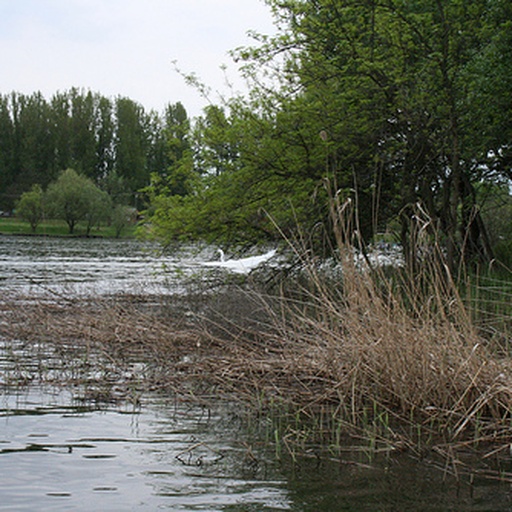}
&
\includegraphics[width=2cm,height=2cm]{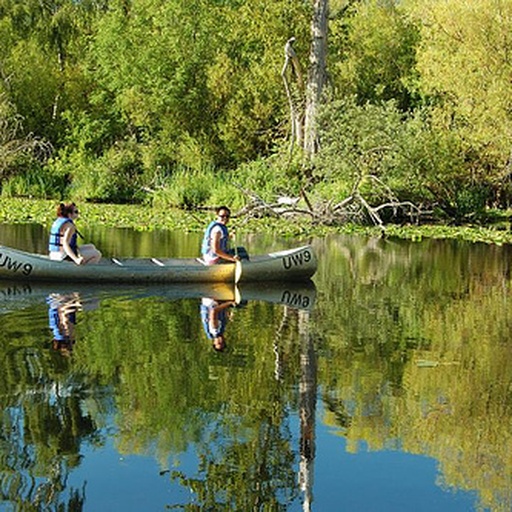}
&
\includegraphics[width=2cm,height=2cm]{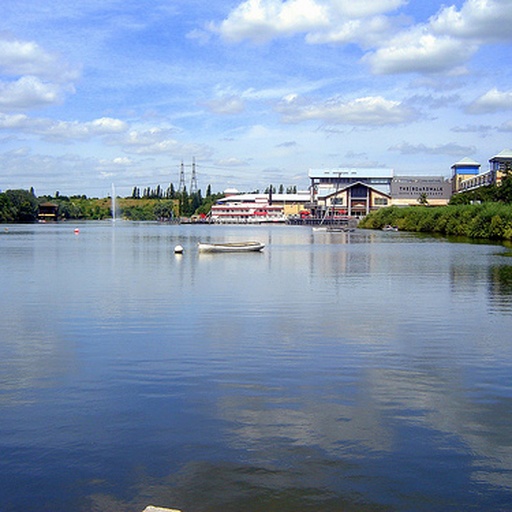}
&
\includegraphics[width=2cm,height=2cm]{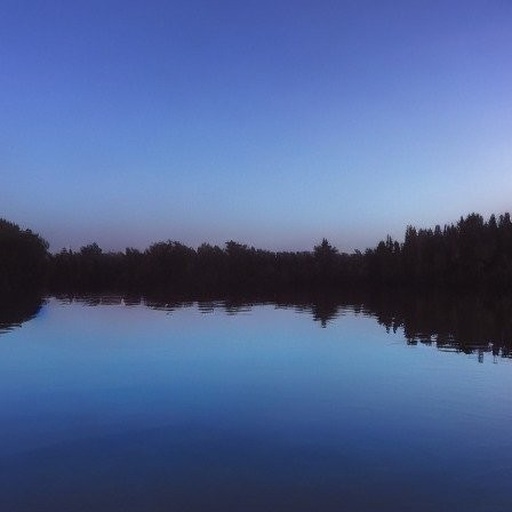} &
\\

&
\raisebox{0pt}[0pt][0pt]{%
\parbox{3.6cm}{\centering
   Cross-stitch\\[3pt]
    \colorbox{gray!20}{\parbox{2.8cm}{\centering Texture \& Material Properties}}
}}
&
\includegraphics[width=2cm,height=2cm]{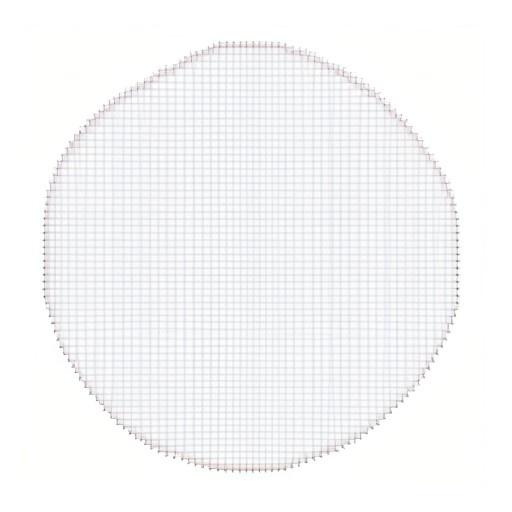}
&
\includegraphics[width=2cm,height=2cm]{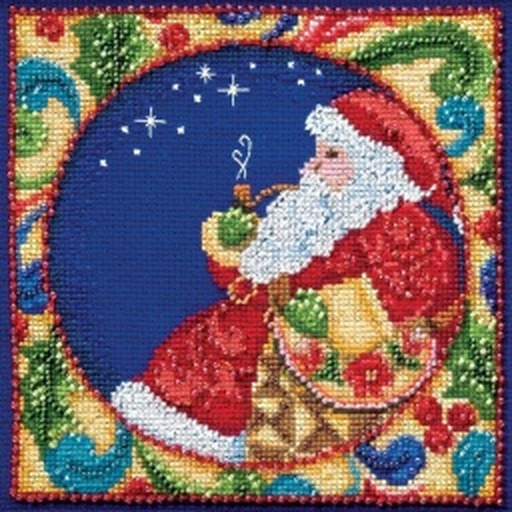}
&
\includegraphics[width=2cm,height=2cm]{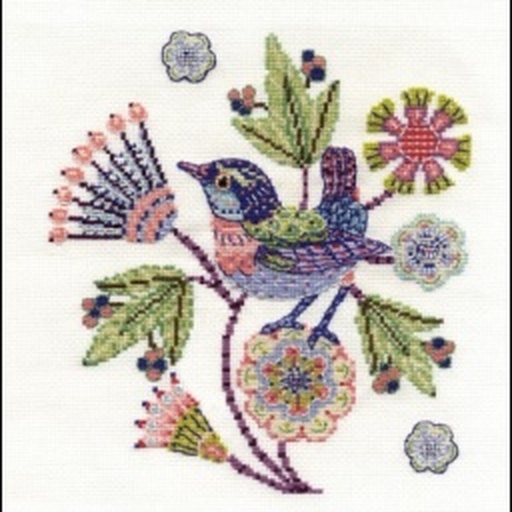}
&
\includegraphics[width=2cm,height=2cm]{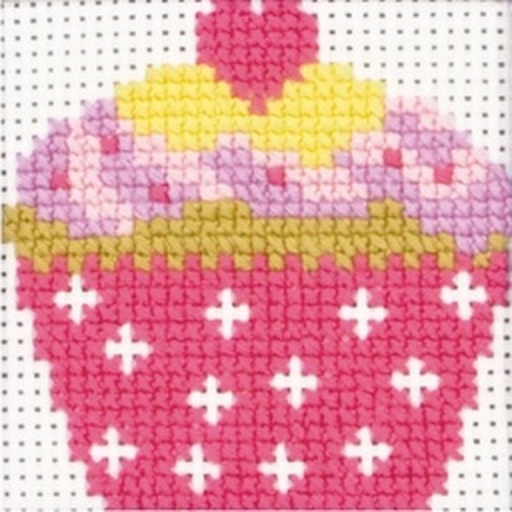}
&
\includegraphics[width=2cm,height=2cm]{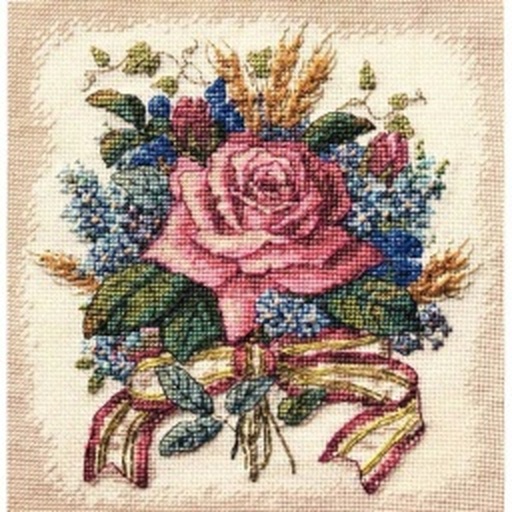} &
\\

&
\raisebox{0pt}[0pt][0pt]{%
\parbox{3.6cm}{\centering
   Pink\\[3pt]
    \colorbox{gray!20}{\parbox{2.8cm}{\centering Color \& Chromatic Qualities}}
}}
&
\includegraphics[width=2cm,height=2cm]{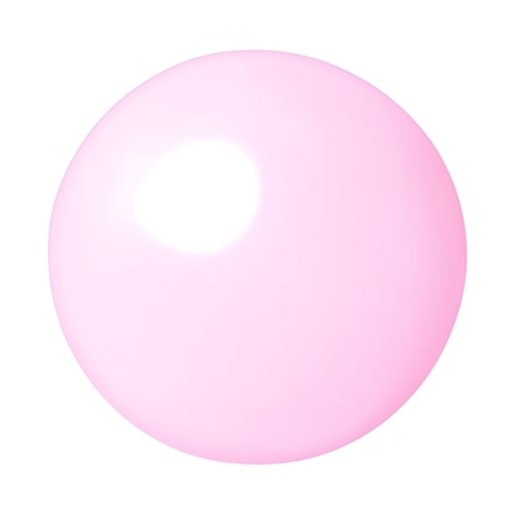}
&
\includegraphics[width=2cm,height=2cm]{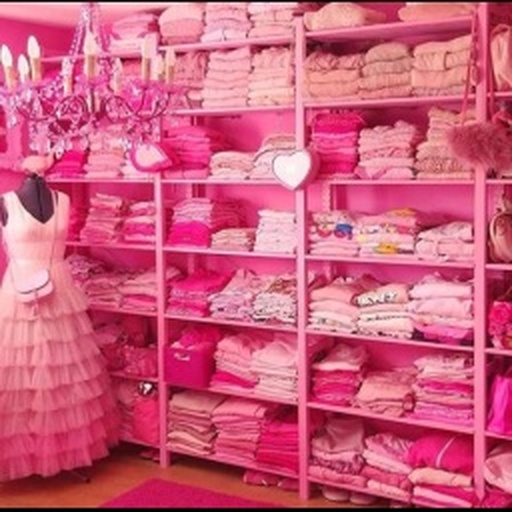}
&
\includegraphics[width=2cm,height=2cm]{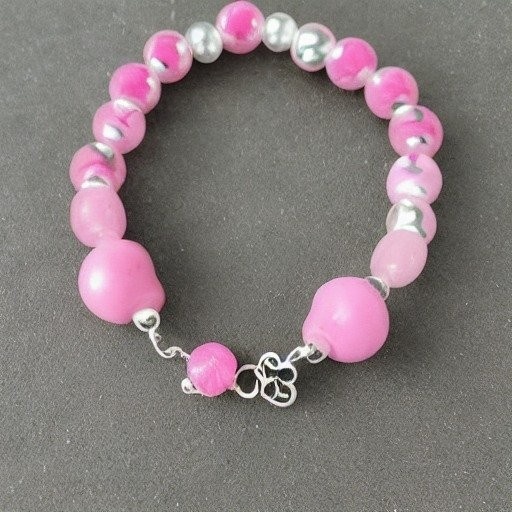}
&
\includegraphics[width=2cm,height=2cm]{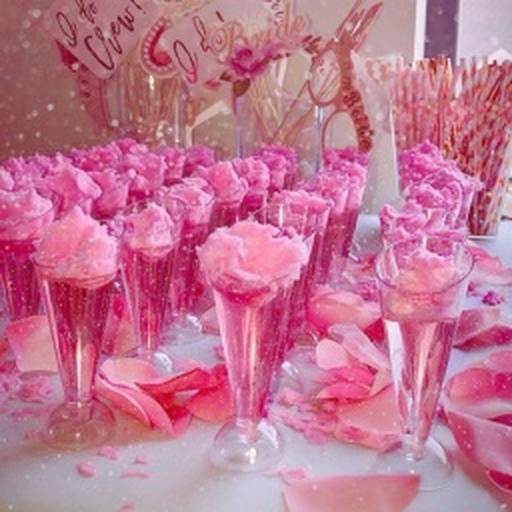}
&
\includegraphics[width=2cm,height=2cm]{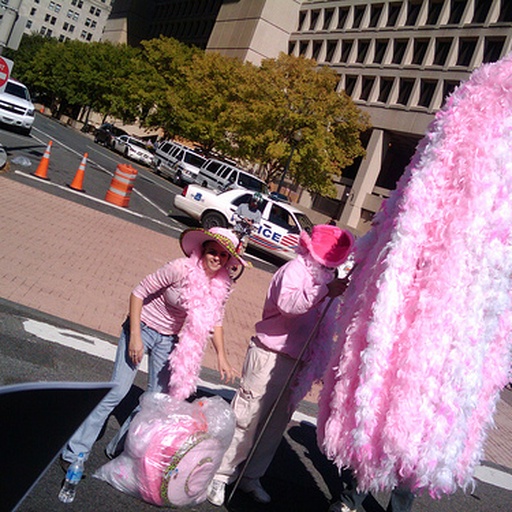} &
\\

&
\raisebox{0pt}[0pt][0pt]{%
\parbox{3.6cm}{\centering
   Flat Design\\[3pt]
    \colorbox{gray!20}{\parbox{2.8cm}{\centering Style \& Cultural / Artistic Aesthetics}}
}}
&
\includegraphics[width=2cm,height=2cm]{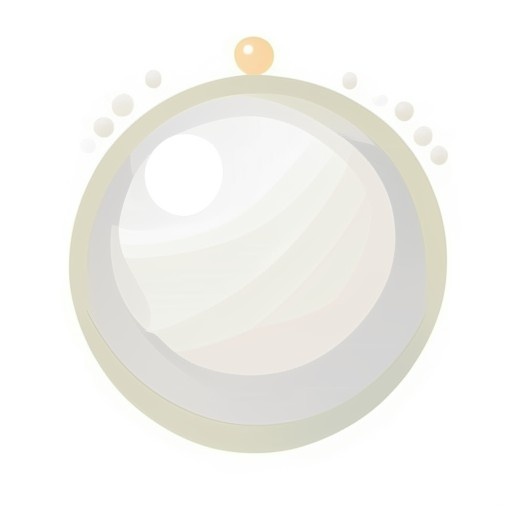}
&
\includegraphics[width=2cm,height=2cm]{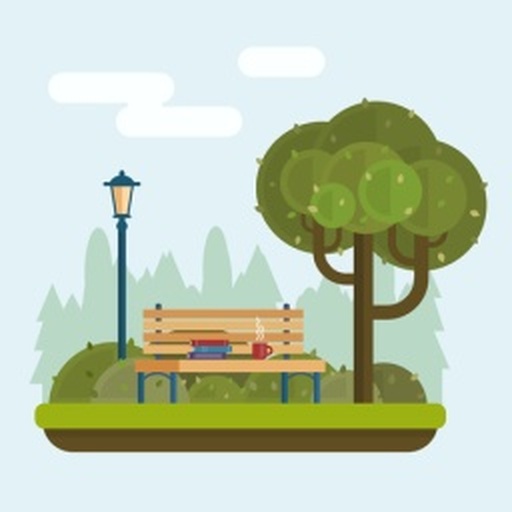}
&
\includegraphics[width=2cm,height=2cm]{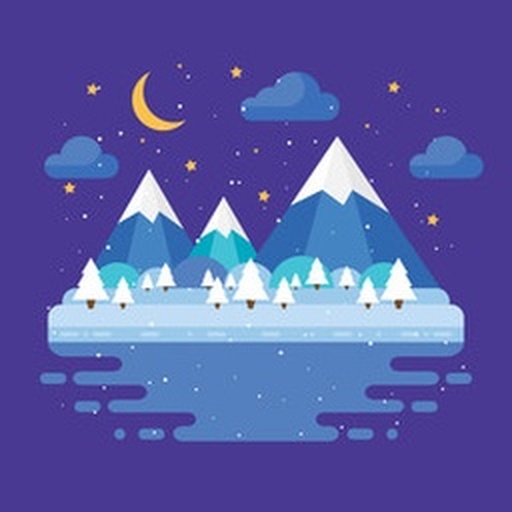}
&
\includegraphics[width=2cm,height=2cm]{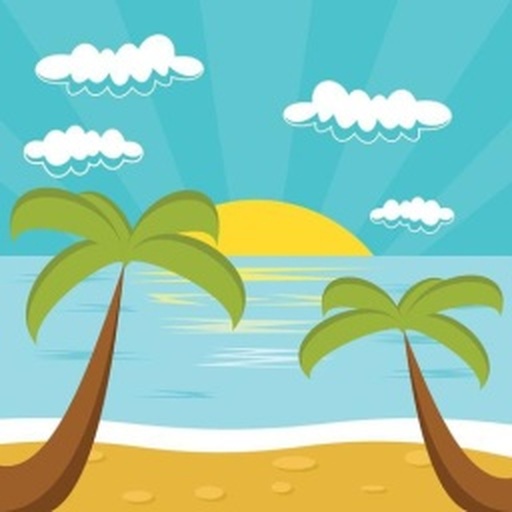}
&
\includegraphics[width=2cm,height=2cm]{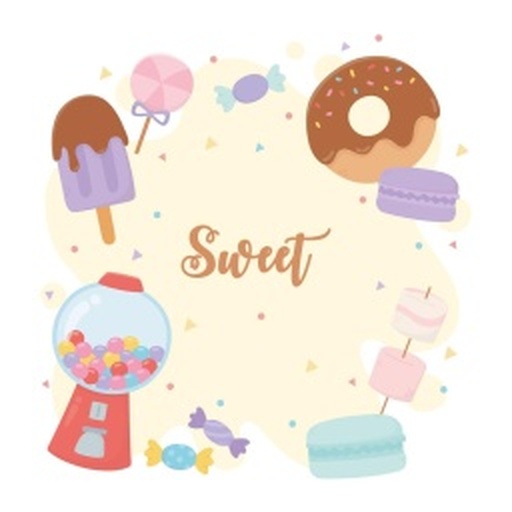} &
\\

&
\raisebox{0pt}[0pt][0pt]{%
\parbox{3.6cm}{\centering
   Chiaroscuro\\[3pt]
    \colorbox{gray!20}{\parbox{2.8cm}{\centering General Lighting \& Shading}}
}}
&
\includegraphics[width=2cm,height=2cm]{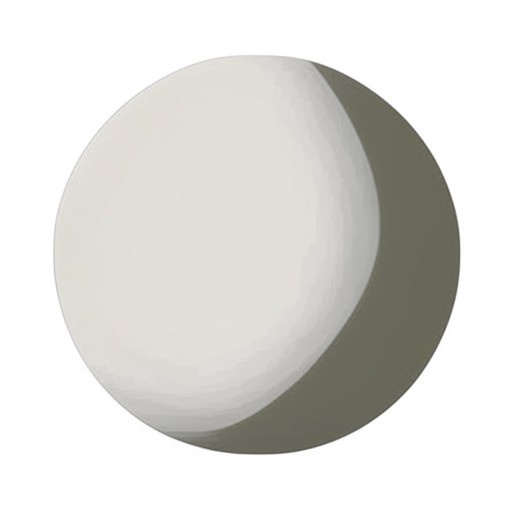}
&
\includegraphics[width=2cm,height=2cm]{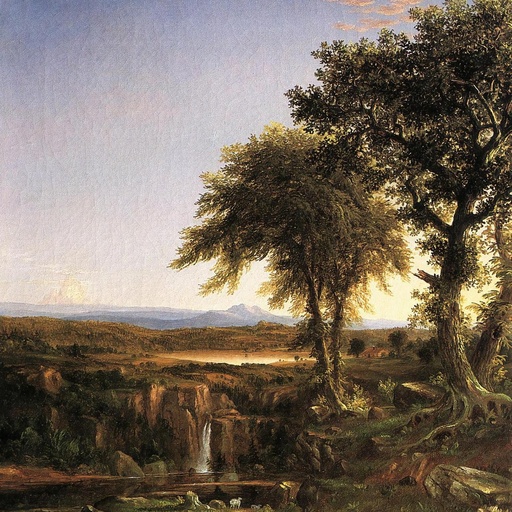}
&
\includegraphics[width=2cm,height=2cm]{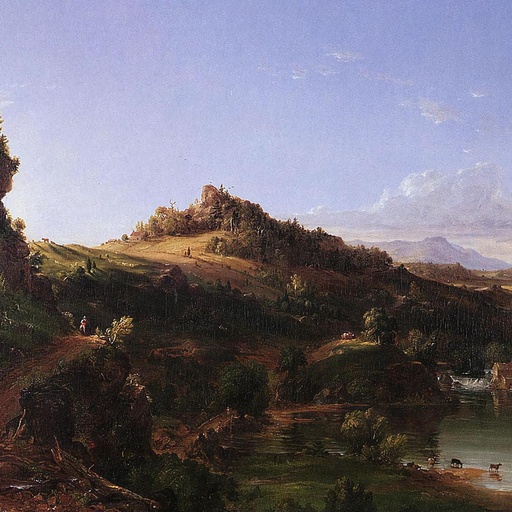}
&
\includegraphics[width=2cm,height=2cm]{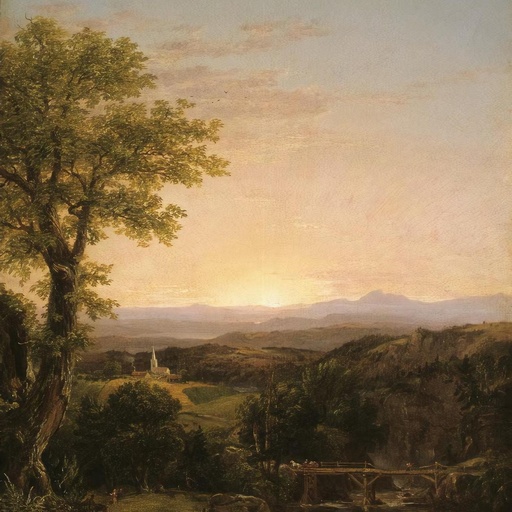}
&
\includegraphics[width=2cm,height=2cm]{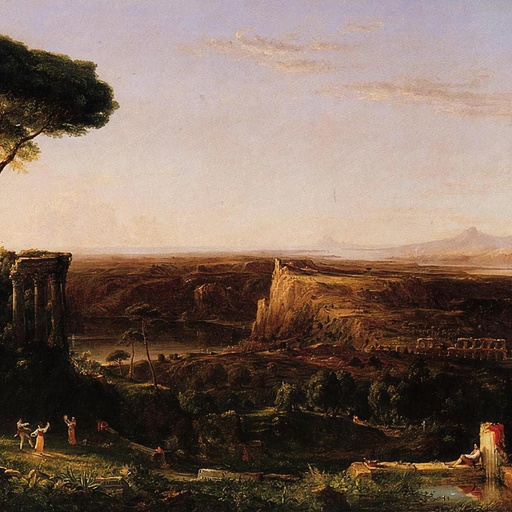} &
\\

&
\raisebox{0pt}[0pt][0pt]{%
\parbox{3.6cm}{\centering
   Technical Drawing\\[3pt]
    \colorbox{gray!20}{\parbox{2.8cm}{\centering Form, Volume, Geometry}}
}}
&
\includegraphics[width=2cm,height=2cm]{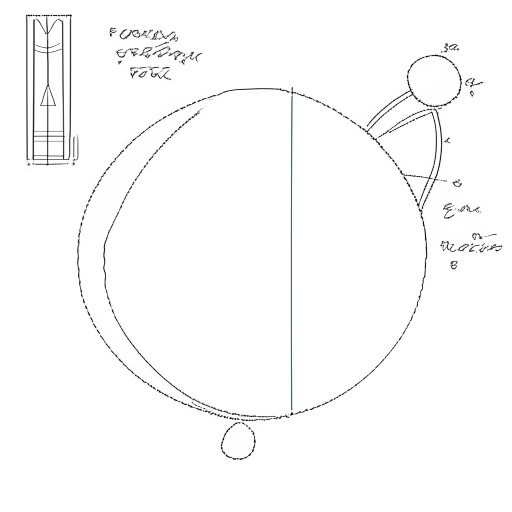}
&
\includegraphics[width=2cm,height=2cm]{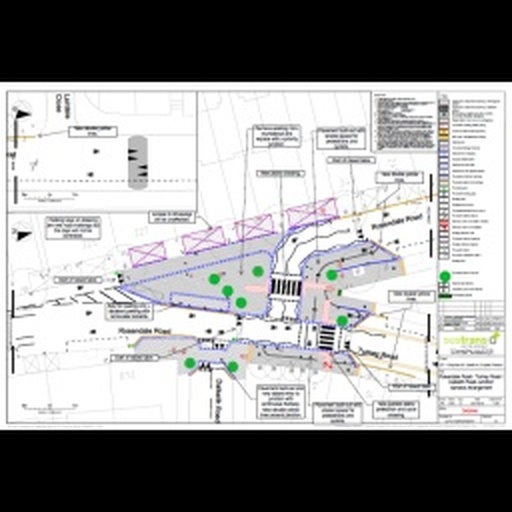}
&
\includegraphics[width=2cm,height=2cm]{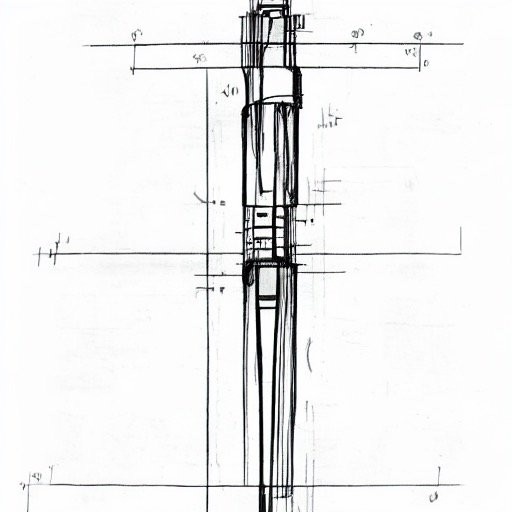}
&
\includegraphics[width=2cm,height=2cm]{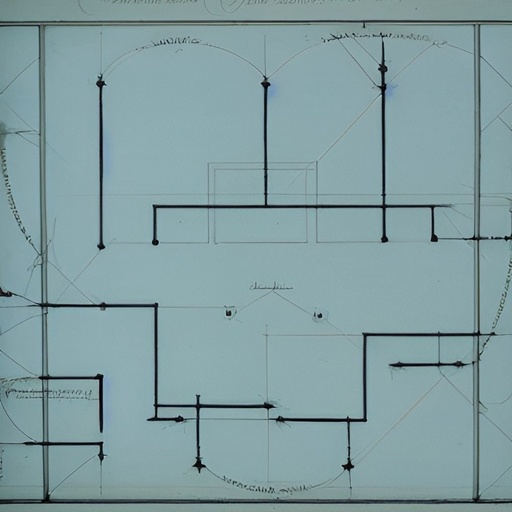}
&
\includegraphics[width=2cm,height=2cm]{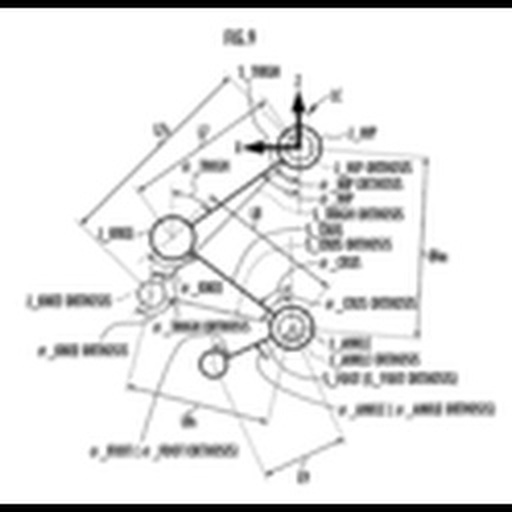} &
\\

&
\raisebox{0pt}[0pt][0pt]{%
\parbox{3.6cm}{\centering
   Polychromatic\\[3pt]
    \colorbox{gray!20}{\parbox{2.8cm}{\centering Color \& Chromatic Qualities}}
}}
&
\includegraphics[width=2cm,height=2cm]{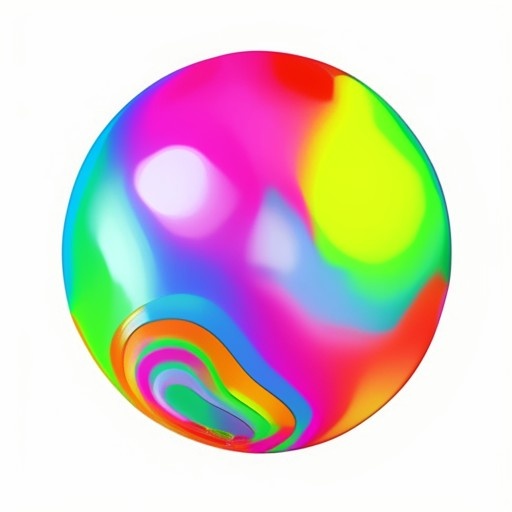}
&
\includegraphics[width=2cm,height=2cm]{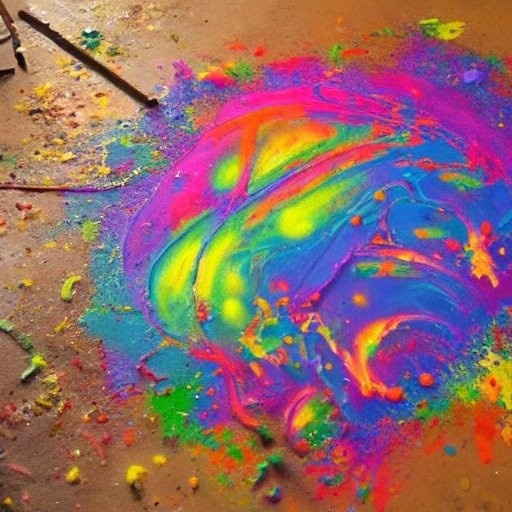}
&
\includegraphics[width=2cm,height=2cm]{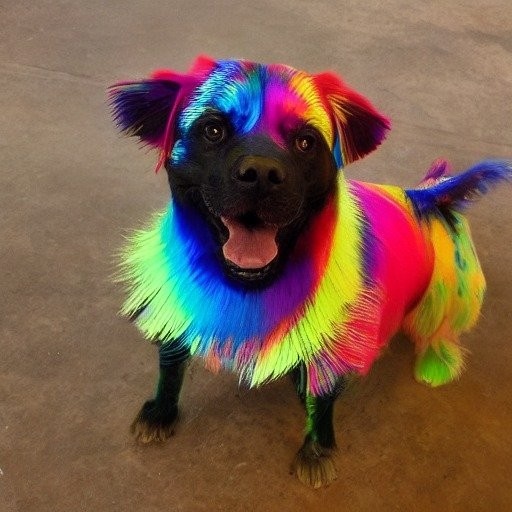}
&
\includegraphics[width=2cm,height=2cm]{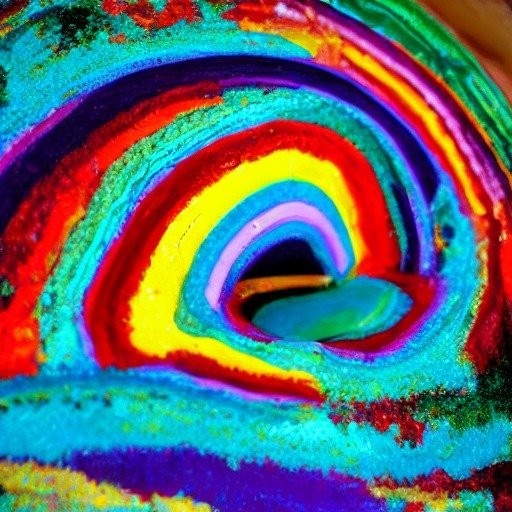}
&
\includegraphics[width=2cm,height=2cm]{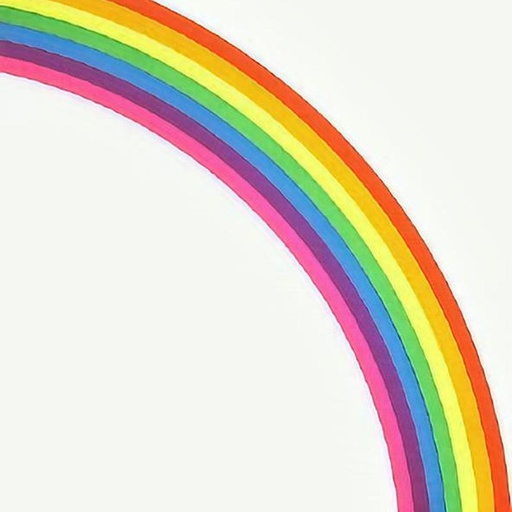} &
\\

\bottomrule
\end{tabular}

\caption{
\textbf{SAE Concept Examples (Part 1/2).} Shown above are the autointerpretability name and label, the prototype visualization, and four representative exemplars for additional concepts learned by \textit{LouvreSAE}-20.
}
\label{fig:more_concepts_part1}
\end{figure*}

\begin{figure*}[h]
\centering
\renewcommand{\arraystretch}{1.05}
\setlength{\tabcolsep}{4pt}

\begin{tabular}{
    >{\centering\arraybackslash}m{0.8cm}
    >{\centering\arraybackslash}m{3.6cm}
    >{\centering\arraybackslash}m{2.0cm}|
    >{\centering\arraybackslash}m{2.0cm}
    >{\centering\arraybackslash}m{2.0cm}
    >{\centering\arraybackslash}m{2.0cm}
    >{\centering\arraybackslash}m{2.0cm}
    >{\centering\arraybackslash}m{0.8cm}
}
\toprule
& \multicolumn{2}{c|}{\textbf{Concept}} &
\multicolumn{4}{c}{\textbf{Representative Exemplars}} & \\
\midrule

&
\raisebox{0pt}[0pt][0pt]{%
\parbox{3.6cm}{\centering
   Curls\\[3pt]
    \colorbox{gray!20}{\parbox{2.8cm}{\centering Form, Volume, Geometry}}
}}
&
\includegraphics[width=2cm,height=2cm]{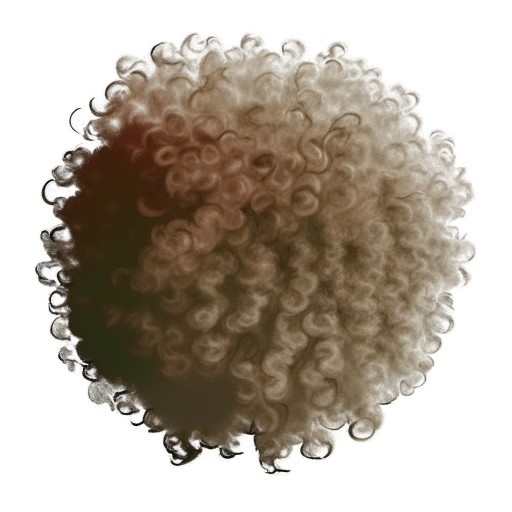}
&
\includegraphics[width=2cm,height=2cm]{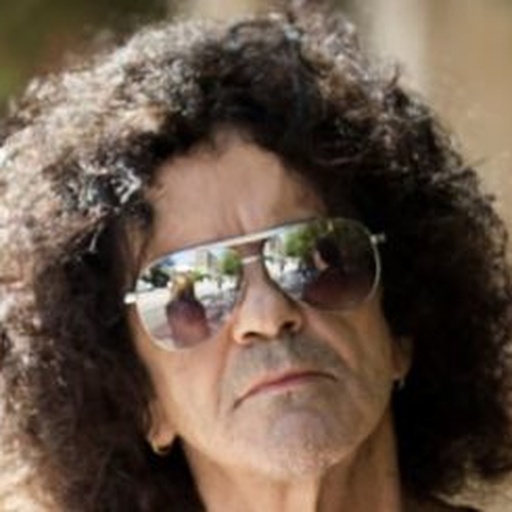}
&
\includegraphics[width=2cm,height=2cm]{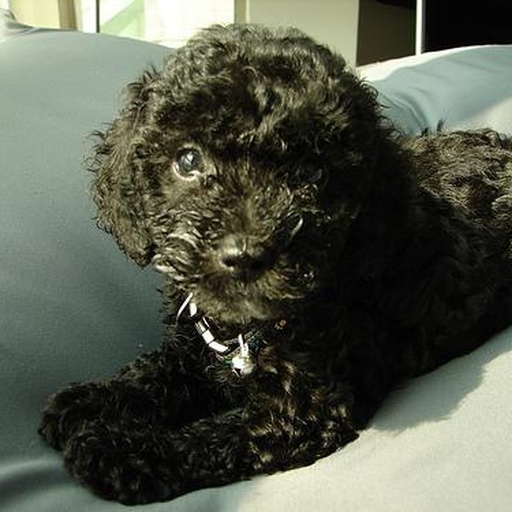}
&
\includegraphics[width=2cm,height=2cm]{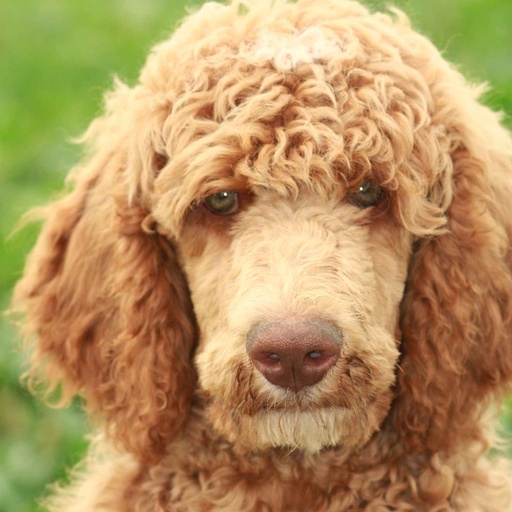}
&
\includegraphics[width=2cm,height=2cm]{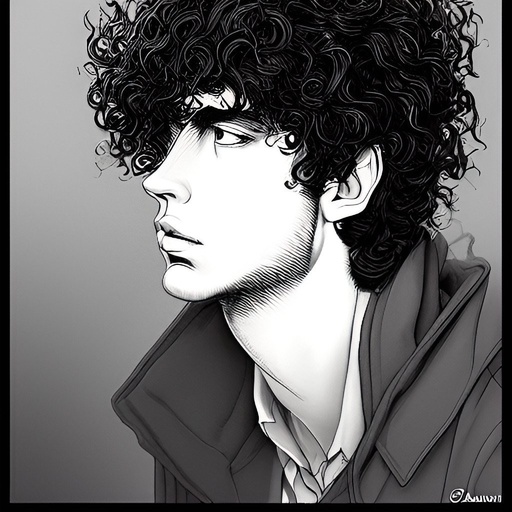} &
\\

&
\raisebox{0pt}[0pt][0pt]{%
\parbox{3.6cm}{\centering
   Specular highlight\\[3pt]
    \colorbox{gray!20}{\parbox{2.8cm}{\centering Reflectivity \& Specular Light Behavior}}
}}
&
\includegraphics[width=2cm,height=2cm]{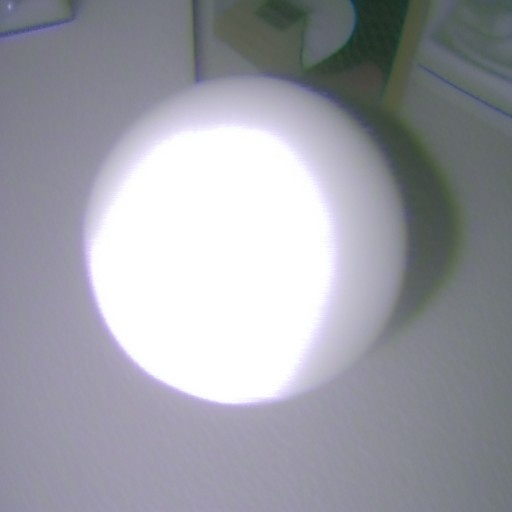}
&
\includegraphics[width=2cm,height=2cm]{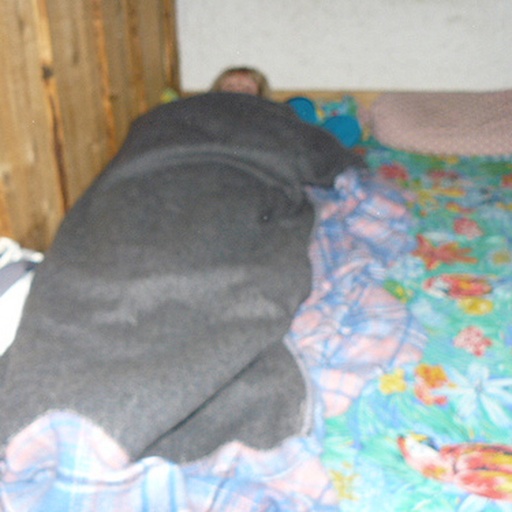}
&
\includegraphics[width=2cm,height=2cm]{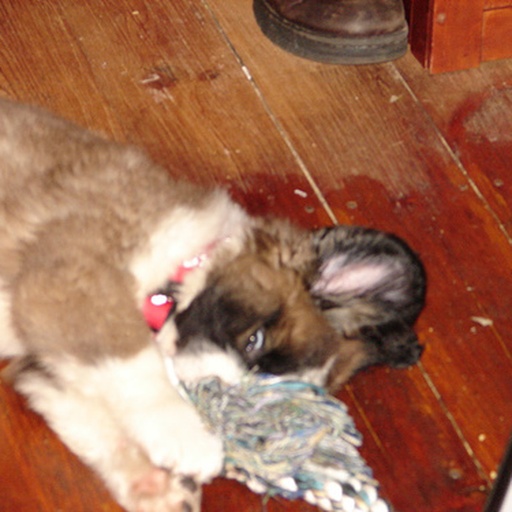}
&
\includegraphics[width=2cm,height=2cm]{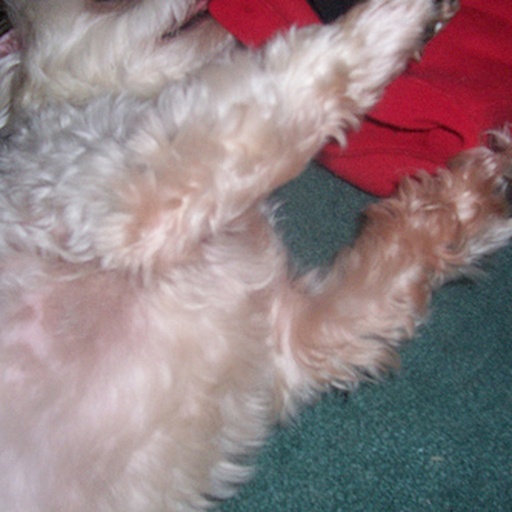}
&
\includegraphics[width=2cm,height=2cm]{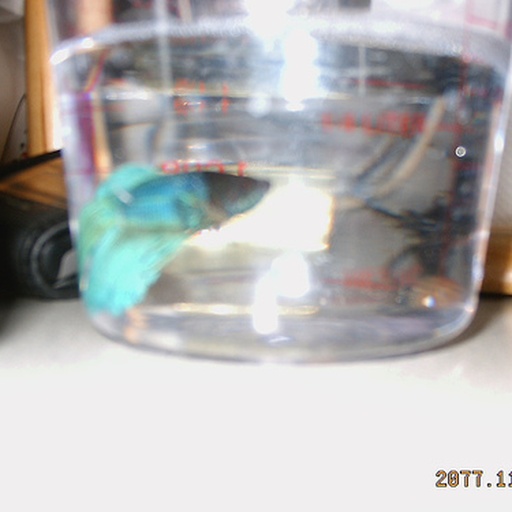} &
\\

&
\raisebox{0pt}[0pt][0pt]{%
\parbox{3.6cm}{\centering
   Horse\\[3pt]
    \colorbox{gray!20}{\parbox{2.8cm}{\centering Objects \& Semantic Concepts}}
}}
&
\includegraphics[width=2cm,height=2cm]{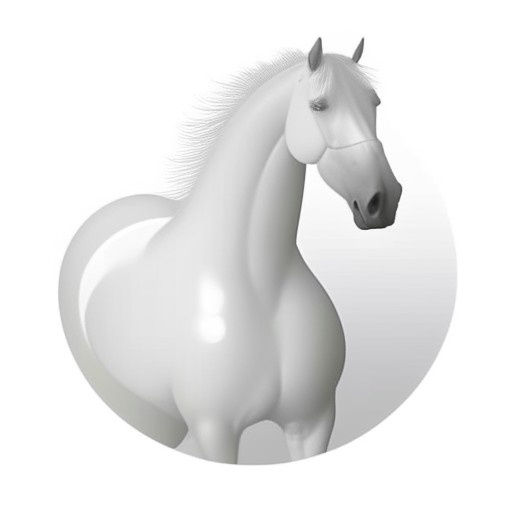}
&
\includegraphics[width=2cm,height=2cm]{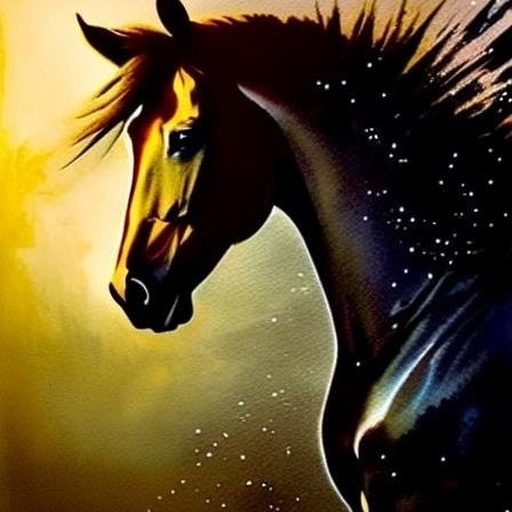}
&
\includegraphics[width=2cm,height=2cm]{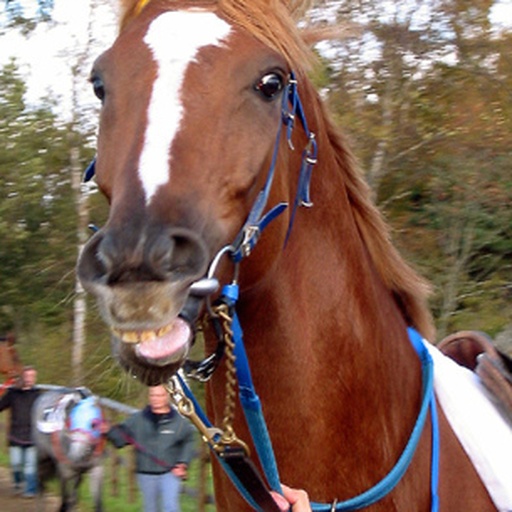}
&
\includegraphics[width=2cm,height=2cm]{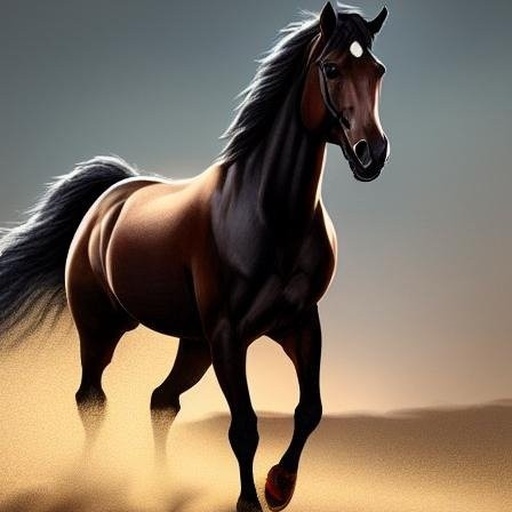}
&
\includegraphics[width=2cm,height=2cm]{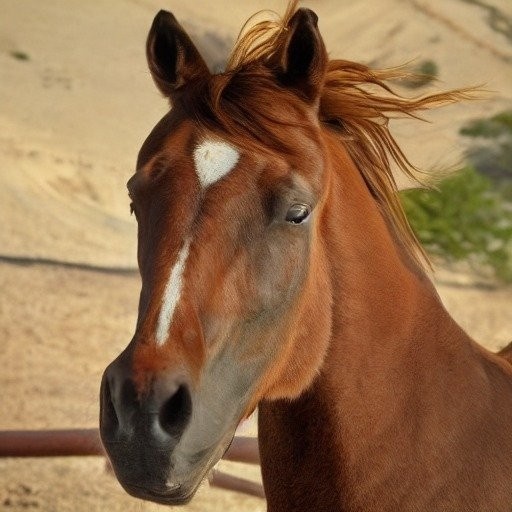} &
\\

&
\raisebox{0pt}[0pt][0pt]{%
\parbox{3.6cm}{\centering
   Glassy\\[3pt]
    \colorbox{gray!20}{\parbox{2.8cm}{\centering Texture \& Material Properties}}
}}
&
\includegraphics[width=2cm,height=2cm]{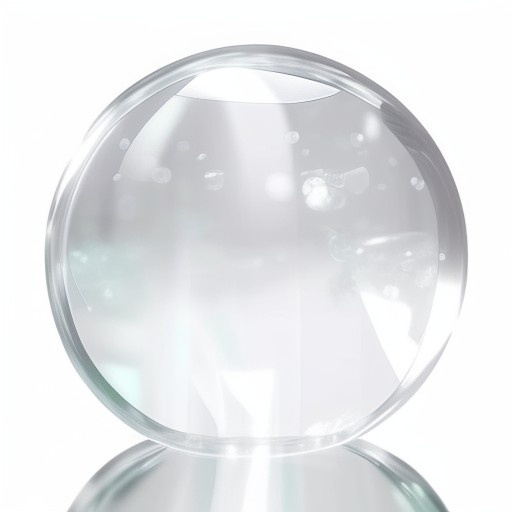}
&
\includegraphics[width=2cm,height=2cm]{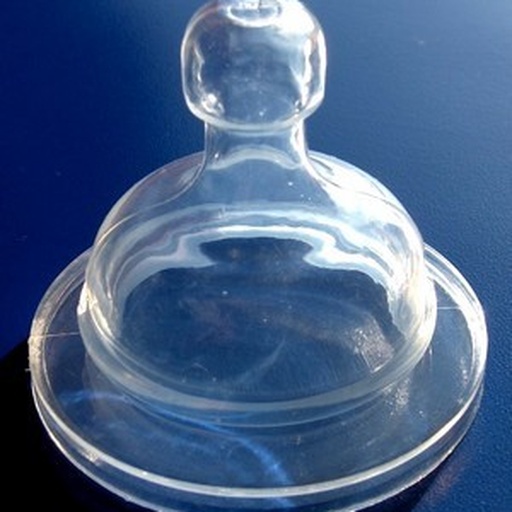}
&
\includegraphics[width=2cm,height=2cm]{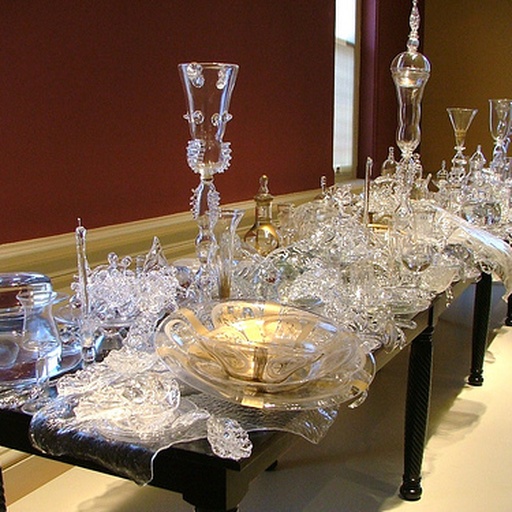}
&
\includegraphics[width=2cm,height=2cm]{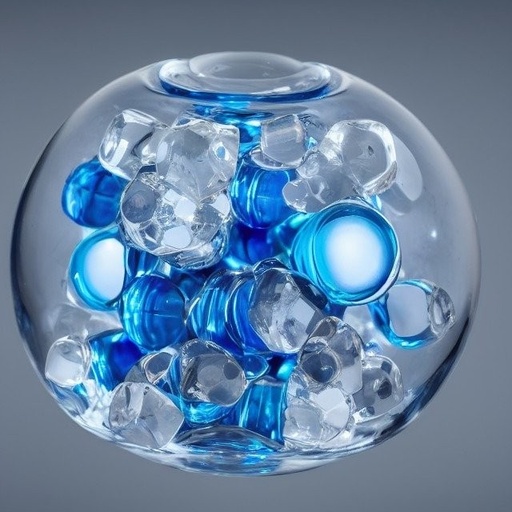}
&
\includegraphics[width=2cm,height=2cm]{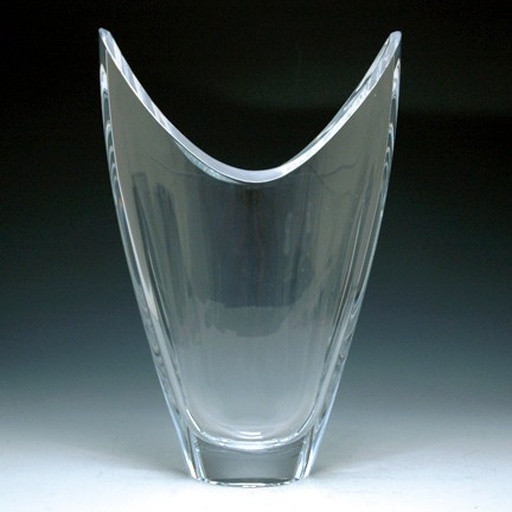} &
\\

\bottomrule
\end{tabular}

\caption{
\textbf{SAE Concept Examples (Part 2/2).} Shown above are the autointerpretability name and label, the prototype visualization, and four representative exemplars for additional concepts learned by \textit{LouvreSAE}-20.
}
\label{fig:more_concepts_part2}
\end{figure*}

\FloatBarrier

\newpage
~
\newpage
\section{Additional Results: SAE Concept Steering}
\label{app:tbd}

\begin{figure*}[h]
    \centering
    \begin{tabular}{
        >{\centering\arraybackslash}m{0.2cm}
        >{\centering\arraybackslash}m{3.6cm}
        >{\centering\arraybackslash}m{2.0cm}|
        >{\centering\arraybackslash}m{2.0cm}
        >{\centering\arraybackslash}m{2.0cm}
        >{\centering\arraybackslash}m{2.0cm}
        >{\centering\arraybackslash}m{2.0cm}
        >{\centering\arraybackslash}m{0.2cm}
    }
    \toprule
    &
    \multicolumn{2}{c|}{\textbf{Concept}} &
    $\boldsymbol{\alpha_1}$ &
    $\boldsymbol{\alpha_2}$ &
    $\boldsymbol{\alpha_3}$ &
    $\boldsymbol{\alpha_4}$ & \\
    \midrule

    &
    \begin{minipage}{3.6cm}\centering
        Water reflection\\[3pt]
        \colorbox{gray!20}{
            \parbox{2.8cm}{\centering Reflectivity \& Specular Light Behavior}
        }
    \end{minipage}
    &
    \includegraphics[width=2cm,height=2cm]{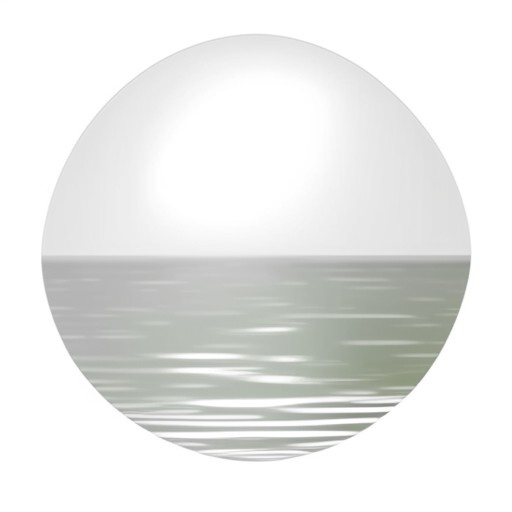} &
    \includegraphics[width=2cm,height=2cm]{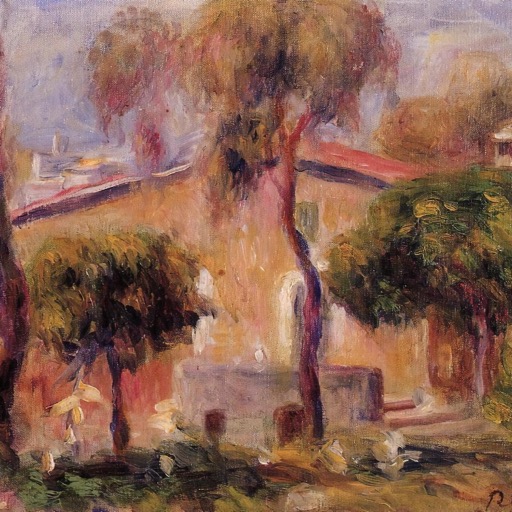} &
    \includegraphics[width=2cm,height=2cm]{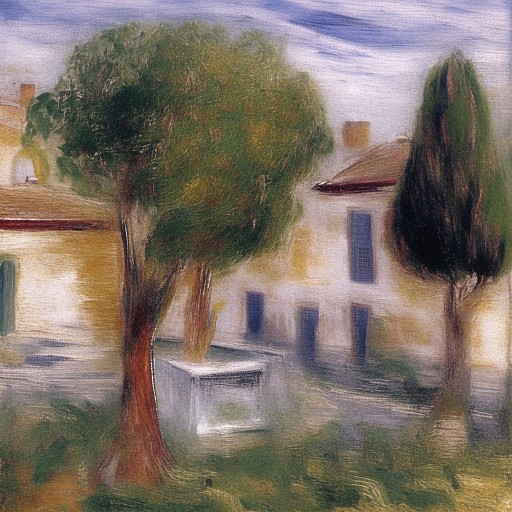} &
    \includegraphics[width=2cm,height=2cm]{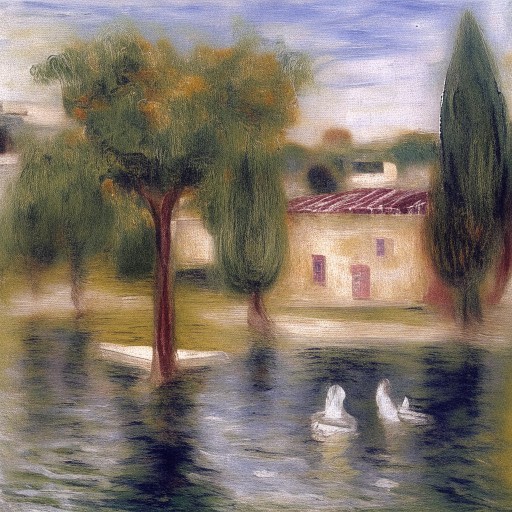} &
    \includegraphics[width=2cm,height=2cm]{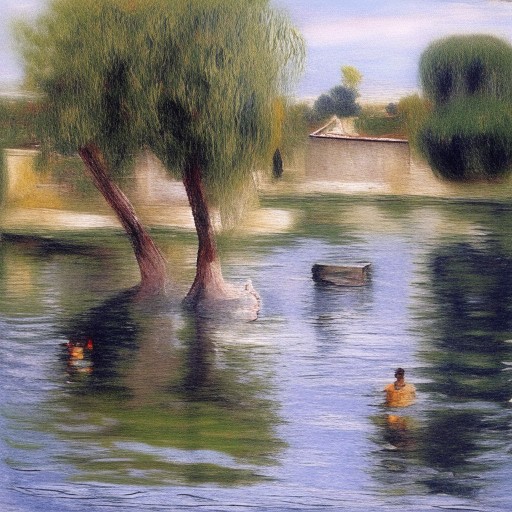} & \\[0.4cm]

    &
    \begin{minipage}{3.6cm}\centering
        Pink\\[3pt]
        \colorbox{gray!20}{
            \parbox{2.8cm}{\centering Color \& Chromatic Qualities}
        }
    \end{minipage}
    &
    \includegraphics[width=2cm,height=2cm]{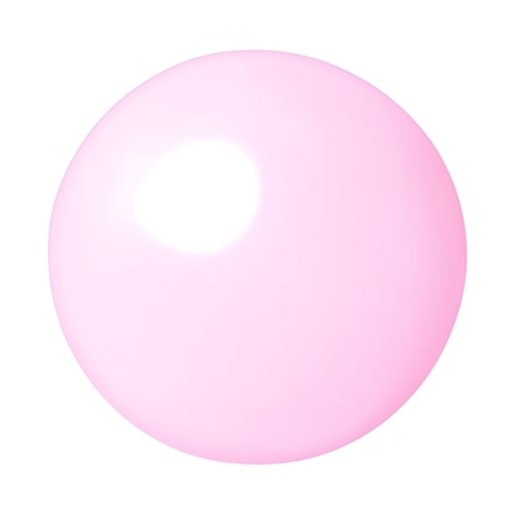} &
    \includegraphics[width=2cm,height=2cm]{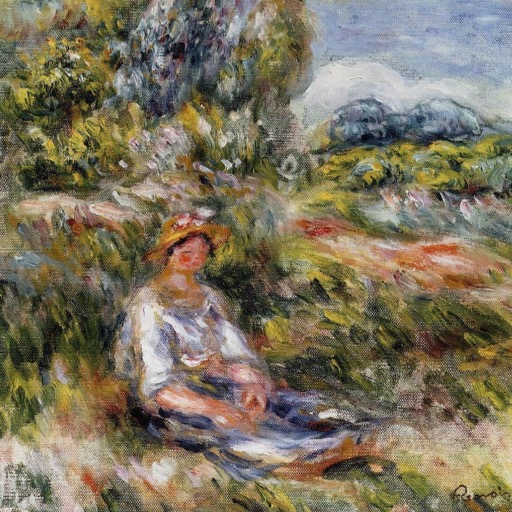} &
    \includegraphics[width=2cm,height=2cm]{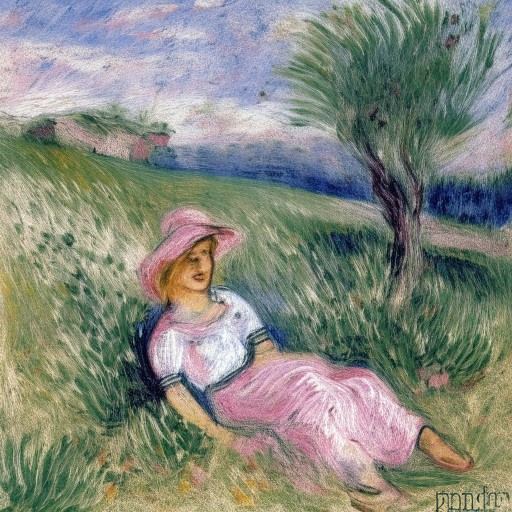} &
    \includegraphics[width=2cm,height=2cm]{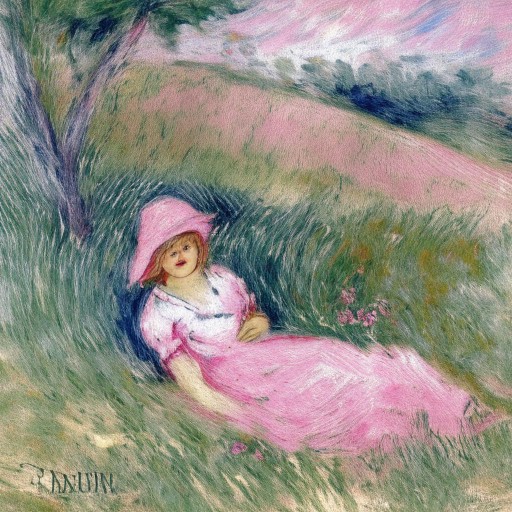} &
    \includegraphics[width=2cm,height=2cm]{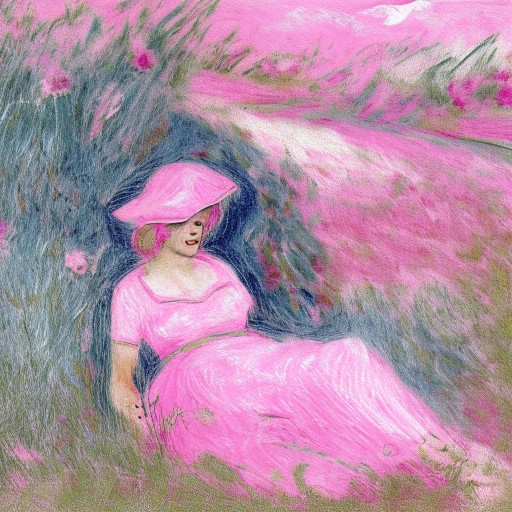} & \\[0.4cm]

    &
    \begin{minipage}{3.6cm}\centering
        Faceted\\[3pt]
        \colorbox{gray!20}{
            \parbox{2.8cm}{\centering Form, Volume, Geometry}
        }
    \end{minipage}
    &
    \includegraphics[width=2cm,height=2cm]{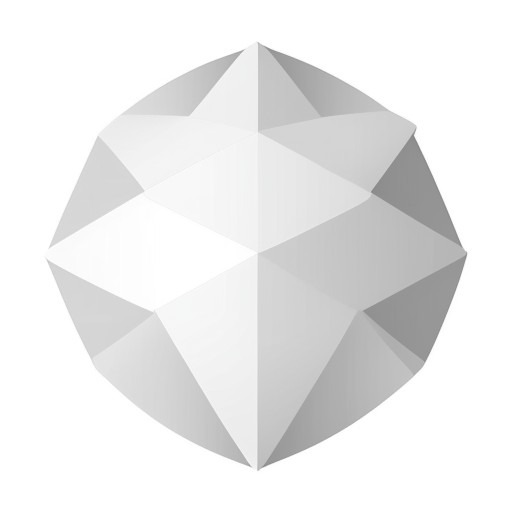} &
    \includegraphics[width=2cm,height=2cm]{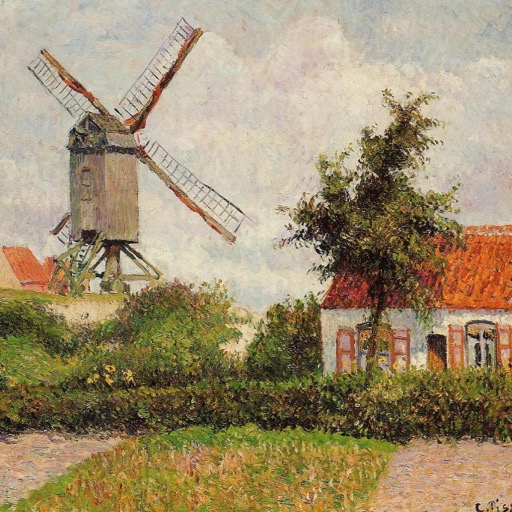} &
    \includegraphics[width=2cm,height=2cm]{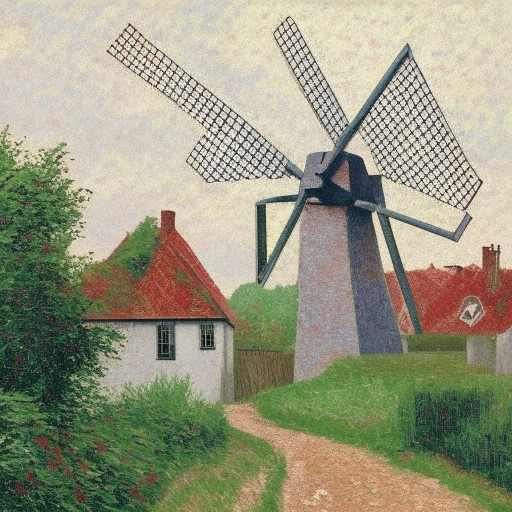} &
    \includegraphics[width=2cm,height=2cm]{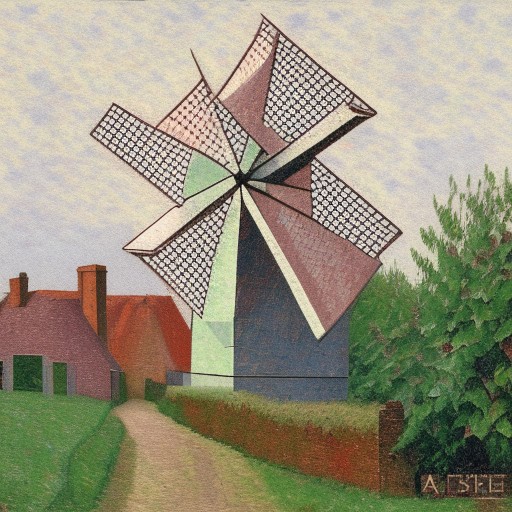} &
    \includegraphics[width=2cm,height=2cm]{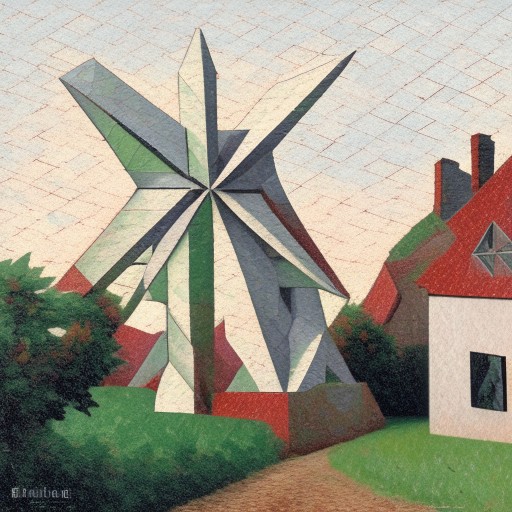} & \\[0.4cm]

    &
    \begin{minipage}{3.6cm}\centering
        Flat Design\\[3pt]
        \colorbox{gray!20}{
            \parbox{2.8cm}{\centering Style \& Cultural / Artistic Aesthetics}
        }
    \end{minipage}
    &
    \includegraphics[width=2cm,height=2cm]{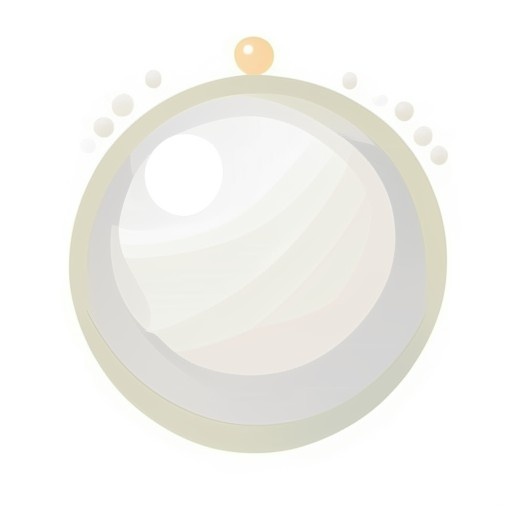} &
    \includegraphics[width=2cm,height=2cm]{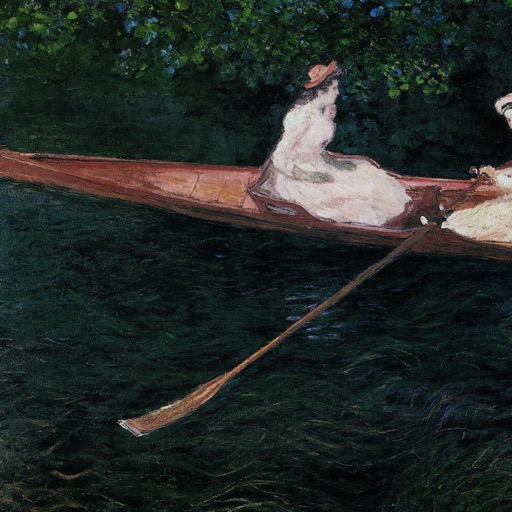} &
    \includegraphics[width=2cm,height=2cm]{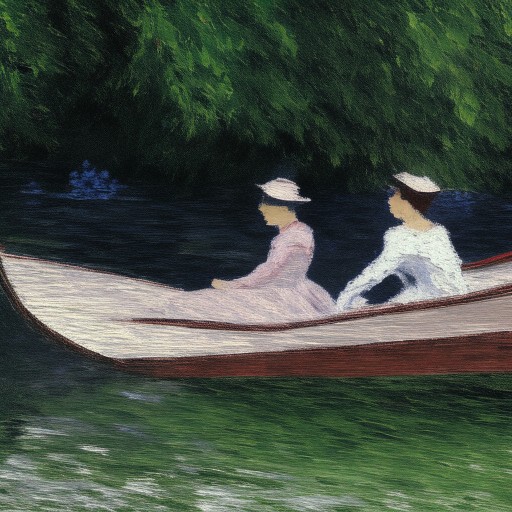} &
    \includegraphics[width=2cm,height=2cm]{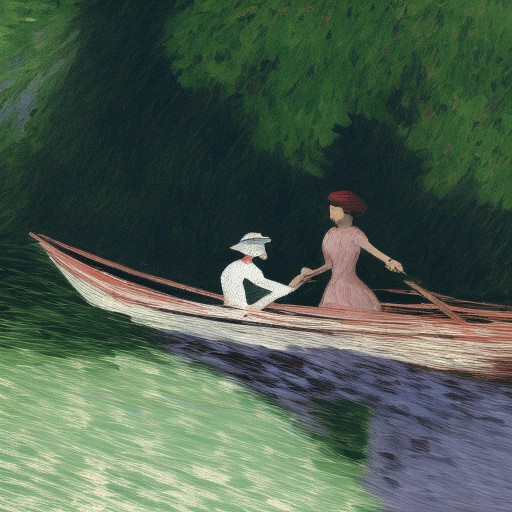} &
    \includegraphics[width=2cm,height=2cm]{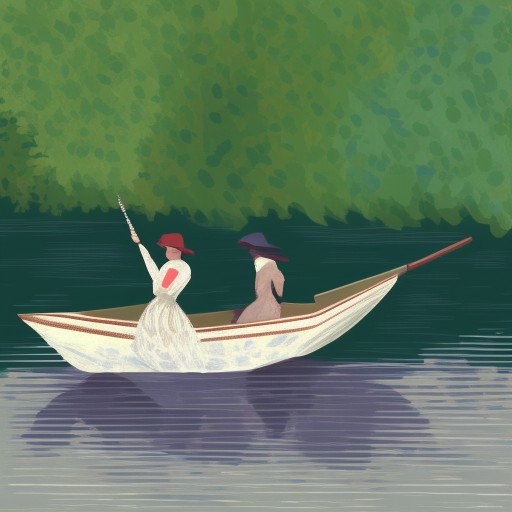} & \\[0.4cm]

    &
    \begin{minipage}{3.6cm}\centering
        Technical Drawing\\[3pt]
        \colorbox{gray!20}{
            \parbox{2.8cm}{\centering Form, Volume, Geometry}
        }
    \end{minipage}
    &
    \includegraphics[width=2cm,height=2cm]{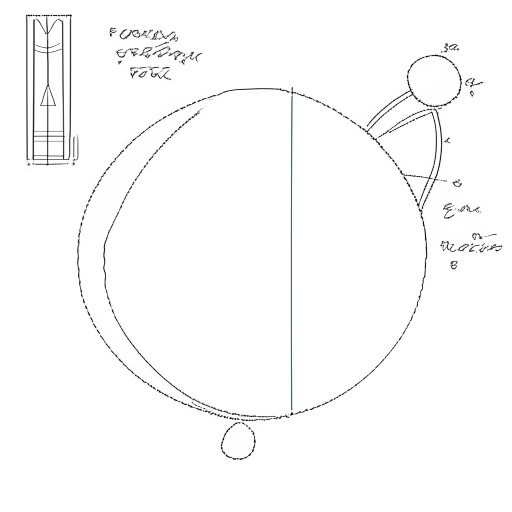} &
    \includegraphics[width=2cm,height=2cm]{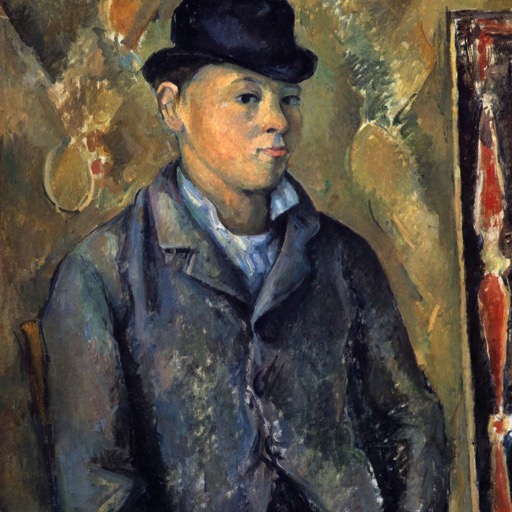} &
    \includegraphics[width=2cm,height=2cm]{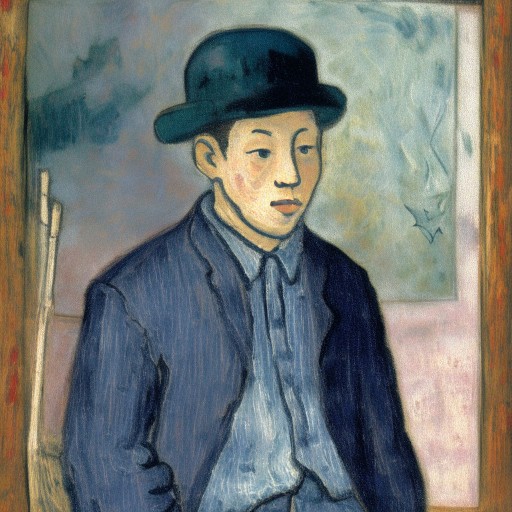} &
    \includegraphics[width=2cm,height=2cm]{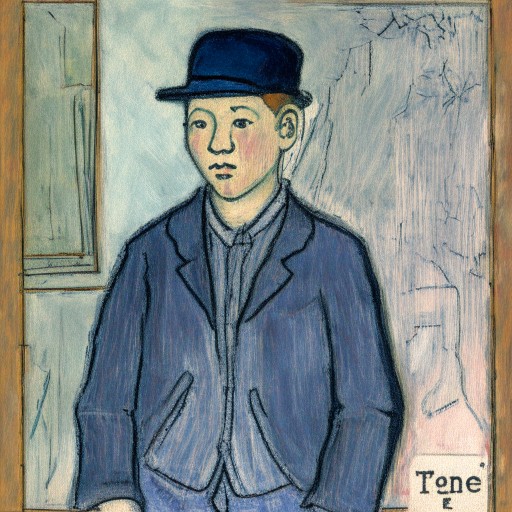} &
    \includegraphics[width=2cm,height=2cm]{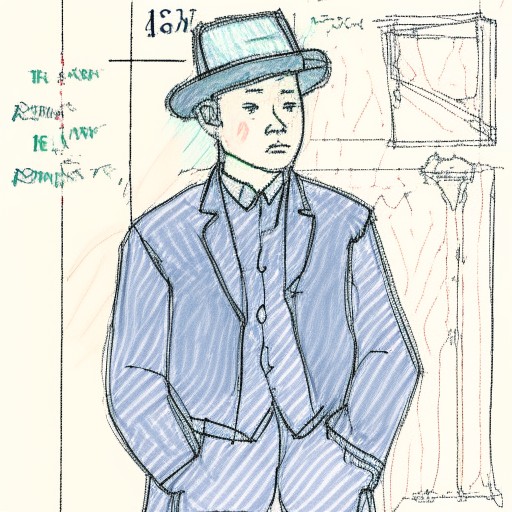} & \\[0.4cm]

    &
    \begin{minipage}{3.6cm}\centering
        Curls\\[3pt]
        \colorbox{gray!20}{
            \parbox{2.8cm}{\centering Form, Volume, Geometry}
        }
    \end{minipage}
    &
    \includegraphics[width=2cm,height=2cm]{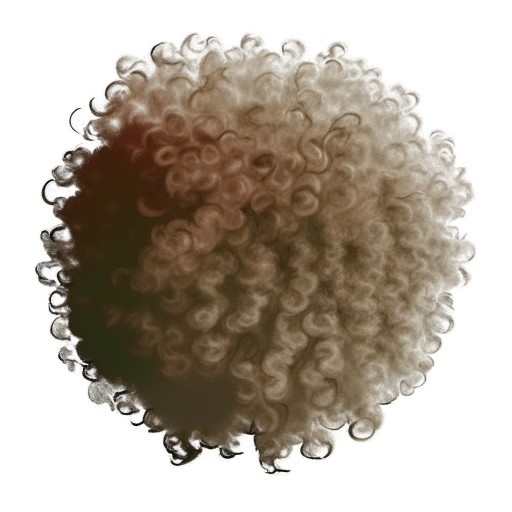} &
    \includegraphics[width=2cm,height=2cm]{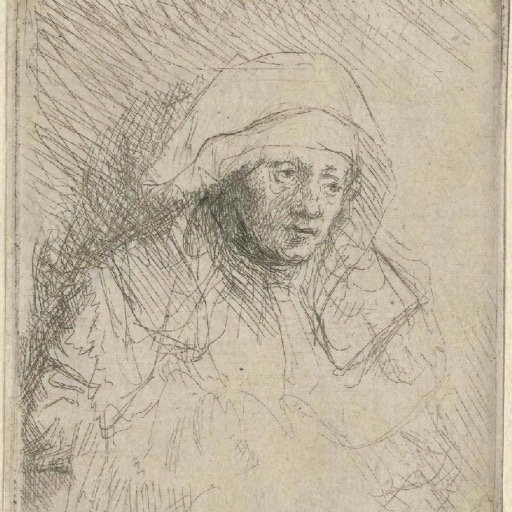} &
    \includegraphics[width=2cm,height=2cm]{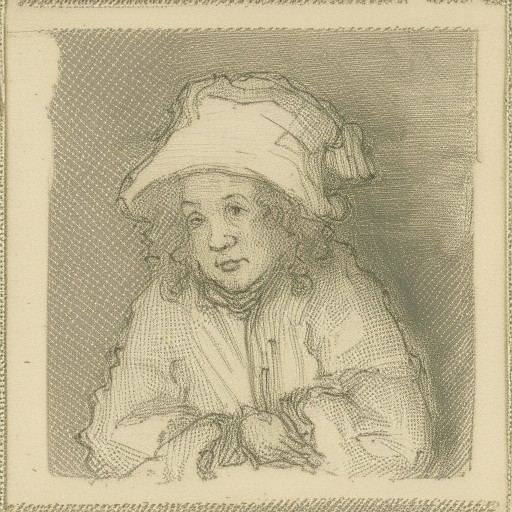} &
    \includegraphics[width=2cm,height=2cm]{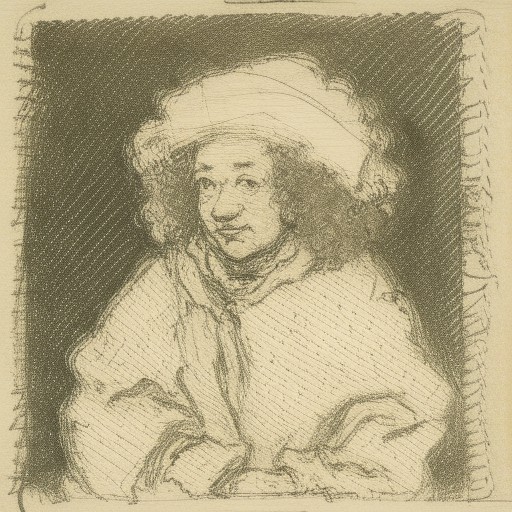} &
    \includegraphics[width=2cm,height=2cm]{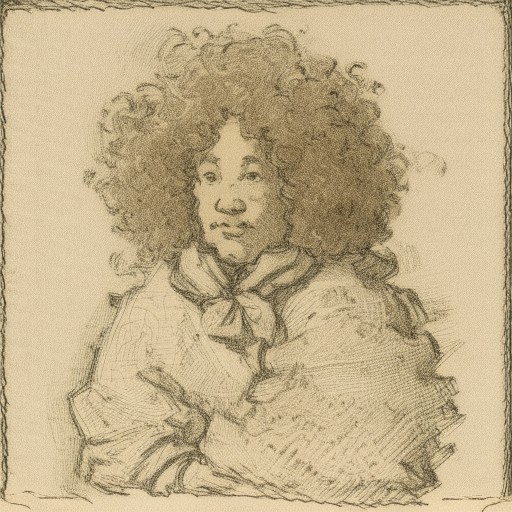} & \\[0.4cm]

    &
    \begin{minipage}{3.6cm}\centering
        Radiance\\[3pt]
        \colorbox{gray!20}{
            \parbox{2.8cm}{\centering Reflectivity \& Specular Light Behavior}
        }
    \end{minipage}
    &
    \includegraphics[width=2cm,height=2cm]{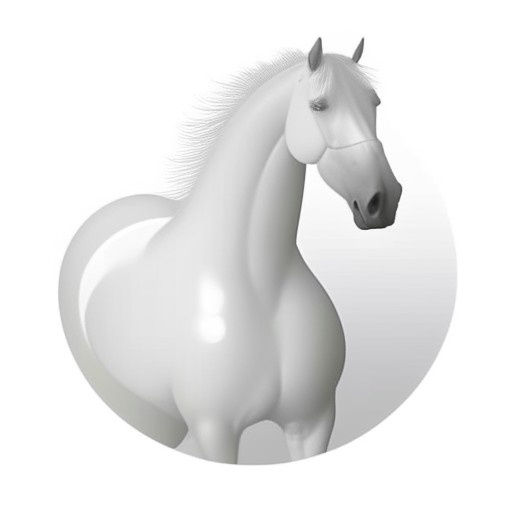} &
    \includegraphics[width=2cm,height=2cm]{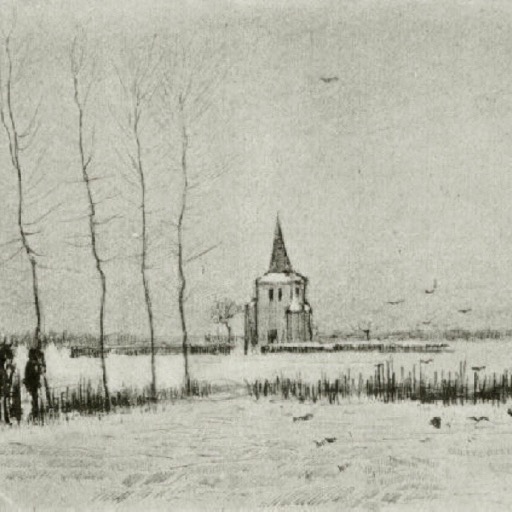} &
    \includegraphics[width=2cm,height=2cm]{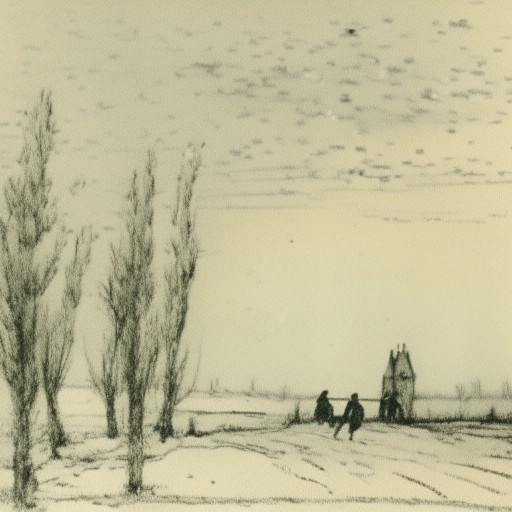} &
    \includegraphics[width=2cm,height=2cm]{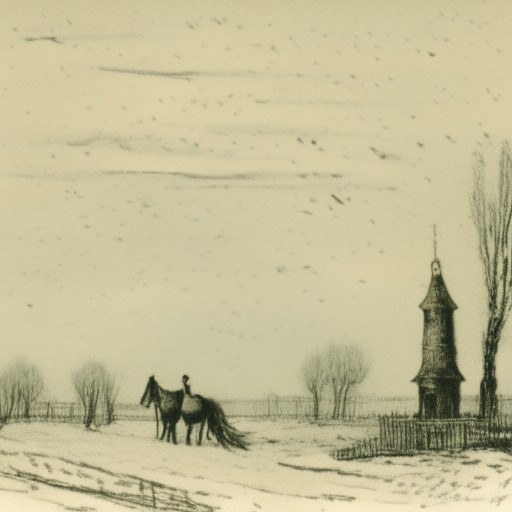} &
    \includegraphics[width=2cm,height=2cm]{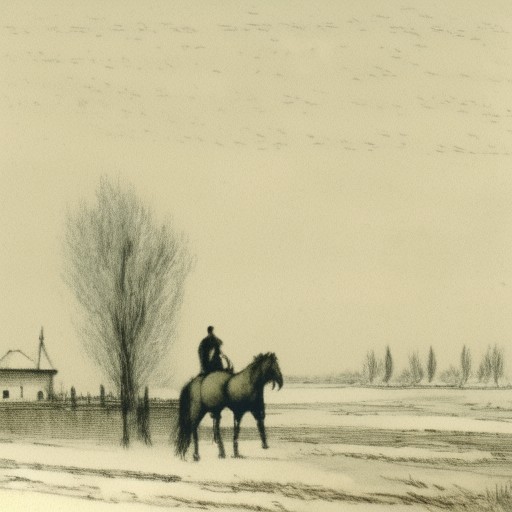} & \\[0.4cm]

    &
    \begin{minipage}{3.6cm}\centering
        Watercolor\\[3pt]
        \colorbox{gray!20}{
            \parbox{2.8cm}{\centering Style \& Cultural / Artistic Aesthetics}
        }
    \end{minipage}
    &
    \includegraphics[width=2cm,height=2cm]{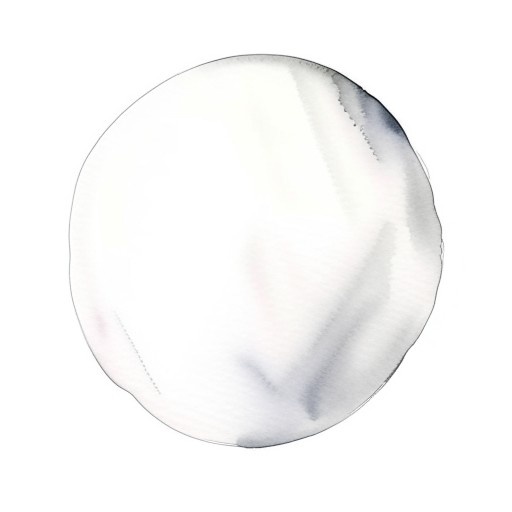} &
    \includegraphics[width=2cm,height=2cm]{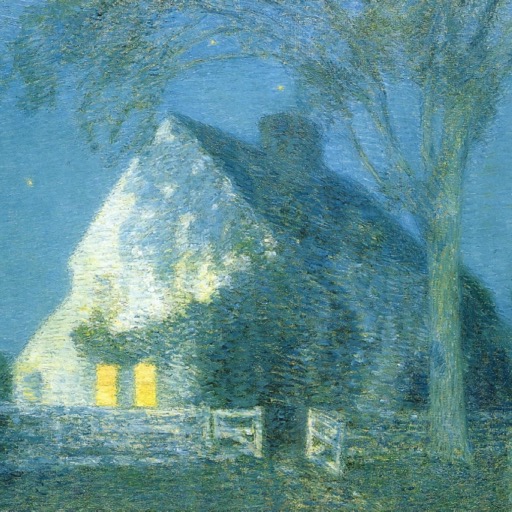} &
    \includegraphics[width=2cm,height=2cm]{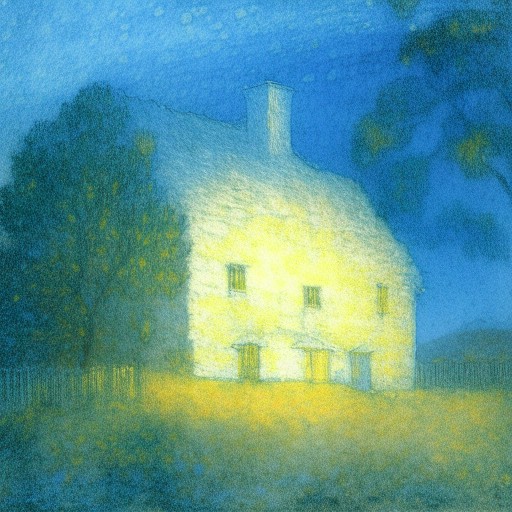} &
    \includegraphics[width=2cm,height=2cm]{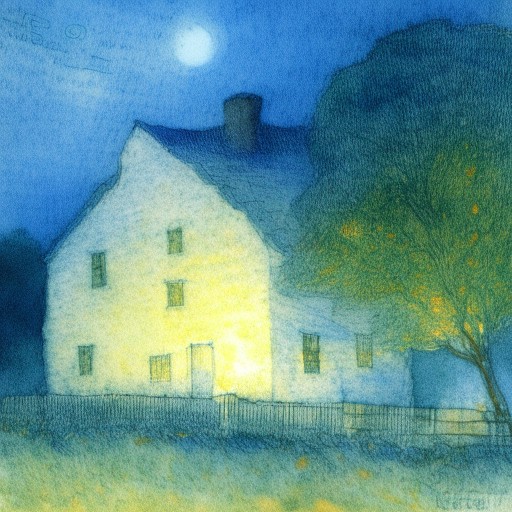} &
    \includegraphics[width=2cm,height=2cm]{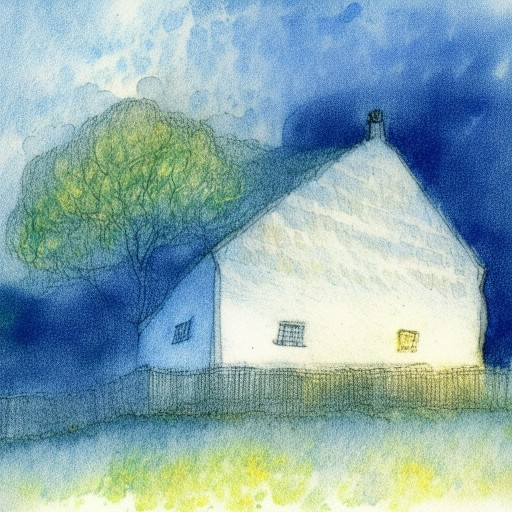} & \\

    \bottomrule
    \end{tabular}

    \caption{\textbf{Additional Results: SAE Concept Steering.} Increasing steering intensity governed by the $\alpha$ coefficient is shown above for a variety of discovered concepts. $\alpha$ is scaled relative to the natural activation intensity of each concept in the training dataset.}
    \label{fig:qual_concept_steer}
\end{figure*}

\FloatBarrier

\newpage
\section{Additional Results: Qualitative Examples}
\label{app:tbd}

\begin{figure*}[h]
    \centering
    \begin{tabular}{
        >{\centering\arraybackslash}m{2.4cm}|
        >{\centering\arraybackslash}m{2.0cm}
        >{\centering\arraybackslash}m{2.0cm}|
        >{\centering\arraybackslash}m{2.0cm}
        >{\centering\arraybackslash}m{2.0cm}|
        >{\centering\arraybackslash}m{2.0cm}
        >{\centering\arraybackslash}m{2.0cm}
    }
    \toprule
    \shortstack{\textbf{Style}} &
    \multicolumn{2}{c|}{\shortstack{\textbf{Reference Samples}}} &
    \multicolumn{2}{c|}{\shortstack{\textbf{Content Images}}} &
    \multicolumn{2}{c}{\shortstack{\textbf{Generated Images}}} \\
    \midrule
    
    \multirow{1}{*}{\centering\shortstack{Camille\\Pissarro}} &
    \includegraphics[width=2cm,height=2cm]{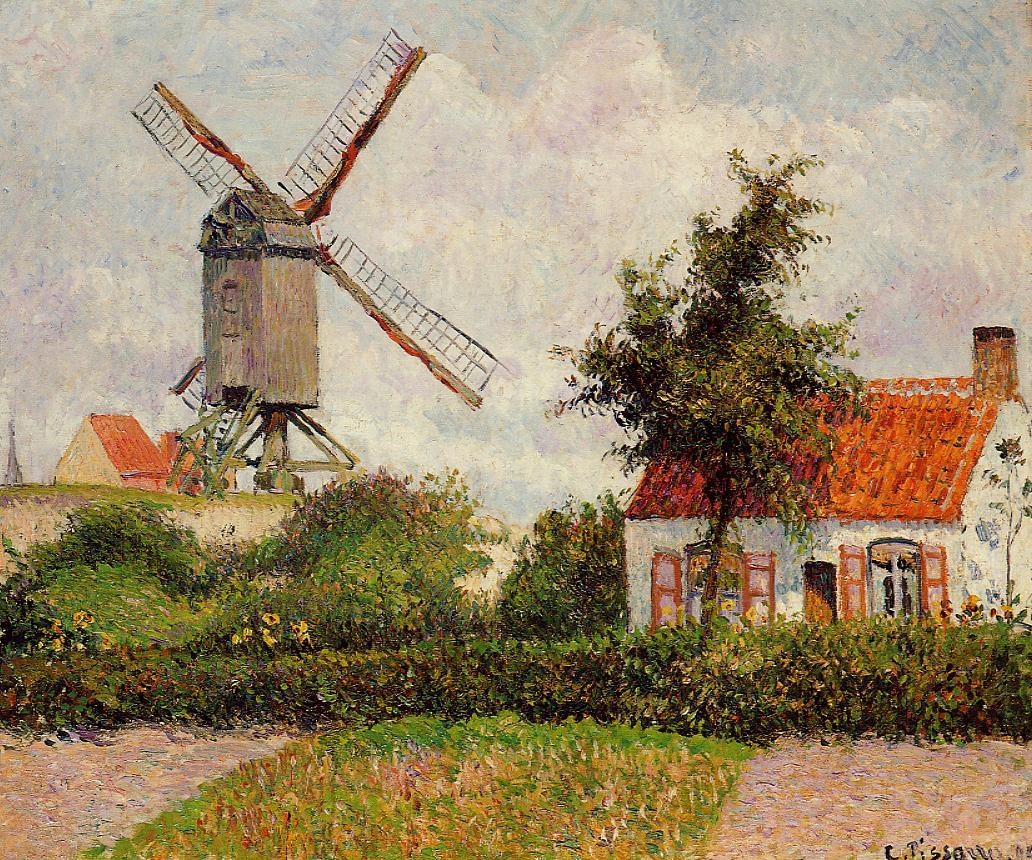} &
    \includegraphics[width=2cm,height=2cm]{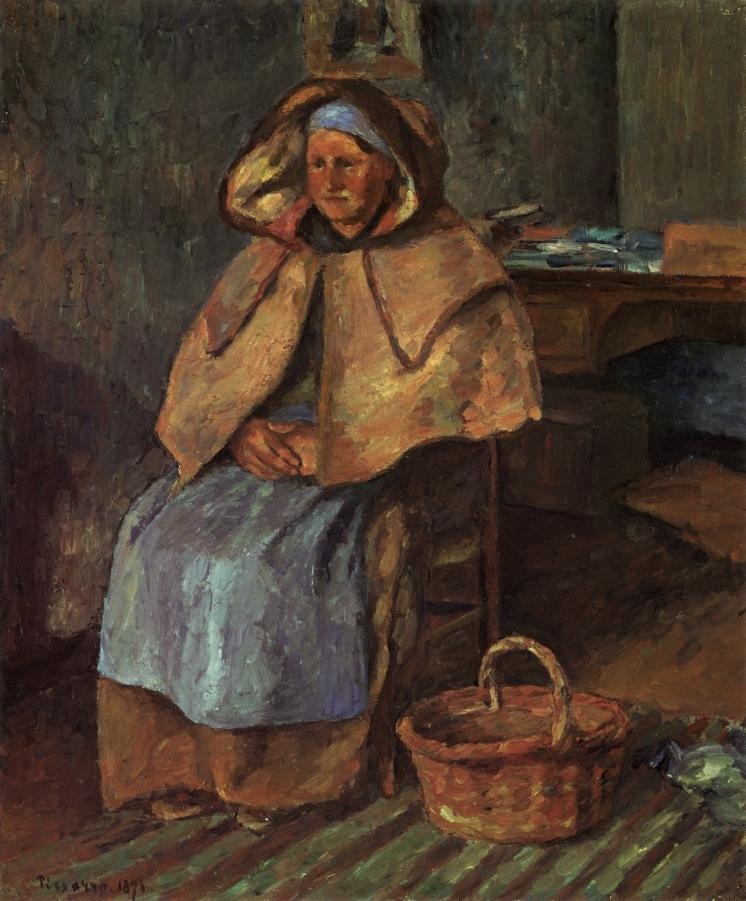} &
    \includegraphics[width=2cm,height=2cm]{fig/fig_11/camille/000091_original_image.jpg} &
    \includegraphics[width=2cm,height=2cm]{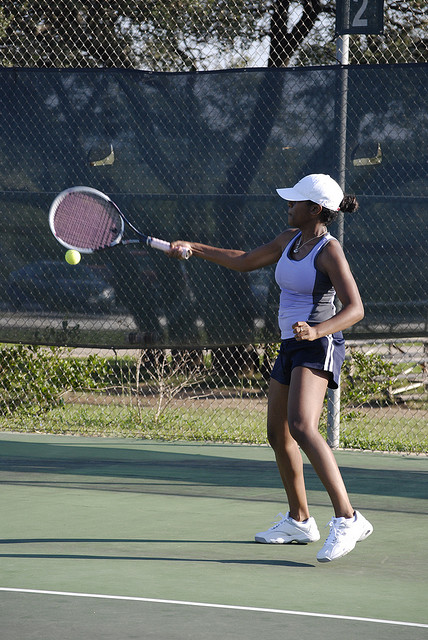} &
    \includegraphics[width=2cm,height=2cm]{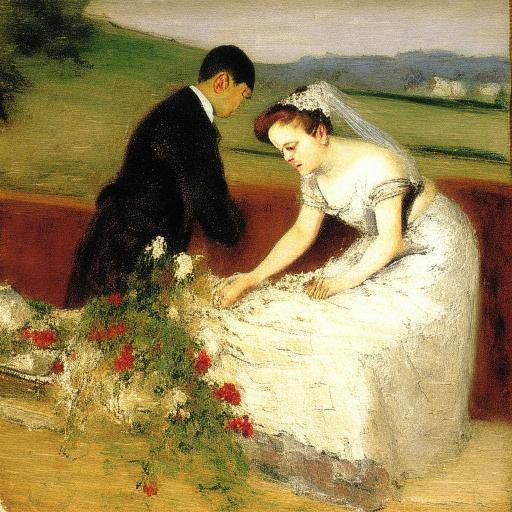} &
    \includegraphics[width=2cm,height=2cm]{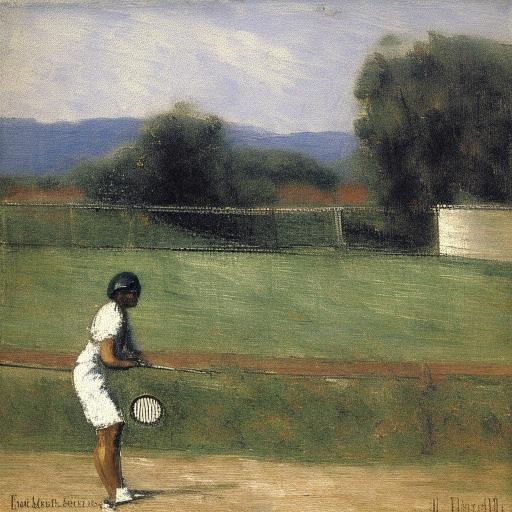} \\[0.4cm]

    \multirow{1}{*}{\centering El Greco} &
    \includegraphics[width=2cm,height=2cm]{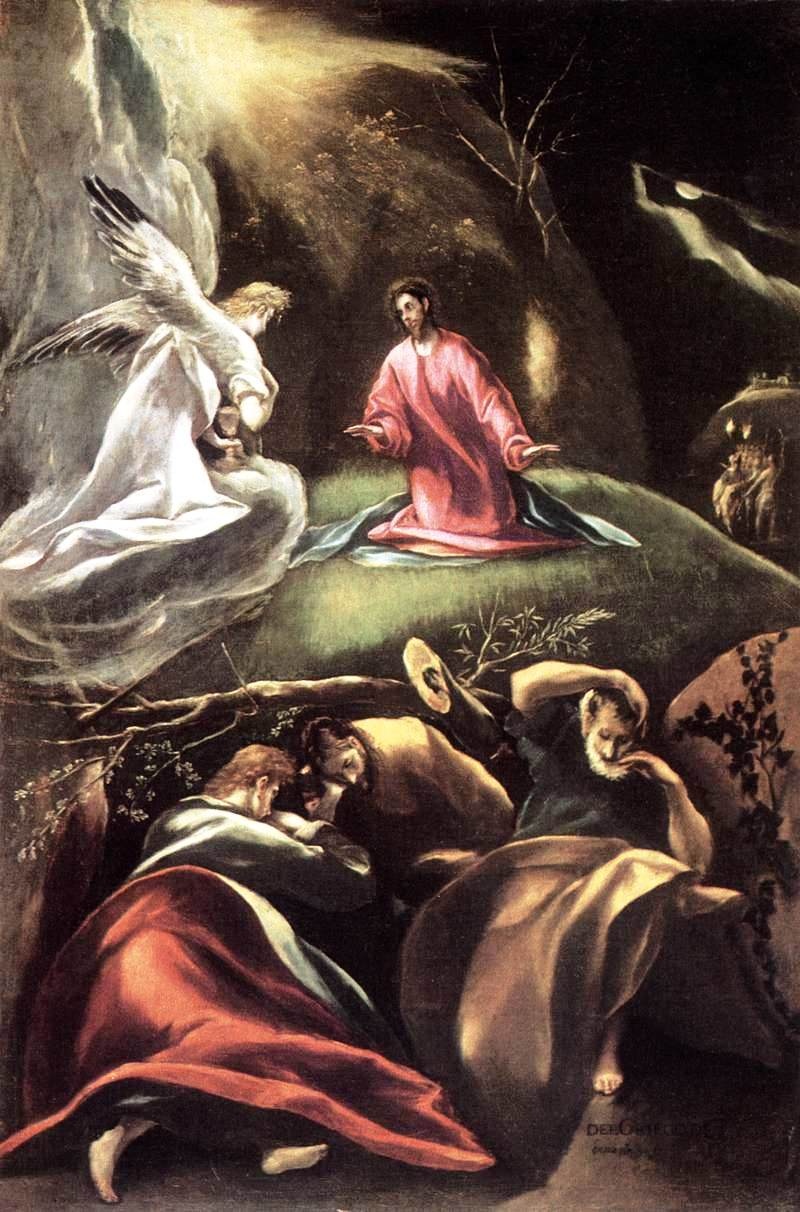} &
    \includegraphics[width=2cm,height=2cm]{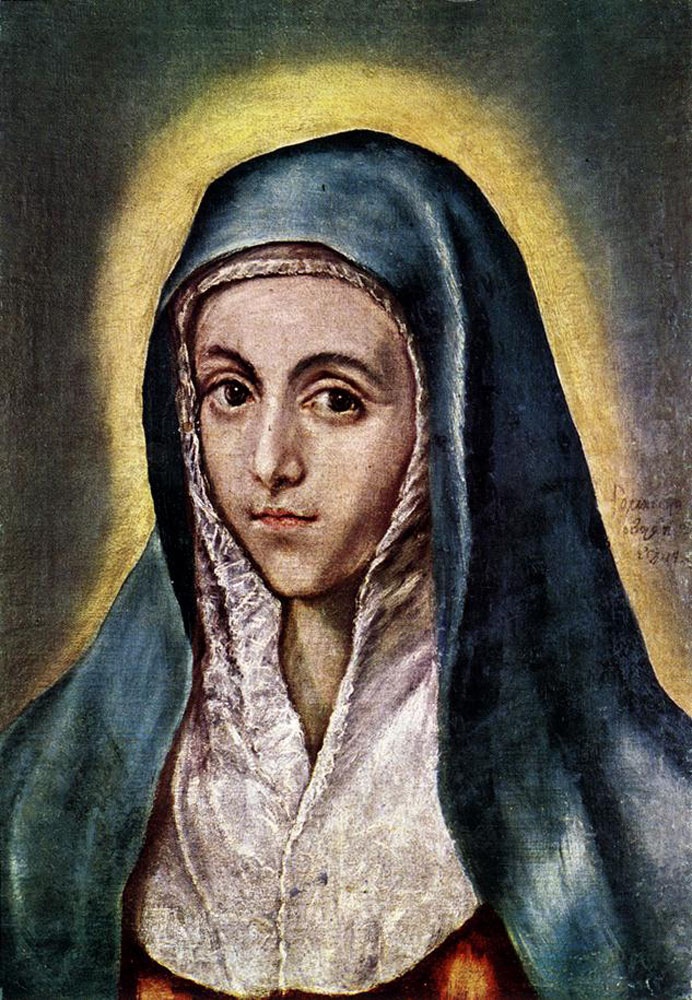} &
    \includegraphics[width=2cm,height=2cm]{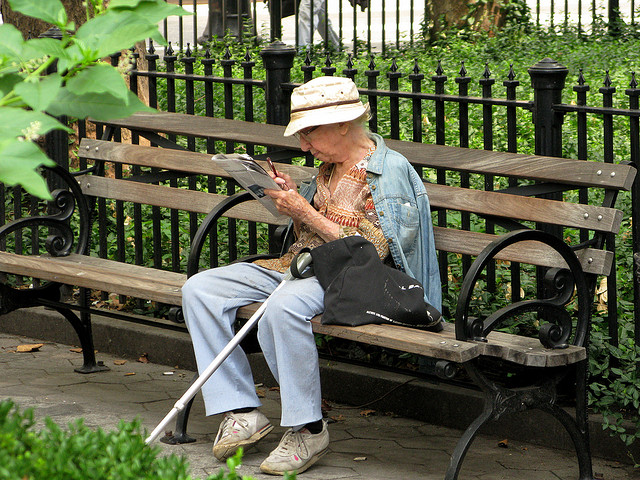} &
    \includegraphics[width=2cm,height=2cm]{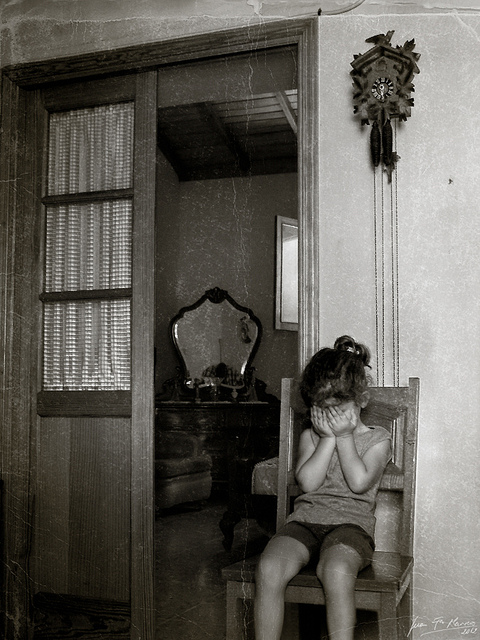} &
    \includegraphics[width=2cm,height=2cm]{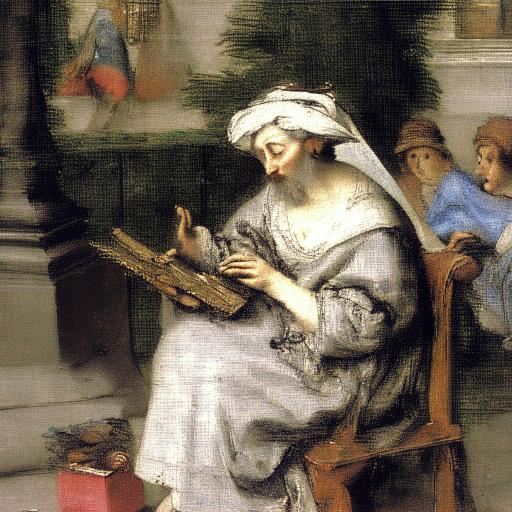} &
    \includegraphics[width=2cm,height=2cm]{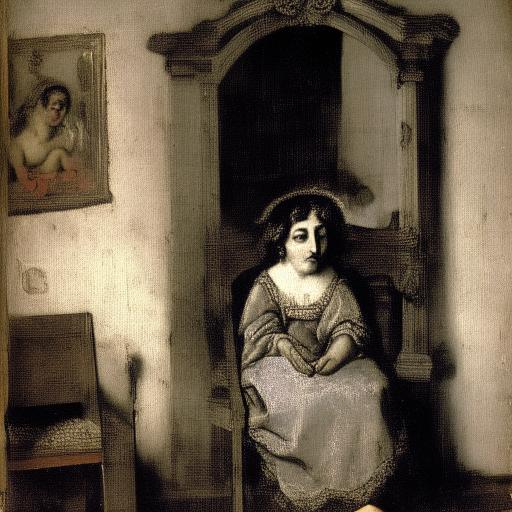} \\[0.4cm]

    \multirow{1}{*}{\centering\shortstack{Egon\\Schiele}} &
    \includegraphics[width=2cm,height=2cm]{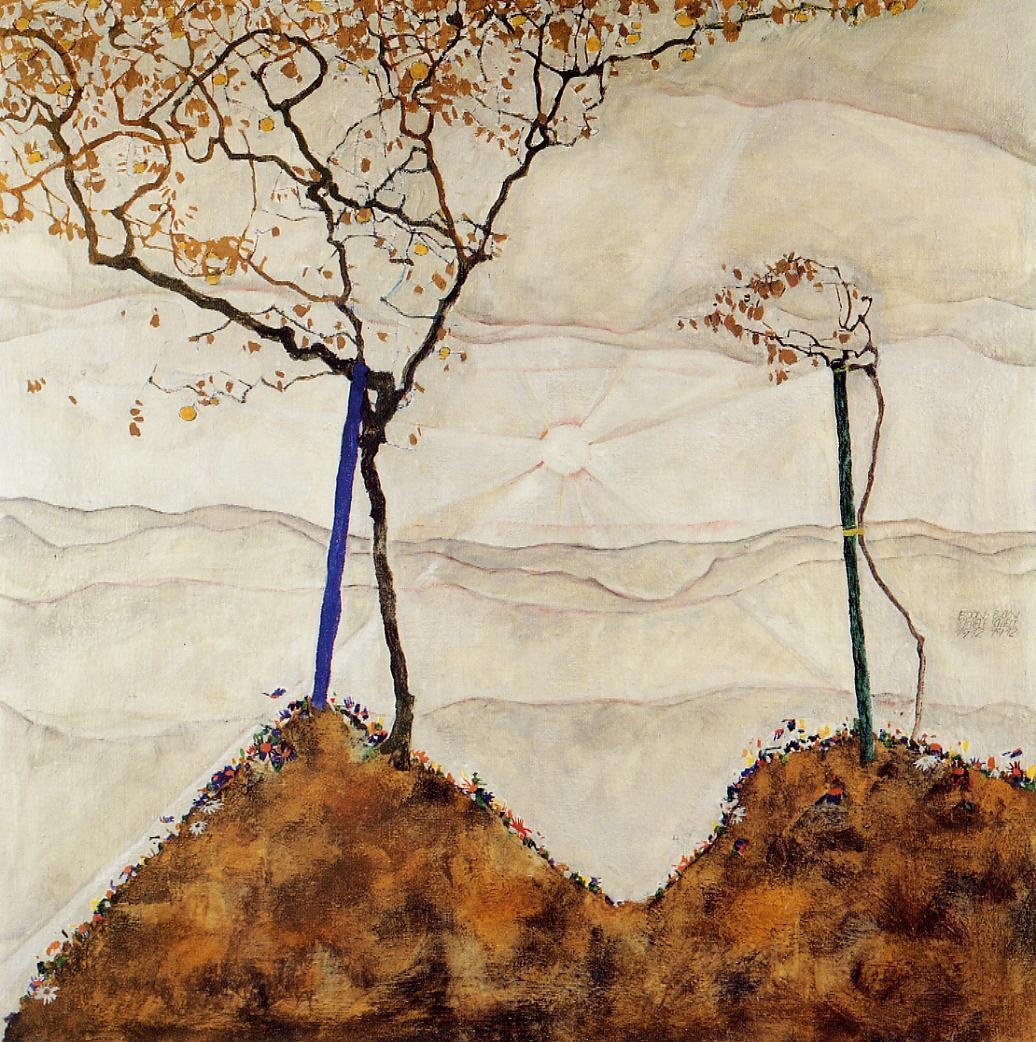} &
    \includegraphics[width=2cm,height=2cm]{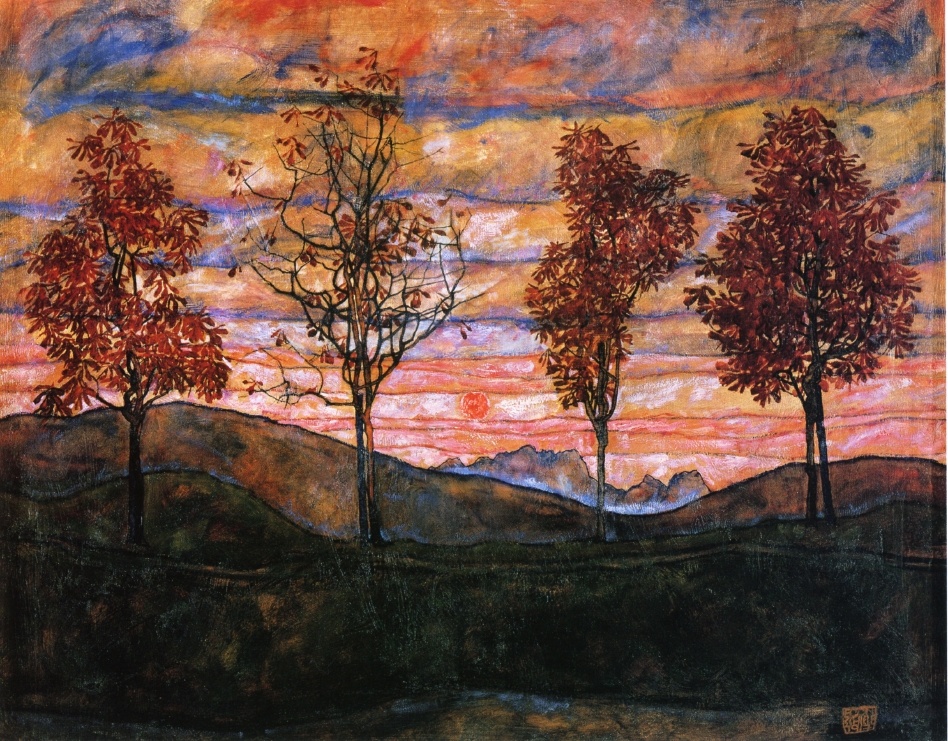} &
    \includegraphics[width=2cm,height=2cm]{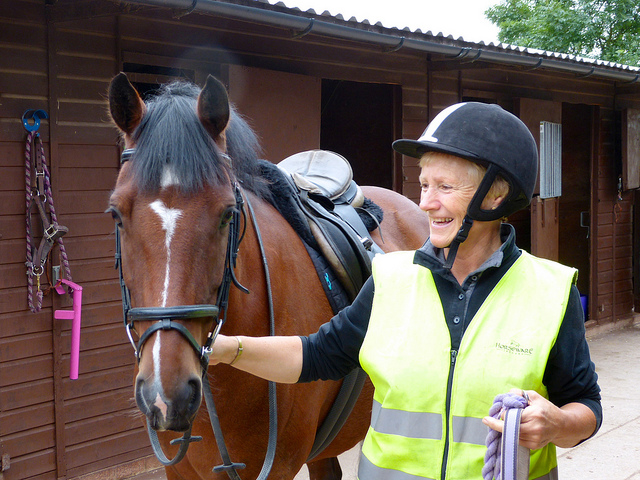} &
    \includegraphics[width=2cm,height=2cm]{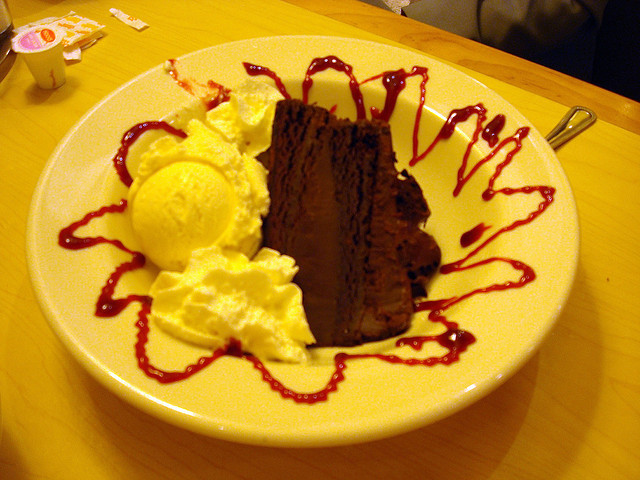} &
    \includegraphics[width=2cm,height=2cm]{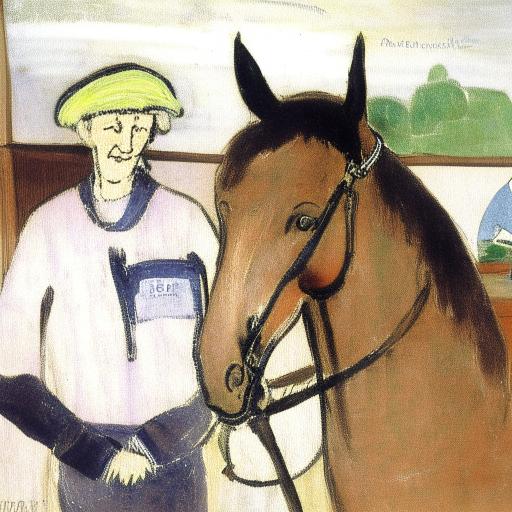} &
    \includegraphics[width=2cm,height=2cm]{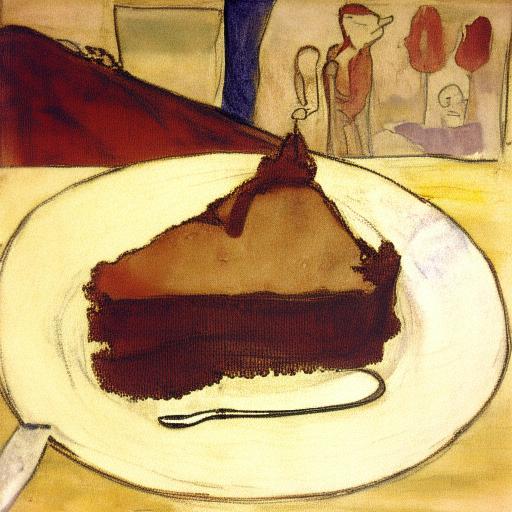} \\[0.4cm]

    \multirow{1}{*}{\centering\shortstack{Hans\\Baldung}} &
    \includegraphics[width=2cm,height=2cm]{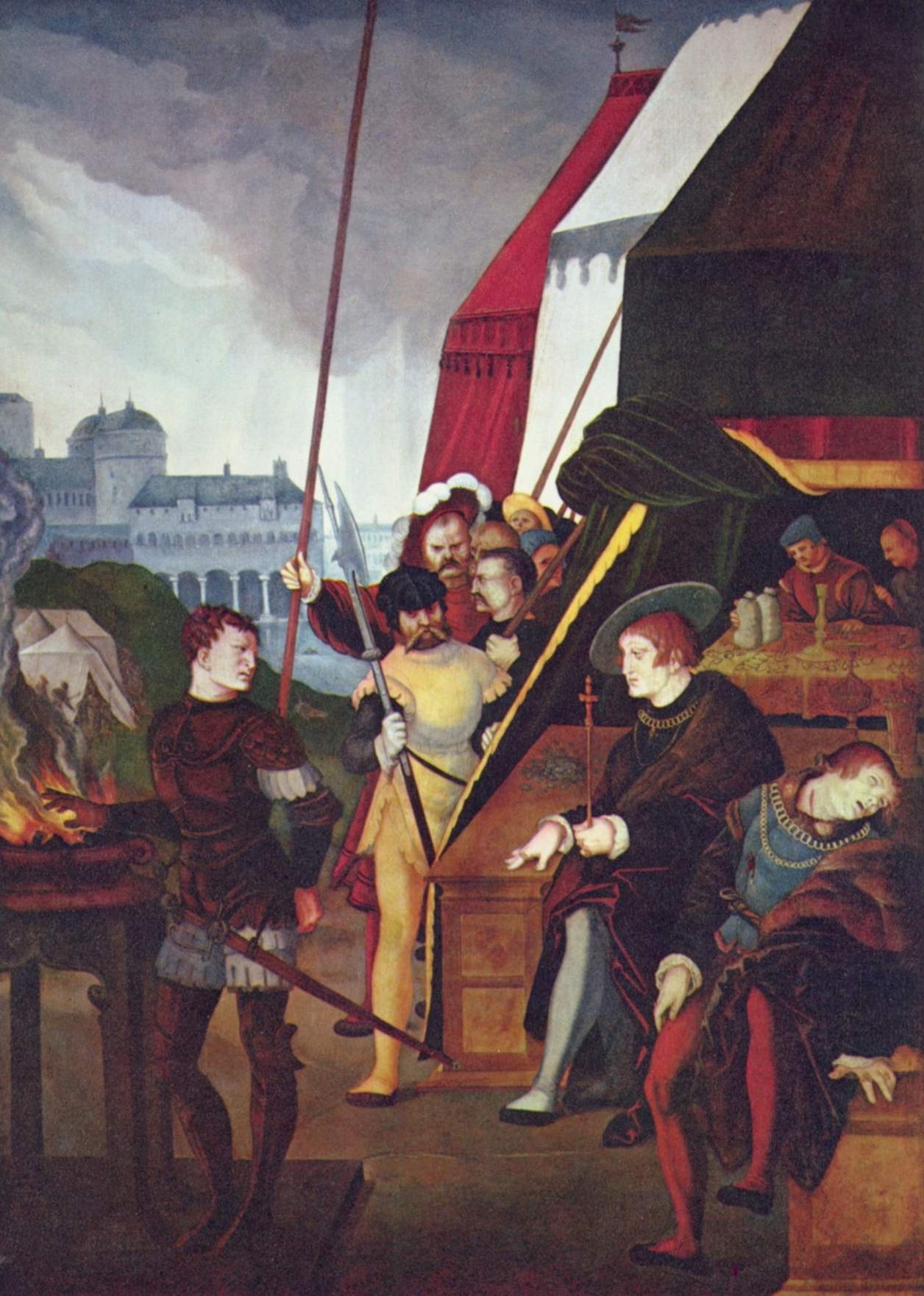} &
    \includegraphics[width=2cm,height=2cm]{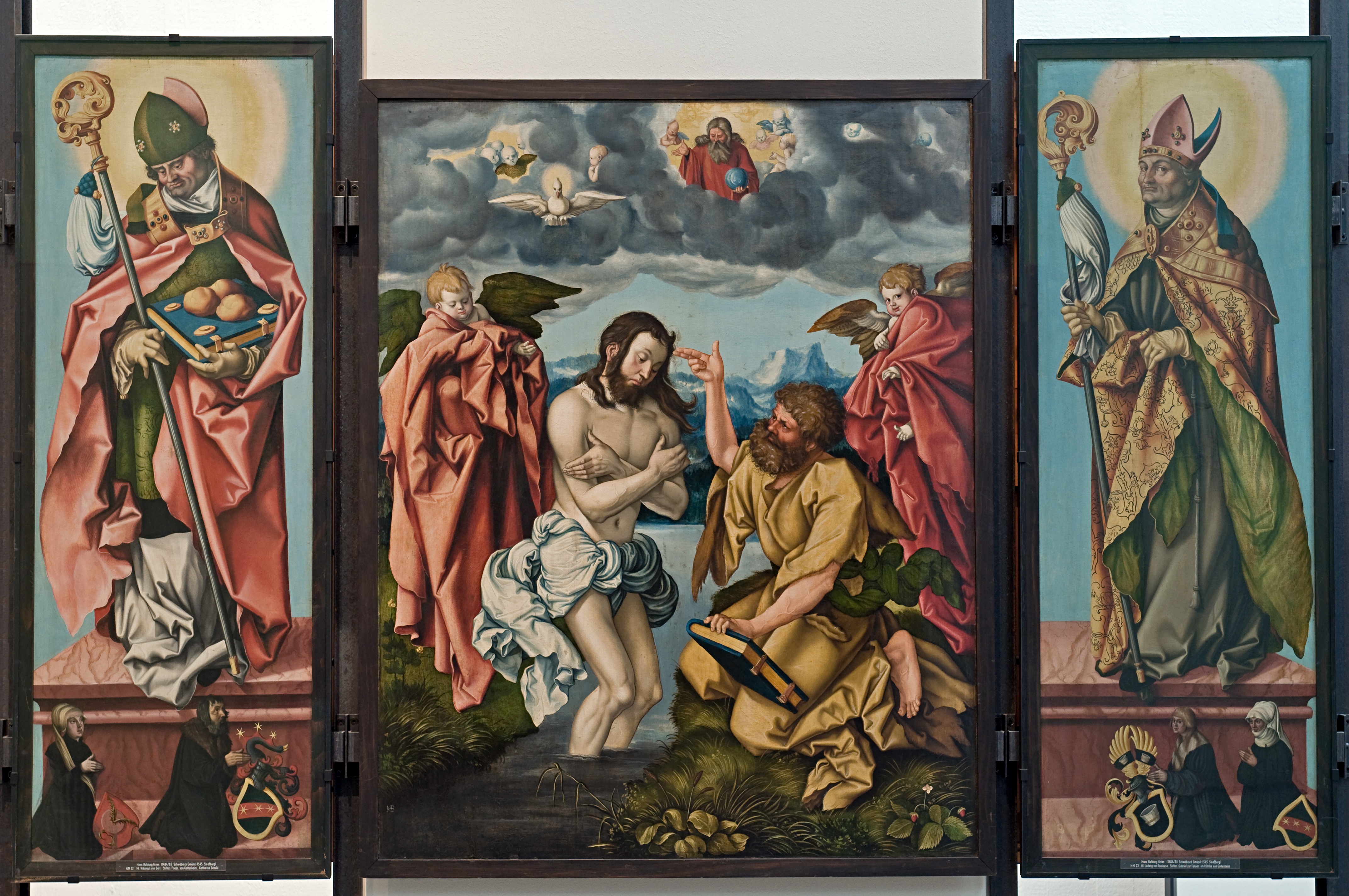} &
    \includegraphics[width=2cm,height=2cm]{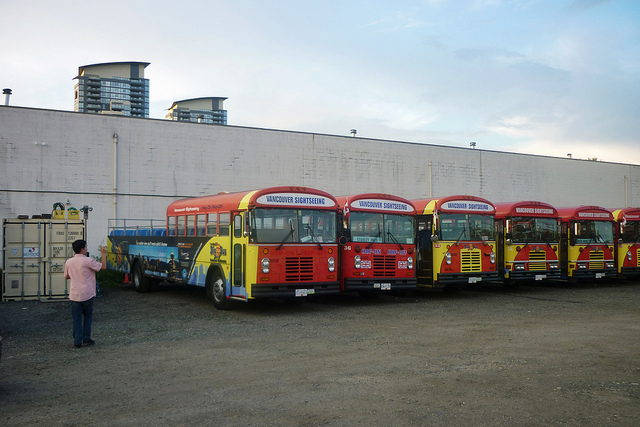} &
    \includegraphics[width=2cm,height=2cm]{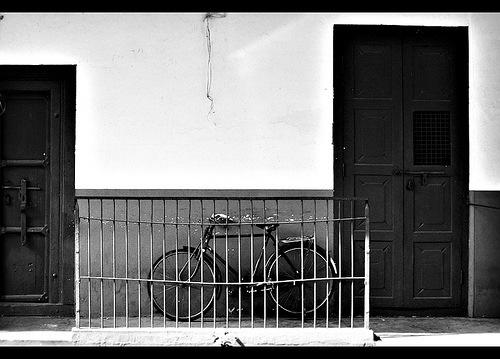} &
    \includegraphics[width=2cm,height=2cm]{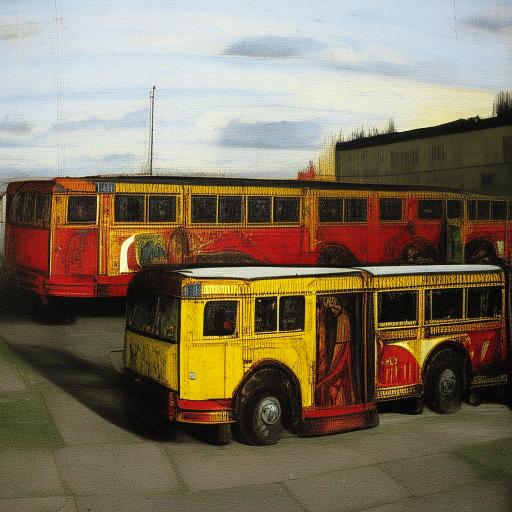} &
    \includegraphics[width=2cm,height=2cm]{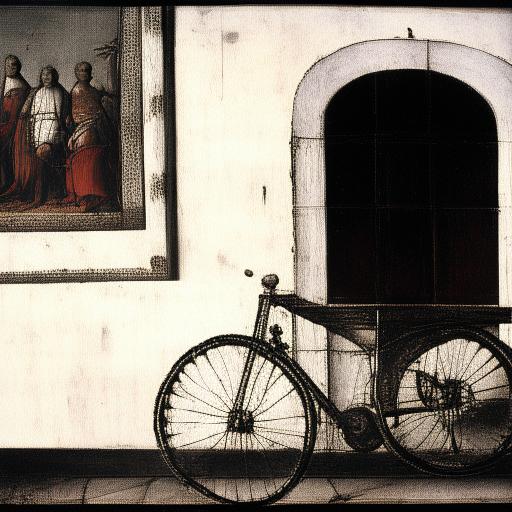} \\[0.4cm]

    \multirow{1}{*}{\centering Ivan Bilibin} &
    \includegraphics[width=2cm,height=2cm]{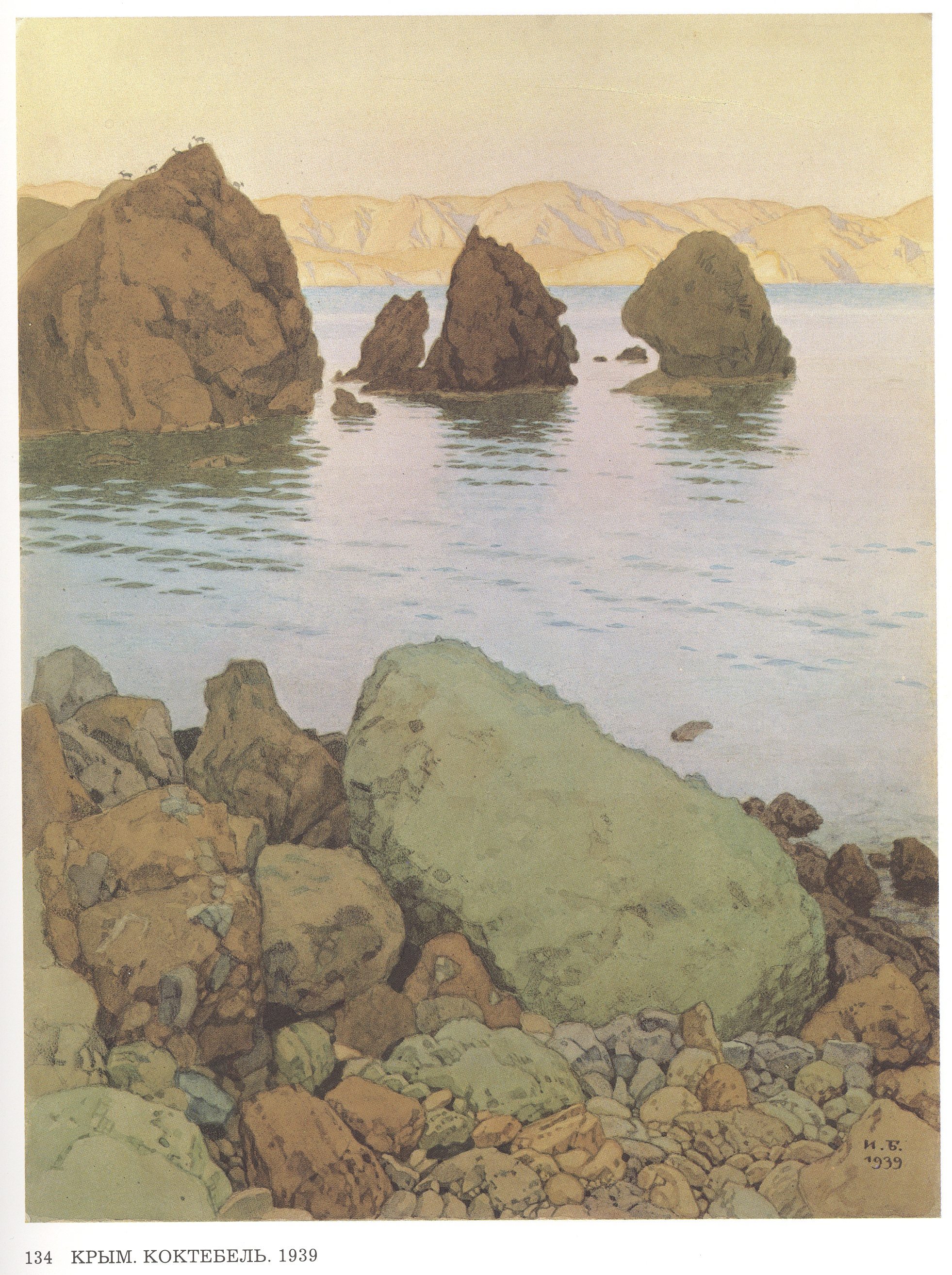} &
    \includegraphics[width=2cm,height=2cm]{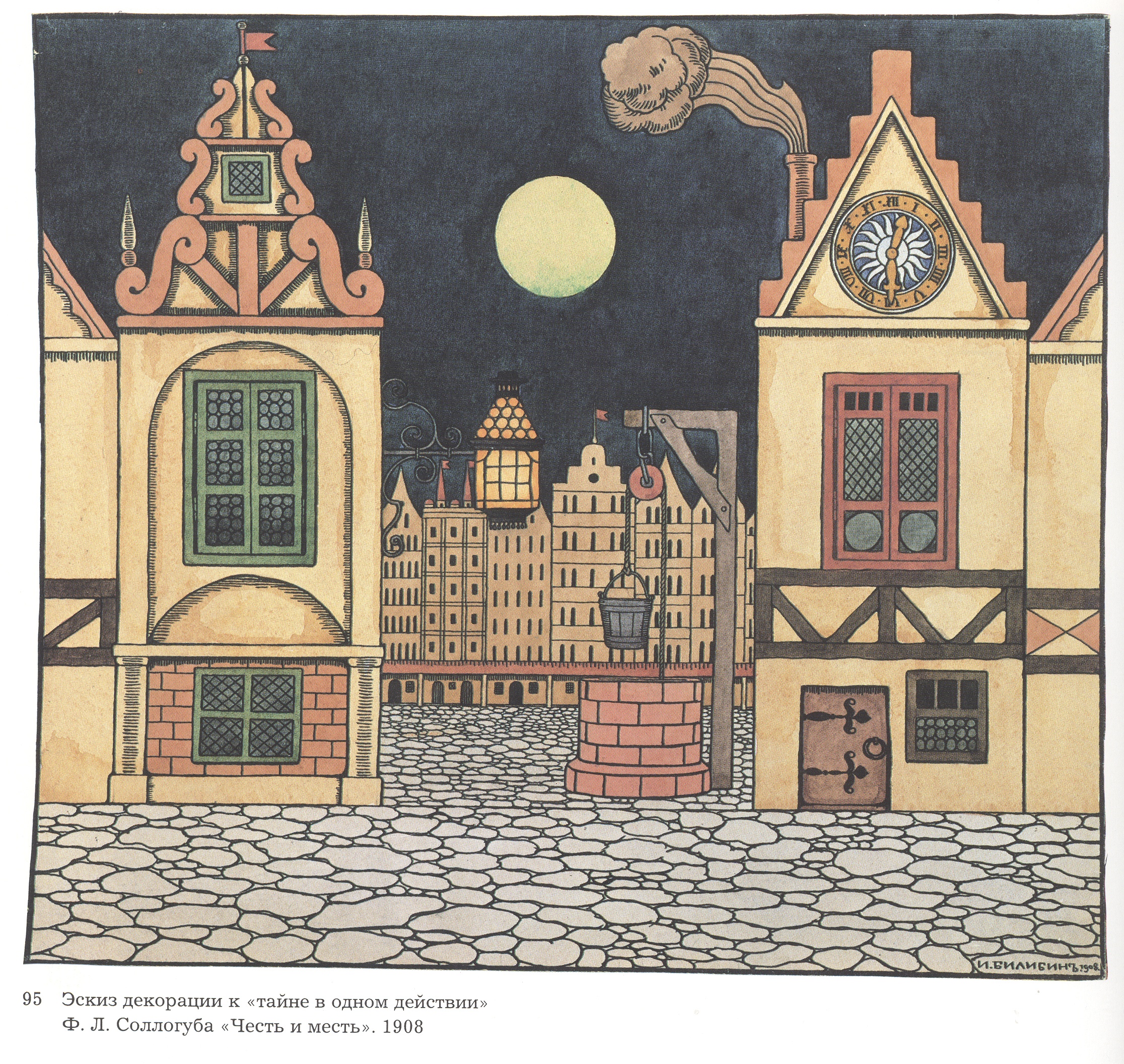} &
    \includegraphics[width=2cm,height=2cm]{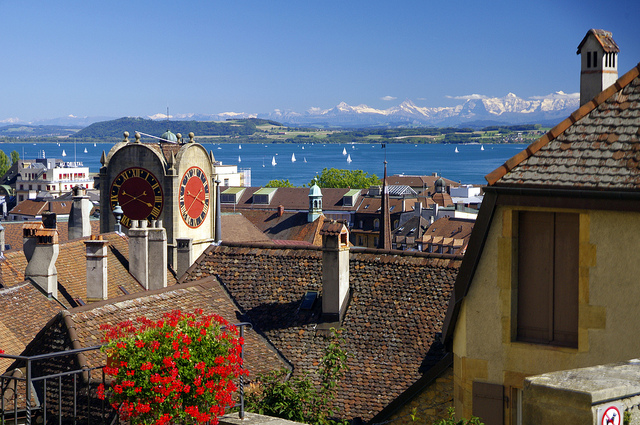} &
    \includegraphics[width=2cm,height=2cm]{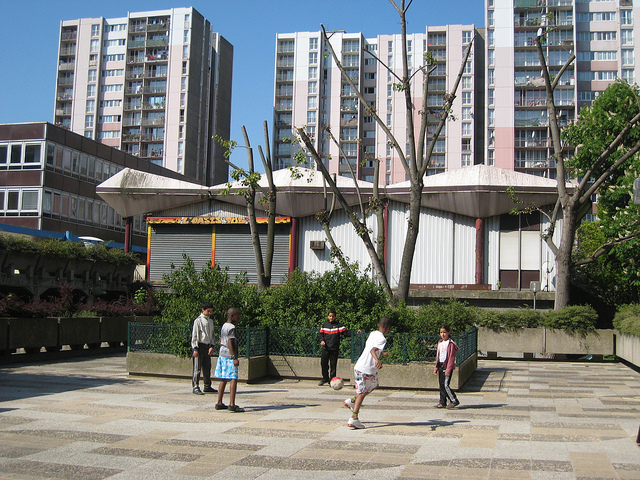} &
    \includegraphics[width=2cm,height=2cm]{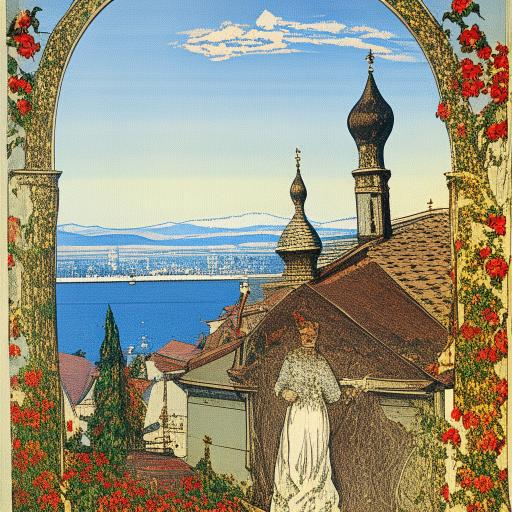} &
    \includegraphics[width=2cm,height=2cm]{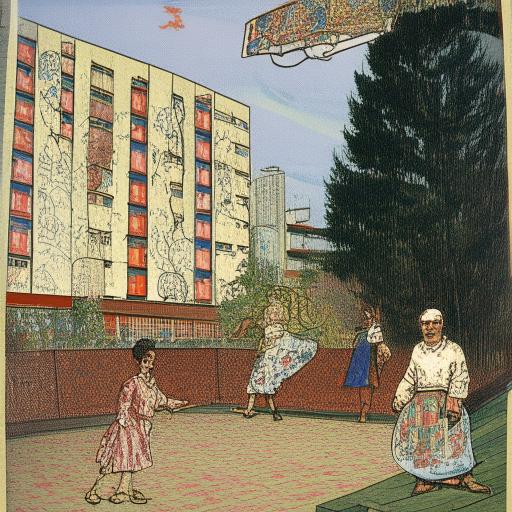} \\[0.4cm]

    \multirow{1}{*}{\centering Paul Gauguin} &
    \includegraphics[width=2cm,height=2cm]{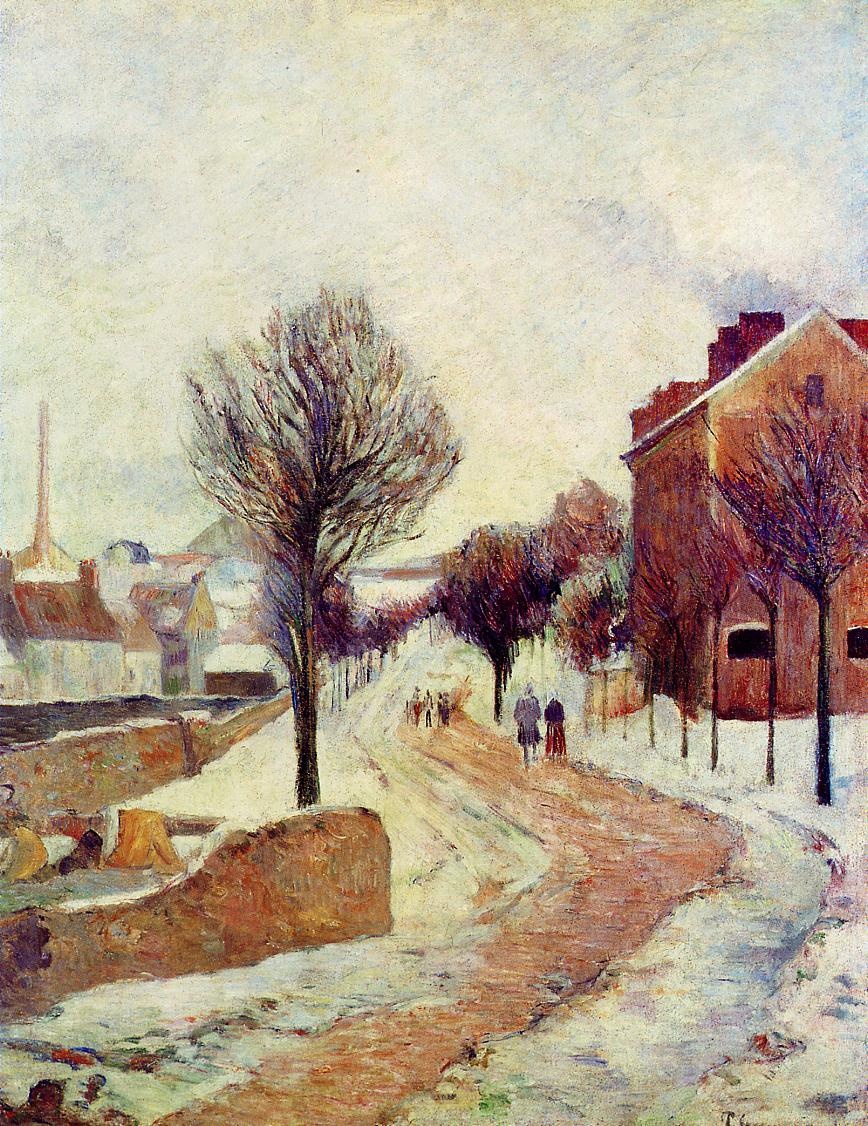} &
    \includegraphics[width=2cm,height=2cm]{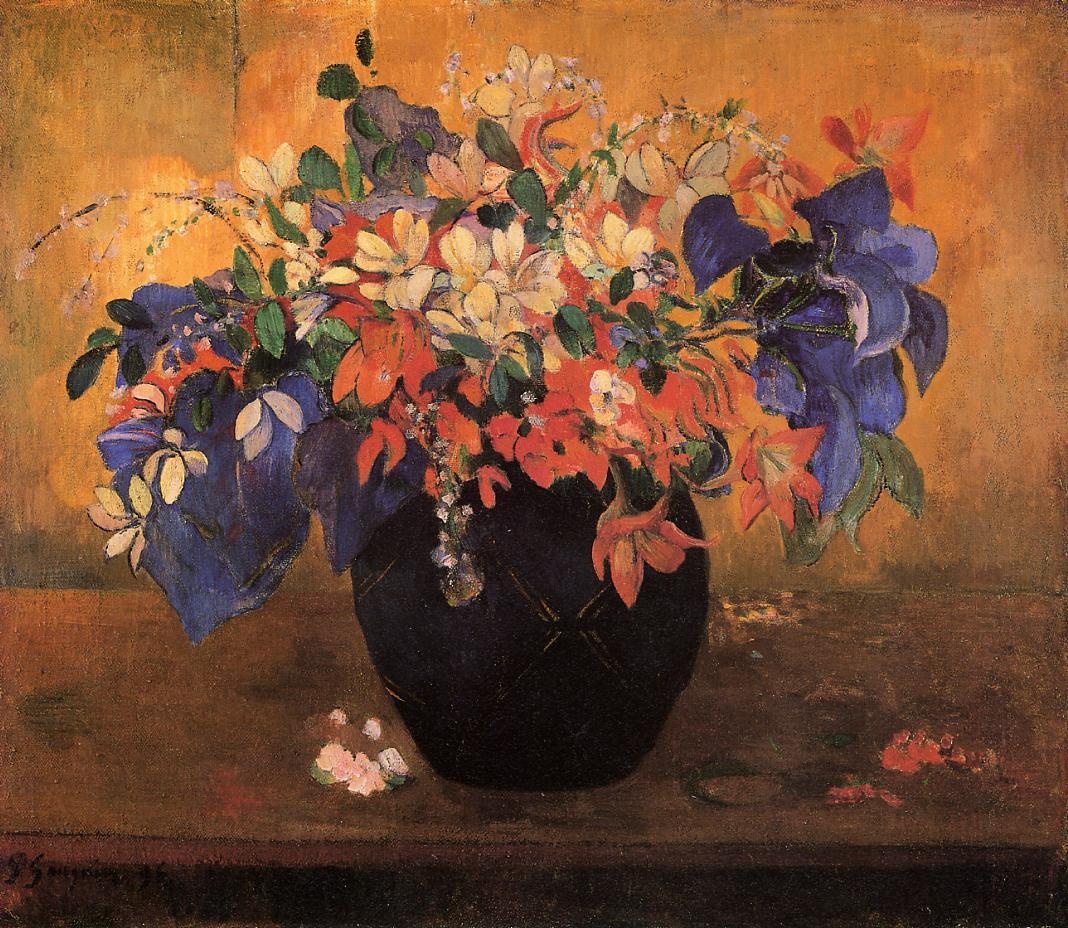} &
    \includegraphics[width=2cm,height=2cm]{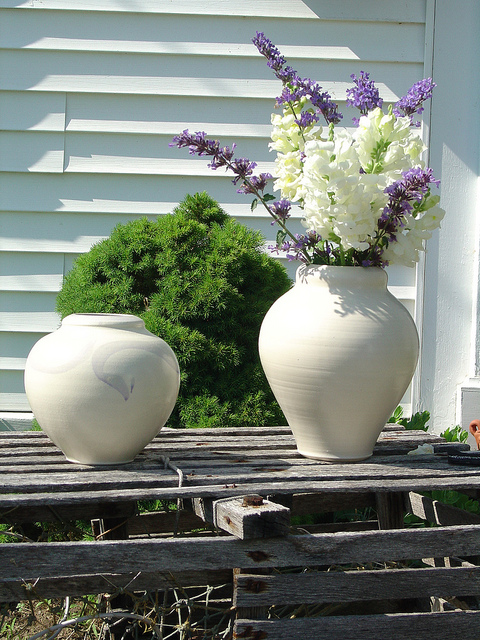} &
    \includegraphics[width=2cm,height=2cm]{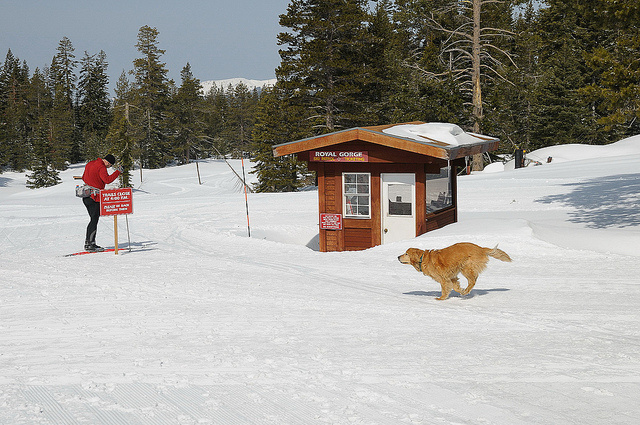}&
    \includegraphics[width=2cm,height=2cm]{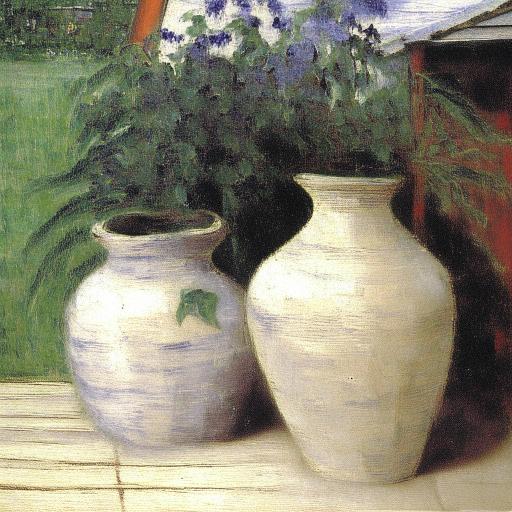} &
    \includegraphics[width=2cm,height=2cm]{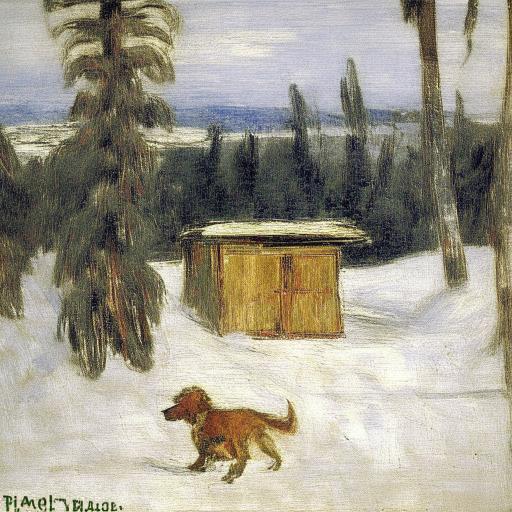} \\[0.4cm]

    \multirow{1}{*}{\centering\shortstack{Zinaida\\Serebriakova}} &
    \includegraphics[width=2cm,height=2cm]{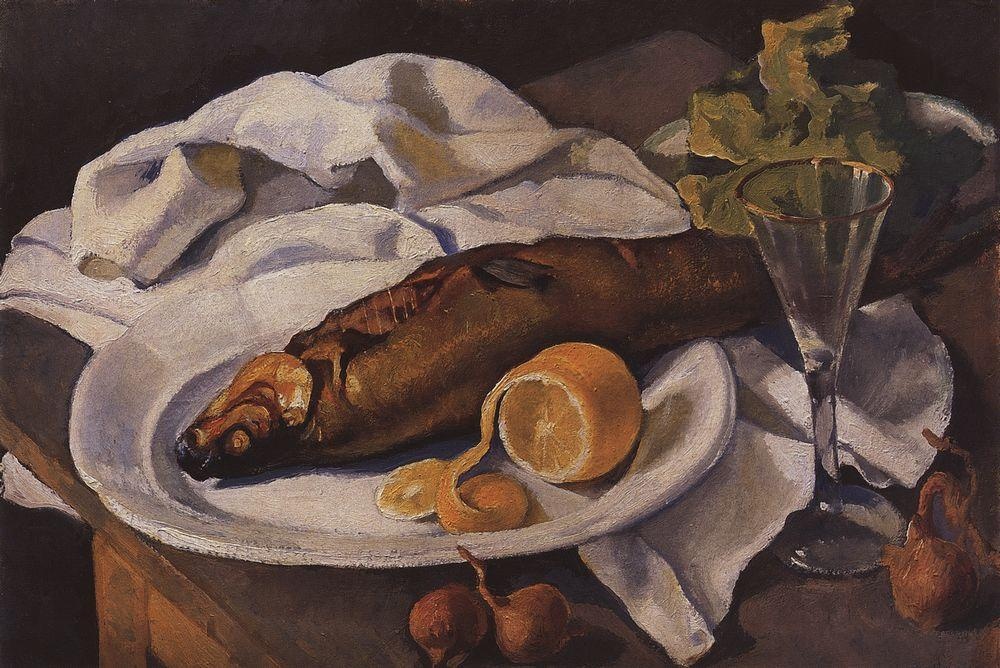} &
    \includegraphics[width=2cm,height=2cm]{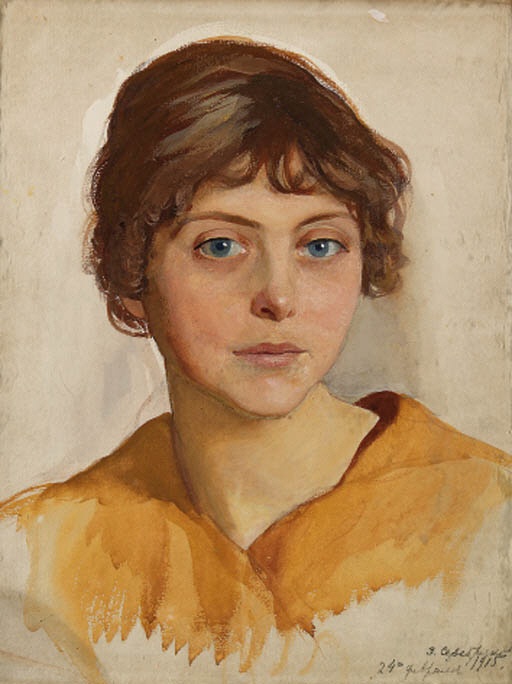} &
    \includegraphics[width=2cm,height=2cm]{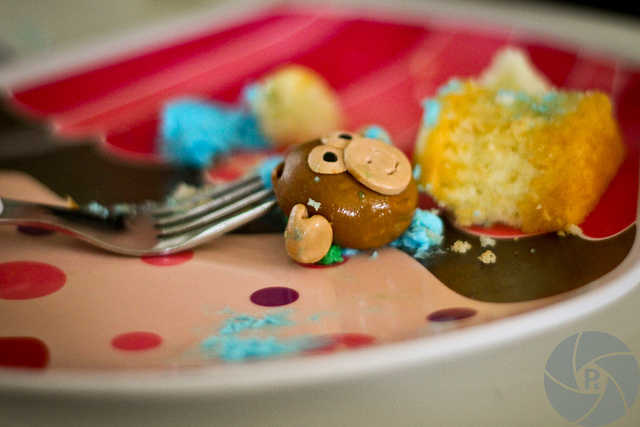} &
    \includegraphics[width=2cm,height=2cm]{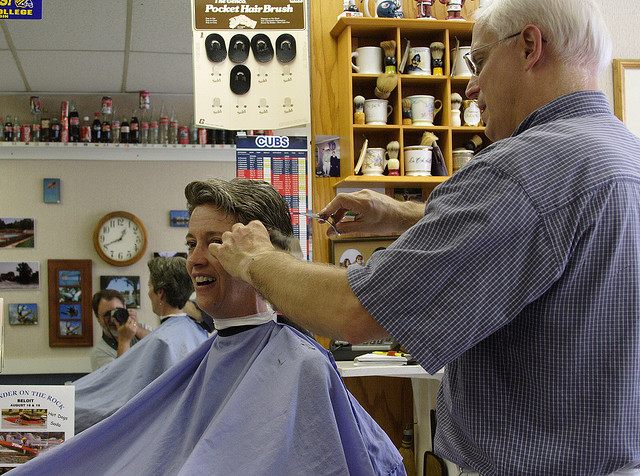} &
    \includegraphics[width=2cm,height=2cm]{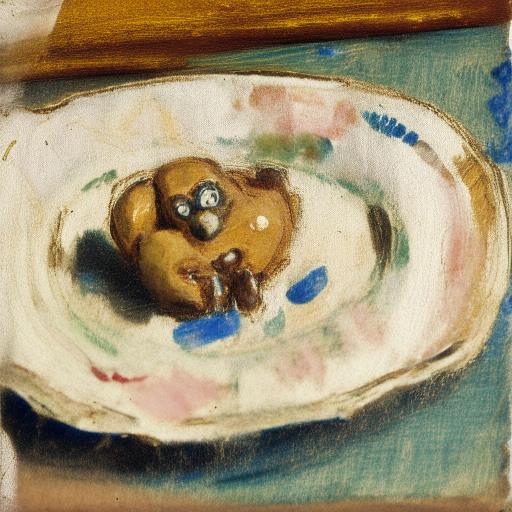} &
    \includegraphics[width=2cm,height=2cm]{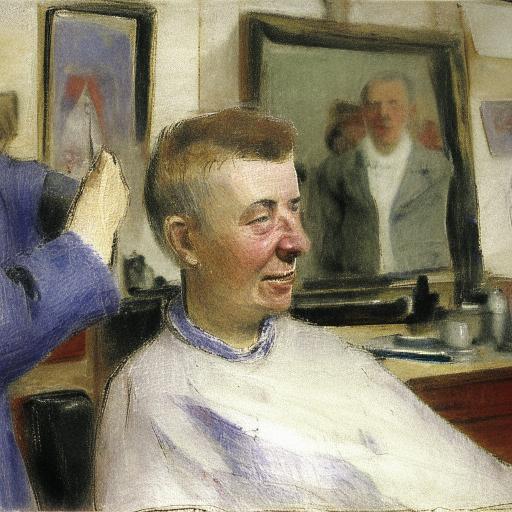} \\

    \bottomrule
    \end{tabular}

    \caption{
        \textbf{Additional Results: Qualitative Examples.} Additional images generated using \textit{LouvreSAE} Steering (namely Kandinsky 2.2 + \textit{LouvreSAE}-20K). In each row, the content images are transferred to the respective target style, whose reference samples are shown.
    }
    \label{fig:qual_baseline_comp}
\end{figure*}
\FloatBarrier

\end{document}